\documentclass[twoside,11pt]{article}
\usepackage[utf8]{inputenc}
\usepackage{jmlr2e}
\usepackage{graphicx}
\graphicspath{ {./figures/} }
\usepackage{listings}
\usepackage{cleveref}
\usepackage{caption}
\usepackage{subcaption}
\usepackage{svg}
\usepackage{soul}
\usepackage{xcolor}
\usepackage{tikz}
\usepackage{fancyvrb}
\usepackage{booktabs} 
\usepackage{multirow}
\usepackage[T5]{fontenc}
\usepackage{makecell}
\usepackage{csquotes}
\usepackage{scalerel}  
\usepackage{etoc} 

\jmlrheading{1}{2000}{1-48}{4/00}{10/00}{BigScience Workshop}

\newcommand{\carboneq}{CO\textsubscript{2}eq}
\newcommand{\carbonintensity}{gCO\textsubscript{2}eq/kWh}

\usepackage{ragged2e,array,longtable}
\newcolumntype{P}[1]{>{\RaggedRight\arraybackslash}p{#1}}
\newcolumntype{T}{c<{\ttfamily}}
\def\tokenizerregex{\scalerel*{\includegraphics{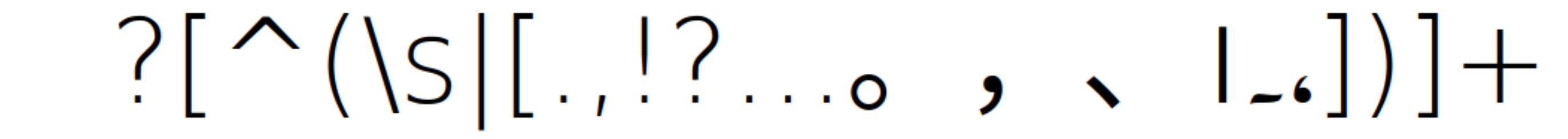}}{\textrm{\textbigcircle}}}

\ShortHeadings{BLOOM}{BigScience Workshop}   
\firstpageno{1}

\begin{document}

\title{BLOOM: A 176B-Parameter Open-Access Multilingual Language Model}


\author{BigScience Workshop\footnote{Please direct correspondence to \url{bigscience-contact@googlegroups.com}. A list of contributions is available in \Cref{sec:contributions}.}
\AND
Major Contributors \\
\names{Teven Le Scao, Angela Fan, Christopher Akiki, Ellie Pavlick, Suzana Ilić, Daniel Hesslow, Roman Castagné, Alexandra Sasha Luccioni, François Yvon, Matthias Gallé, Jonathan Tow, Alexander M. Rush, Stella Biderman, Albert Webson, Pawan Sasanka Ammanamanchi, Thomas Wang, Benoît Sagot, Niklas Muennighoff, Albert Villanova del Moral, Olatunji Ruwase, Rachel Bawden, Stas Bekman, Angelina McMillan-Major, Thomas Wolf, Iz Beltagy, Huu Nguyen, Lucile Saulnier, Samson Tan, Pedro Ortiz Suarez, Victor Sanh, Hugo Laurençon, Yacine Jernite, Julien Launay, Margaret Mitchell, Colin Raffel}
\AND
Dataset \\
\names{Aaron Gokaslan, Adi Simhi, Aitor Soroa, Albert Villanova del Moral, Alexandra Sasha Luccioni, Alham Fikri Aji, Amit Alfassy, Angelina McMillan-Major, Anna Rogers, Ariel Kreisberg Nitzav, Canwen Xu, Chenghao Mou, Chris Emezue, Christopher Akiki, Christopher Klamm, Colin Leong, Colin Raffel, Daniel van Strien, David Ifeoluwa Adelani, Dragomir Radev, Eduardo González Ponferrada, Efrat Levkovizh, Ethan Kim, Eyal Bar Natan, Francesco De Toni, Gérard Dupont, Germán Kruszewski, Giada Pistilli, Hady Elsahar, Hamza Benyamina, Hieu Tran, Hugo Laurençon, Huu Nguyen, Ian Yu, Idris Abdulmumin, Isaac Johnson, Itziar Gonzalez-Dios, Javier de la Rosa, Jenny Chim, Jesse Dodge, Jian Zhu, Jonathan Chang, Jörg Frohberg, Joseph Tobing, Joydeep Bhattacharjee, Khalid Almubarak, Kimbo Chen, Kyle Lo, Leandro Von Werra, Leon Weber, Long Phan, Loubna Ben allal, Lucile Saulnier, Ludovic Tanguy, Manan Dey, Manuel Romero Muñoz, Maraim Masoud, Margaret Mitchell, María Grandury, Mario Šaško, Max Huang, Maximin Coavoux, Mayank Singh, Mike Tian-Jian Jiang, Minh Chien Vu, Mohammad A. Jauhar, Mustafa Ghaleb, Nishant Subramani, Nora Kassner, Nurulaqilla Khamis, Olivier Nguyen, Omar Espejel, Ona de Gibert, Paulo Villegas, Pawan Sasanka Ammanamanchi, Pedro Ortiz Suarez, Peter Henderson, Pierre Colombo, Priscilla Amuok, Quentin Lhoest, Rheza Harliman, Rishi Bommasani, Roberto Luis López, Roman Castagné, Rui Ribeiro, Salomey Osei, Sampo Pyysalo, Samson Tan, Sebastian Nagel, Shamik Bose, Shamsuddeen Hassan Muhammad, Shanya Sharma, Shayne Longpre, Somaieh Nikpoor, Stanislav Silberberg, Stella Biderman, Suhas Pai, Suzana Ilić, Sydney Zink, Teven Le Scao, Thomas Wang, Tiago Timponi Torrent, Timo Schick, Tristan Thrush, Valentin Danchev, Vassilina Nikoulina, Veronika Laippala, Violette Lepercq, Vrinda Prabhu, Yacine Jernite, Zaid Alyafeai, Zeerak Talat}
\AND
Tokenization \\
\names{Arun Raja, Benjamin Heinzerling, Benoît Sagot, Chenglei Si, Colin Raffel, Davut Emre Taşar, Elizabeth Salesky, Lucile Saulnier, Manan Dey, Matthias Gallé, Pedro Ortiz Suarez, Roman Castagné, Sabrina J. Mielke, Samson Tan, Teven Le Scao, Thomas Wang, Wilson Y. Lee, Zaid Alyafeai}
\AND
Prompt Engineering \\
\names{Abheesht Sharma, Albert Webson, Alexander M. Rush, Alham Fikri Aji, Andrea Santilli, Antoine Chaffin, Arnaud Stiegler, Arun Raja, Canwen Xu, Colin Raffel, Debajyoti Datta, Dragomir Radev, Eliza Szczechla, Gunjan Chhablani, Han Wang, Harshit Pandey, Hendrik Strobelt, Jason Alan Fries, Jonathan Chang, Jos Rozen, Khalid Almubarak, Leo Gao, Lintang Sutawika, M Saiful Bari, Maged S. Al-shaibani, Manan Dey, Matteo Manica, Mike Tian-Jian Jiang, Nihal Nayak, Niklas Muennighoff, Rachel Bawden, Ryan Teehan, Samuel Albanie, Shanya Sharma, Sheng Shen, Srulik Ben-David, Stella Biderman, Stephen H. Bach, Taewoon Kim, Tali Bers, Teven Le Scao, Thibault Fevry, Thomas Wang, Thomas Wolf, Trishala Neeraj, Urmish Thakker, Victor Sanh, Vikas Raunak, Xiangru Tang, Zaid Alyafeai, Zheng-Xin Yong, Zhiqing Sun, Shaked Brody, Yallow Uri, Hadar Tojarieh}
\AND
Architecture and Objective \\
\names{Adam Roberts, Colin Raffel, Daniel Hesslow, Hady Elsahar, Hyung Won Chung, Iz Beltagy, Jaesung Tae, Jason Phang, Julien Launay, Lintang Sutawika, Lucile Saulnier, M Saiful Bari, Niklas Muennighoff, Ofir Press, Sheng Shen, Stas Bekman, Stella Biderman, Teven Le Scao, Thomas Wang, Vassilina Nikoulina, Victor Sanh, Zheng-Xin Yong}
\AND
Engineering \\
\names{Conglong Li, Deepak Narayanan, Hatim Bourfoune, Jared Casper, Jeff Rasley, Max Ryabinin, Mayank Mishra, Minjia Zhang, Mohammad Shoeybi, Myriam Peyrounette, Nicolas Patry, Niklas Muennighoff, Nouamane Tazi, Olatunji Ruwase, Omar Sanseviero, Patrick von Platen, Pierre Cornette, Pierre François Lavallée, Rémi Lacroix, Samyam Rajbhandari, Sanchit Gandhi, Shaden Smith, Stas Bekman, Stéphane Requena, Suraj Patil, Teven Le Scao, Thomas Wang, Tim Dettmers}
\AND
Evaluation and Interpretability \\
\names{Ahmed Baruwa, Albert Webson, Alexandra Sasha Luccioni, Alham Fikri Aji, Amanpreet Singh, Anastasia Cheveleva, Anne-Laure Ligozat, Arjun Subramonian, Aurélie Névéol, Charles Lovering, Dan Garrette, Deepak Tunuguntla, Dragomir Radev, Ehud Reiter, Ekaterina Taktasheva, Ekaterina Voloshina, Eli Bogdanov, Ellie Pavlick, François Yvon, Genta Indra Winata, Hailey Schoelkopf, Jaesung Tae, Jan-Christoph Kalo, Jekaterina Novikova, Jessica Zosa Forde, Jordan Clive, Jungo Kasai, Ken Kawamura, Khalid Almubarak, Liam Hazan, Lintang Sutawika, Manan Dey, Maraim Masoud, Margaret Mitchell, Marine Carpuat, Miruna Clinciu, Najoung Kim, Newton Cheng, Niklas Muennighoff, Oleg Serikov, Omer Antverg, Oskar van der Wal, Pawan Sasanka Ammanamanchi, Pierre Colombo, Rachel Bawden, Rui Zhang, Ruochen Zhang, Samson Tan, Sebastian Gehrmann, Shachar Mirkin, Shani Pais, Shanya Sharma, Shayne Longpre, Stella Biderman, Tatiana Shavrina, Thomas Scialom, Tian Yun, Tomasz Limisiewicz, Urmish Thakker, Vassilina Nikoulina, Verena Rieser, Vikas Raunak, Vitaly Protasov, Vladislav Mikhailov, Wilson Y. Lee, Yada Pruksachatkun, Yonatan Belinkov, Zachary Bamberger, Zdeněk Kasner, Zeerak Talat, Zheng-Xin Yong}
\AND
Broader Impacts \\
\names{Aaron Gokaslan, Alexandra Sasha Luccioni, Alham Fikri Aji, Alice Rueda, Amanda Pestana, Amir Feizpour, Ammar Khan, Amy Faranak, Ana Santos, Angelina McMillan-Major, Anthony Hevia, Antigona Unldreaj, Arash Aghagol, Arezoo Abdollahi, Aycha Tammour, Azadeh HajiHosseini, Bahareh Behroozi, Benjamin Ajibade, Bharat Saxena, Carlos Muñoz Ferrandis, Chenghao Mou, Minh Chien Vu, Christopher Akiki, Daniel McDuff, Danish Contractor, David Ifeoluwa Adelani, David Lansky, Davis David, Douwe Kiela, Duong A. Nguyen, Edward Tan, Emi Baylor, Ezinwanne Ozoani, Fatima Mirza, Frankline Ononiwu, Gérard Dupont, Giada Pistilli, Habib Rezanejad, Hessie Jones, Huu Nguyen, Ian Yu, Indrani Bhattacharya, Irene Solaiman, Irina Sedenko, Isar Nejadgholi, Jaesung Tae, Jenny Chim, Jesse Dodge, Jesse Passmore, Josh Seltzer, Julien Launay, Julio Bonis Sanz, Khalid Almubarak, Livia Dutra, Long Phan, Mairon Samagaio, Manan Dey, Maraim Masoud, Margaret Mitchell, Margot Mieskes, Marissa Gerchick, Martha Akinlolu, Michael McKenna, Mike Qiu, Muhammed Ghauri, Mykola Burynok, Nafis Abrar, Nazneen Rajani, Niklas Muennighoff, Nishant Subramani, Nour Elkott, Nour Fahmy, Olanrewaju Samuel, Olivier Nguyen, Paulo Villegas, Pawan Sasanka Ammanamanchi, Priscilla Amuok, Ran An, Rasmus Kromann, Ryan Hao, Samira Alizadeh, Sarmad Shubber, Shanya Sharma, Shayne Longpre, Silas Wang, Somaieh Nikpoor, Sourav Roy, Stas Bekman, Stella Biderman, Suhas Pai, Suzana Ilić, Sylvain Viguier, Teven Le Scao, Thanh Le, Tobi Oyebade, Trieu Le, Tristan Thrush, Yacine Jernite, Yoyo Yang, Zach Nguyen, Zeerak Talat, Zheng-Xin Yong}
\AND
Applications \\
\names{Abhinav Ramesh Kashyap, Albert Villanova del Moral, Alfredo Palasciano, Alison Callahan, Anima Shukla, Antonio Miranda-Escalada, Ayush Singh, Benjamin Beilharz, Bo Wang, Caio Brito, Carlos Muñoz Ferrandis, Chenxi Zhou, Chirag Jain, Christopher Akiki, Chuxin Xu, Clémentine Fourrier, Daniel León Periñán, Daniel Molano, Daniel van Strien, Danish Contractor, David Lansky, Debajyoti Datta, Dian Yu, Enrique Manjavacas, Fabio Barth, Florian Fuhrimann, Francesco De Toni, Gabriel Altay, Giyaseddin Bayrak, Gully Burns, Helena U. Vrabec, Imane Bello, Ishani Dash, Jason Alan Fries, Javier de la Rosa, Jenny Chim, Jihyun Kang, John Giorgi, Jonas Golde, Jose David Posada, Karthik Rangasai Sivaraman, Leon Weber, Lokesh Bulchandani, Lu Liu, Luisa Shinzato, Madeleine Hahn de Bykhovetz, Maiko Takeuchi, Marc Pàmies, Maria A Castillo, Marianna Nezhurina, Mario Sänger, Matthias Samwald, Michael Cullan, Michael Weinberg, Michiel De Wolf, Mina Mihaljcic, Minh Chien Vu, Minna Liu, Moritz Freidank, Myungsun Kang, Natasha Seelam, Nathan Dahlberg, Nicholas Michio Broad, Nikolaus Muellner, Pascale Fung, Patrick Haller, Ramya Chandrasekhar, Renata Eisenberg, Robert Martin, Rodrigo Canalli, Rosaline Su, Ruisi Su, Samuel Cahyawijaya, Samuele Garda, Shamik Bose, Shlok S Deshmukh, Shubhanshu Mishra, Sid Kiblawi, Simon Ott, Sinee Sang-aroonsiri, Srishti Kumar, Stefan Schweter, Stella Biderman, Stephen H. Bach, Sushil Bharati, Tanmay Laud, Théo Gigant, Tomoya Kainuma, Trishala Neeraj, Wojciech Kusa, Yanis Labrak, Yash Shailesh Bajaj, Yash Venkatraman, Yifan Xu, Yingxin Xu, Yu Xu, Zhe Tan, Zhongli Xie, Zifan Ye}
\AND
Organization \\
\names{Angela Fan, Christopher Akiki, Douwe Kiela, Giada Pistilli, Margot Mieskes, Mathilde Bras, Matthias Gallé, Suzana Ilić, Yacine Jernite, Younes Belkada, Thomas Wolf}
}
\editor{}

\maketitle

\begin{abstract}
Large language models (LLMs) have been shown to be able to perform new tasks based on a few demonstrations or natural language instructions.
While these capabilities have led to widespread adoption, most LLMs are developed by resource-rich organizations and are frequently kept from the public.
As a step towards democratizing this powerful technology, we present BLOOM, a 176B-parameter open-access language model designed and built thanks to a collaboration of hundreds of researchers.
BLOOM is a decoder-only Transformer language model that was trained on the ROOTS corpus, a dataset comprising hundreds of sources in 46 natural and 13 programming languages (59 in total).
We find that BLOOM achieves competitive performance on a wide variety of benchmarks, with stronger results after undergoing multitask prompted finetuning.
To facilitate future research and applications using LLMs, we publicly release our models and code under the Responsible AI  License.\footnote{\href{https://hf.co/bigscience/bloom}{\texttt{hf.co/bigscience/bloom}}}
\end{abstract}


\section{Introduction}

Pretrained language models have become a cornerstone of modern natural language processing (NLP) pipelines because they often produce better performance from smaller quantities of labeled data.
The development of ELMo~\citep{peters2018deep}, ULMFiT~\citep{howard2018universal}, GPT~\citep{radford2018improving}, and BERT~\citep{devlin2019bert} led to the widespread use of pretrained models as an initialization for finetuning on downstream tasks.
The subsequent finding that pretrained language models can perform useful tasks without any additional training~\citep{radford2019language,brown2020language} further demonstrated their utility.
In addition, the empirical observation that a language model's performance tends to increase as the model is made larger---sometimes predictably~\citep{hestness2017deep,kaplan2020scaling,hoffmann2022training} and sometimes suddenly~\citep{wei2022emergent}---has led to a trend of increasing scale~\citep{zeng2021pangu, rae2021scaling, smith2022using, chowdhery2022palm}. 
Apart from environmental concerns~\citep{strubell2019energy,lacoste2019quantifying,schwartz2020green}, the costs of training large language models (LLMs) are only affordable for well-resourced organizations.
Furthermore, until recently, most LLMs were not publicly released.
As a result, the majority of the research community has been excluded from the development of LLMs.
This exclusion has had concrete consequences; for example, most LLMs are primarily trained on English-language text (with notable exceptions in Chinese and Korean, e.g.~\citealp{wang2021ernie,zeng2021pangu,kim2021changes}).

To address these issues, we present the BigScience Large Open-science Open-access Multilingual Language Model (BLOOM,~\citealp{bloom_model}).
BLOOM is a 176 billion parameter language model trained on 46 natural languages and 13 programming languages that was developed and released by a collaboration of hundreds of researchers.
The compute for training BLOOM was provided through a French public grant from GENCI and IDRIS, leveraging IDRIS' Jean Zay supercomputer.
To build BLOOM, we undertook a thorough design process for each of its components, including the training dataset (\Cref{sec:dataset}), model architecture and training objective (\Cref{sec:modeling}), and engineering strategy for distributed learning (\Cref{sec:engineering}).
We also performed an analysis of the model's capabilities (\Cref{sec:evaluation}).
Our overall aim is not only to publicly release a large-scale multilingual language model with performance comparable to recently developed systems, but also to document the coordinated process that went into its development (\Cref{sec:organization}).
The purpose of this paper is to provide a high-level overview of these design steps while referencing the individual reports we produced over the course of developing BLOOM.

\section{Background}
\label{sec:background}

Before describing the BLOOM model itself, in this section we provide necessary background on LLMs as well as an organizational overview of the BigScience effort.

\subsection{Language Modeling}
Language modeling refers to the task of modeling the probability of a sequence of tokens in a text \citep{shannon1948mathematical}, where a token is a unit of text (e.g.\ word, subword, character or byte, etc., as discussed by~\citealp{mielke2021between}).
In this work (and in most current applications of language modeling) we model the joint probability of tokens in a text as:
\begin{equation}
    p(x) = p(x_1, \ldots, x_T) = \prod_{t = 1}^T p(x_t | x_{< t})
\end{equation}
where $x$ is a sequence of tokens, $x_t$ is the $t^{\mathrm{th}}$ token, and $x_{< t}$ is the sequence of tokens preceding $x_t$.
This approach is referred to as autoregressive language modeling and can be seen as iteratively predicting the probability of the next token.

\paragraph{Early Language Models}
Language models have a long history of application in NLP.
Early language models (such as those developed by~\citealp{shannon1948mathematical}) were primarily $n$-gram models that estimate the probability of a length-$n$ sequence of tokens in accordance with the number of times it appears in a training corpus.
In practice, $n$-gram models face two major issues: first, they grow exponentially in size as $n$ is increased; and second, they have no direct way of producing a probability for a sequence of tokens that does not appear in their training data.
Advances on these problems enabled $n$-gram models to see widespread use across most areas of NLP~\citep{goodman2001bit}.

\paragraph{Neural Language Models}
An alternative to $n$-gram models, first proposed by~\citet{miikkulainen1991natural} and \citet{schmidhuber1996sequential} and later popularized by~\citet{bengio2000neural}, is to use a neural network to estimate the probability of the next token given prior tokens.
While early work used feed-forward networks with a fixed-length history window,~\citet{mikolov2010recurrent,sutskever2011generating,graves2013generating} proposed to use recurrent neural networks instead and found that this significantly improved performance.
More recently, language models based on the Transformer architecture~\citep{vaswani2017attention} were shown to be more effective than recurrent neural networks~\citep{radford2018improving,al2019character,kaplan2020scaling}.
Consequently, the Transformer has become the \textit{de facto} choice for language models.

\paragraph{Transfer Learning}
In tandem with advances in language modeling using neural networks, NLP pipelines have increasingly adopted the framework of transfer learning.
In transfer learning, the parameters of a model are first pretrained on a data-rich task before being finetuned on a downstream task.
A historically common approach to obtaining pretrained parameters were word vectors~\citep{mikolov2013distributed} trained so that the dot product of co-occurring word vectors is large.
However, subsequent work by~\citet{peters2018deep,howard2018universal,radford2018improving,devlin2019bert} showed that the framework of~\citet{collobert2011natural}, where the entire model is pretrained before being finetuned, can attain stronger performance.
In particular,~\citet{radford2018improving,devlin2019bert} demonstrated strong results using pretrained Transformer language models, prompting work on progressively better models~\citep[etc.]{liu2019roberta,yang2019xlnet,lewis2020bart,raffel2020exploring,zhang2019ernie}.

\paragraph{Few- and Zero-Shot Learning}
While finetuning a pretrained model remains an effective way of attaining high performance with limited labeled data, a parallel line of work has demonstrated that pretrained language models can be induced to perform tasks without any subsequent training.
After~\citet{vinyals2015neural} observed limited task-performing behavior in a neural dialog model,~\citet{radford2019language} later demonstrated that Transformer-based language models trained on text scraped from the web could perform various tasks to varying degrees.
Notably,~\citet{radford2019language} found that performance improved with model scale, inspiring work to characterize~\citep{kaplan2020scaling,hoffmann2022training} and exploit~\citep{shoeybi2019megatron,brown2020language,smith2022using,chowdhery2022palm,rae2021scaling,wang2021ernie,zeng2021pangu,zhang2022opt} the benefits of scale.
A major factor in the success of this approach is the way that task-specific examples are formatted when fed into the model.
\citet{brown2020language} popularized the idea of designing ``prompts'' that provide natural-language descriptions of the task and also allow inputting a few demonstrations of input-output behavior.

\paragraph{Social Limitations of LLM Development}
While the continued increase in the size of large language models 
has resulted in improvements
across a wide range of tasks,
it has also 
exacerbated issues with their development and use~\citep{bender2021dangers}. The computational expense of large models also prohibits the majority of the research community from participating in their development, evaluation and routine use.
Moreover, the computational costs have also lead to concerns about the carbon footprint stemming from the training and use of large language models
~\citep{strubell2019energy,lacoste2019quantifying,schwartz2020green,bannour-etal-2021-evaluating}, and existing carbon footprint studies have likely under-estimated emissions~\citep{bannour-etal-2021-evaluating}. 
Contributing to an increase in the global carbon footprint exacerbates climate change which most severely affects already-marginalized communities \citep{westra2001faces}.
Furthermore, the concentration of resources within a handful of (typically industrial) institutions with primarily technical expertise hinders prospects for an inclusive, collaborative, and reliable governance of the technology. First, public narratives about the technology that are driven by industry actors can lead to inflated expectations about its suitability for use~\citep{brennen2018hype,brennen2022expect}, leading to misaligned research and policy priorities~\citep{raji2022fallacy} and potentially dire consequences in e.g.\ medical applications~\citep{wong2021sepsis}. Second, in a world mediated by technology, choices at all stages of its development end up shaping people's lives in a way that can be most closely compared to regulations~\citep{winner1977technology,winner2017artifacts}, albeit without the same explicit consultation of stakeholders in the process. 
When the development efforts are guided by prioritizing internal definitions of performance over their impact on society, the values of the developers come to be emphasized over those of the direct and indirect users \citep{birhane2022values}.
Despite the substantial social dangers in allowing this technology to be developed unilaterally by corporations, EleutherAI \citep{phang2022eleutherai} was the only non-corporate entity outside of China that was developing large language models before the BigScience Workshop was convened.

\subsection{BigScience}\label{sec:organization}

\paragraph{Participants}
BLOOM's development was coordinated by BigScience, an open research collaboration whose goal was the public release of an LLM.
The project started after being awarded by GENCI a compute grant on its Jean Zay supercomputer at IDRIS/CNRS.
It was initially built around a concerted effort from Hugging Face and the French NLP community (the ``founding members''), and quickly opened up to grow into a broader international collaboration to support its aims of linguistic, geographical, and scientific diversity.
In the end, over 1200 people registered as participants in BigScience and were given access to its communication channels.
They had background not only in machine learning and computer science, but also linguistics, statistics, socio-cultural anthropology, philosophy, law, and other fields.
Of those, hundreds of individuals have directly contributed to one of the project’s released artifacts.
While the largest number of participants ultimately originated from the US, 38 countries were represented.

\begin{figure}[ht]
\centering
\includegraphics[width=\textwidth]{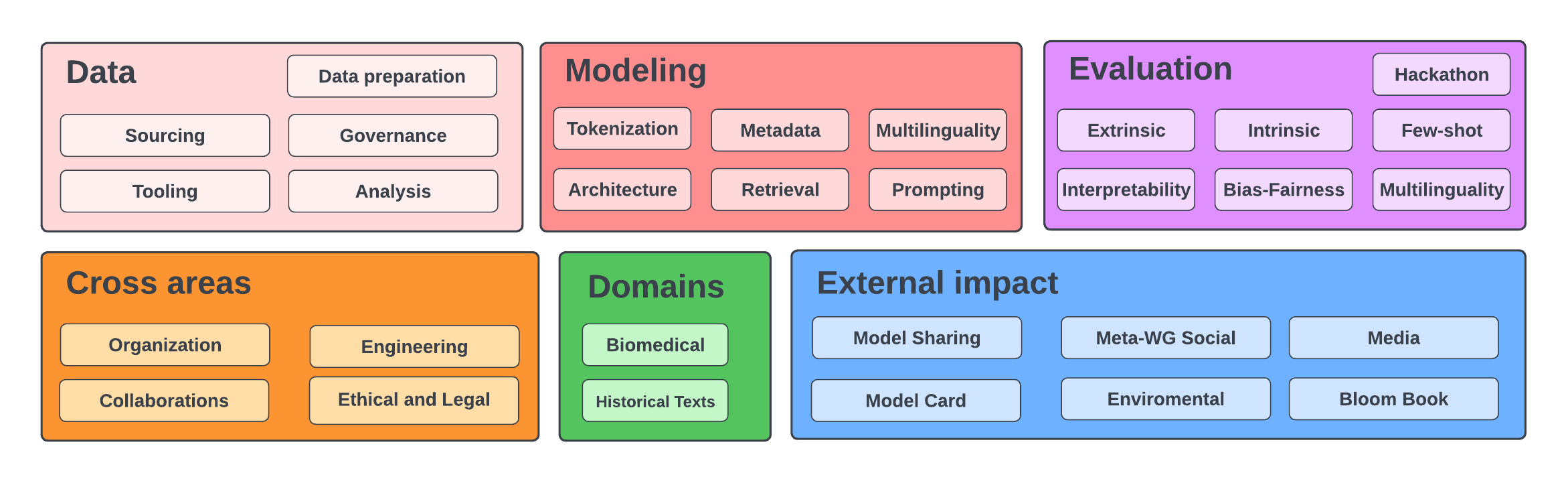}
\caption{Organization of BigScience working groups.}
\label{fig:workinggroups}
\end{figure}

\paragraph{Organization}
The set of related research questions tackled by the BigScience effort was reflected in the project's organization into working groups. 
Each working group comprised several participants with various levels of involvement, including chairs whose role was to self-organize around a specific aspect of the overall project. 
Importantly, participants were encouraged to join more than one working group in order to share experiences and information, which resulted in the set of 30 working groups presented in Figure~\ref{fig:workinggroups}. Most of the working groups focused on tasks directly linked to the development of BLOOM. In addition, a few groups focused on the evaluation of LLMs and dataset development in specific domains, such as biomedical texts~\citep{bigbio} and historical texts~\citep{de-toni-etal-2022-entities}.
A larger overview of the motivations behind this initiative, its history and some of the lessons learned can be found in \citet{bigscience2022org}.

\paragraph{Ethical Considerations within BigScience}
In order to acknowledge and start addressing social limitations of LLM development within BigScience, the workshop relied on a collaboratively designed Ethical Charter\footnote{\href{https://bigscience.huggingface.co/blog/bigscience-ethical-charter}{\texttt{bigscience.huggingface.co/blog/bigscience-ethical-charter}}} and original research on applicable regulations in jurisdictions outside of the US\footnote{\href{https://bigscience.huggingface.co/blog/legal-playbook-for-natural-language-processing-researchers}{\texttt{bigscience.huggingface.co/blog/legal-playbook-for-natural-language-processing-researchers}}} to guide its choices throughout the project. In particular, the charter emphasizes values of \textbf{inclusivity and diversity}, \textbf{openness and reproducibility}, and \textbf{responsibility} in various aspects of the organization~\citep{bigscience2022org}. Each of these values are showcased in different ways in the dataset curation (Section~\ref{sec:dataset}), modeling (Section~\ref{sec:modeling}), engineering (Section~\ref{sec:engineering}), evaluation (Section~\ref{sec:evaluation}), and other social impact (throughout) aspects of the project.

\section{BLOOM}

In this section, we document the design of BLOOM, including its training dataset (\Cref{sec:dataset}), architecture (\Cref{sec:modeling}), tokenizer (\Cref{sec:tokenization}), computing infrastructure (\Cref{sec:engineering}), and training hyperparameters (\Cref{sec:training}).

\subsection{Training Dataset}
\label{sec:dataset}

\begin{figure}
\centering
\includegraphics[width=0.7\textwidth]{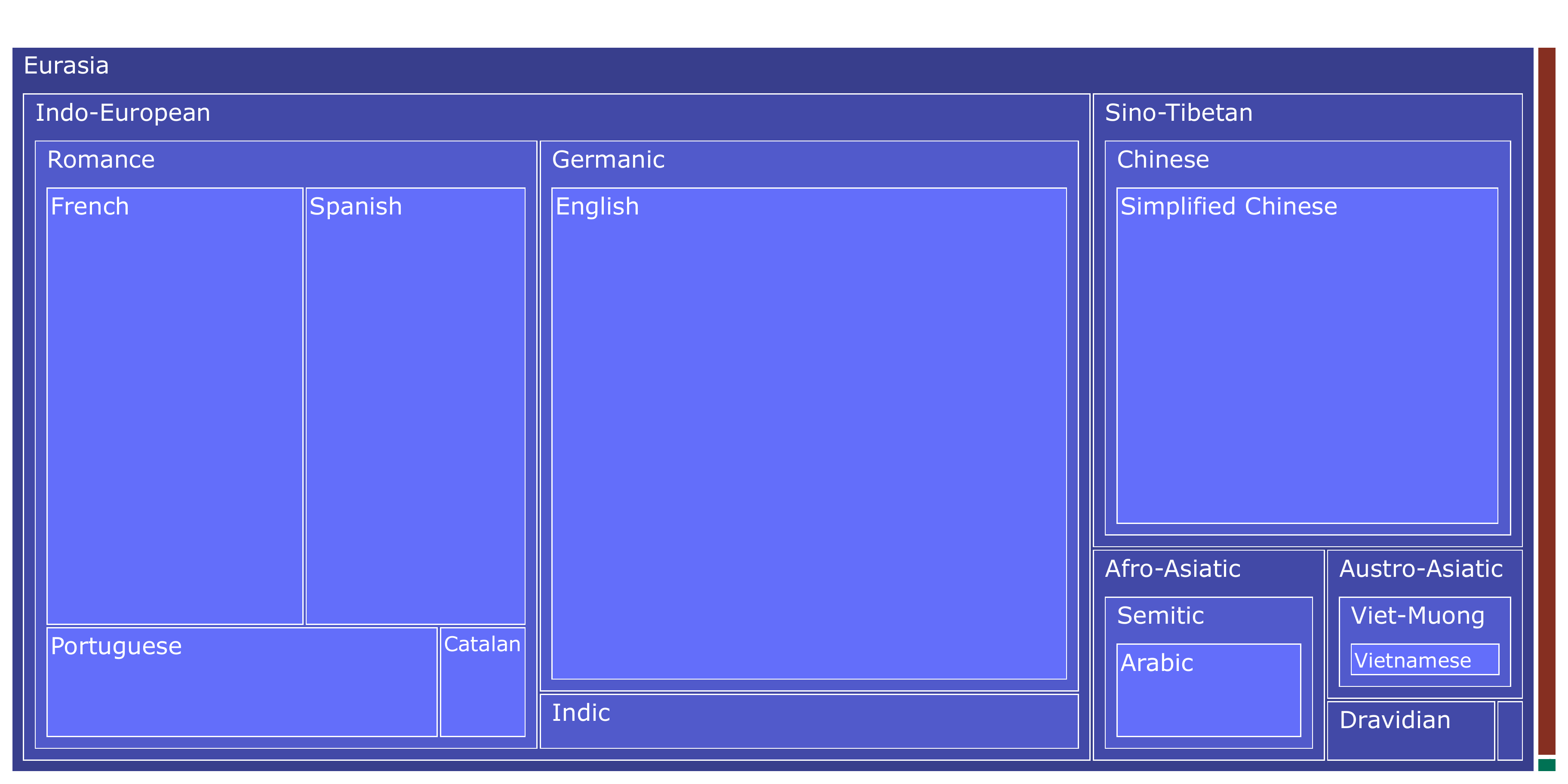}
\hfill
\includegraphics[width=0.21\textwidth]{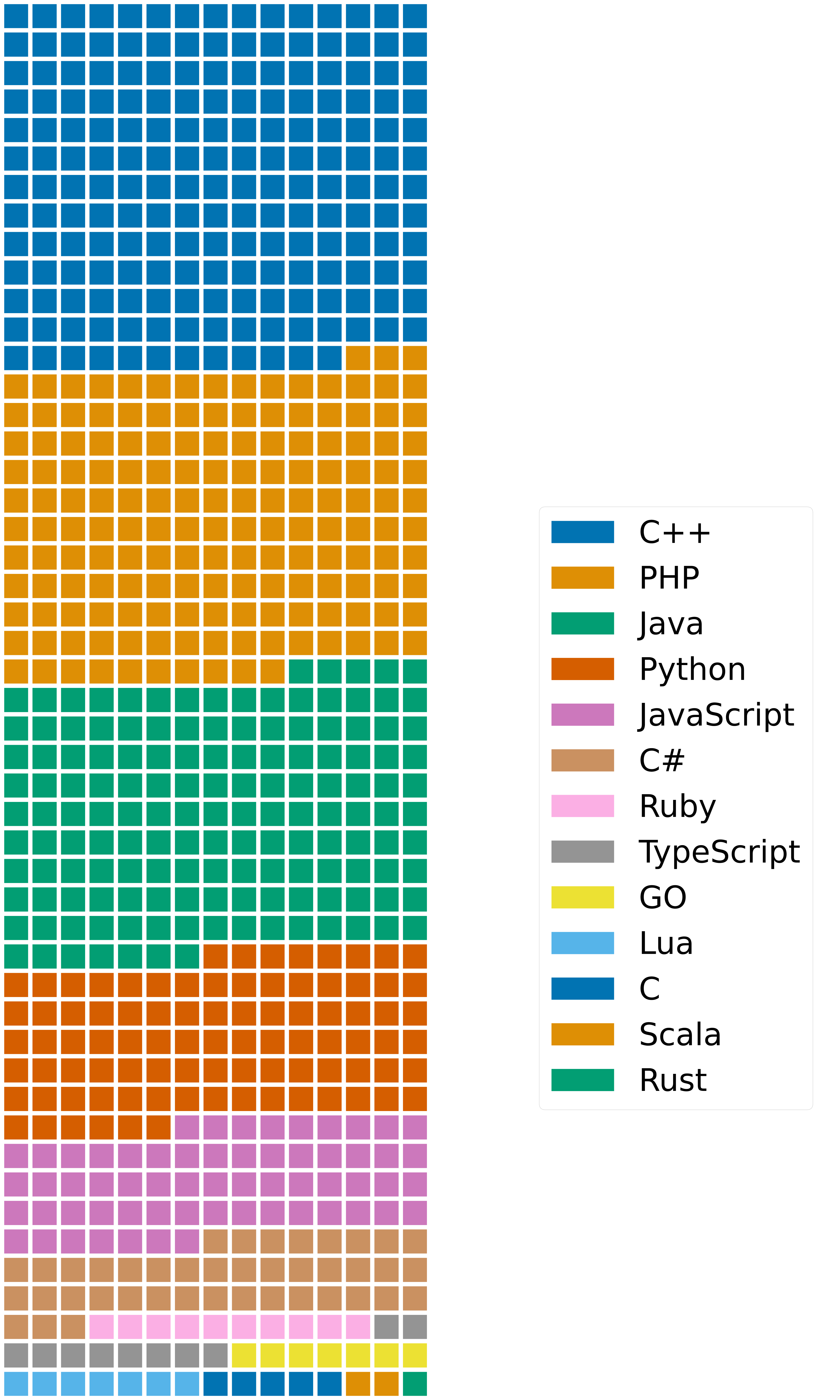}
\caption{
Graphical overview of the ROOTS corpus.
\textbf{Left:} A treemap plot of the language families of all 46 natural languages where surface is proportional to the number of bytes. Indo-European and Sino-Tibetan families overwhelm the plot with a combined total of 1321.89 GB. The thin orange surface represents 18GB of Indonesian data and the green rectangle 0.4GB constituting the Niger-Congo language family subset.
\textbf{Right:} A waffle plot of the distribution of the 13 programming languages by size, where one square represents approximately 200MB.
}
\label{fig:corpus_distribution}
\end{figure}

BLOOM was trained on the ROOTS corpus~\citep{laurencon2022bigscience}, a composite collection of 498~Hugging~Face~datasets~\citep{lhoest-etal-2021-datasets} amounting to 1.61~terabytes of text that span 46 natural languages and 13 programming languages. A high-level overview of this dataset can be seen in~\Cref{fig:corpus_distribution}, while a detailed itemized list of every language along with its linguistic genus, family and macroarea is presented in~\Cref{tab:language_families}.
\begin{table}[]
\centering
\resizebox{\textwidth}{!}{%
\begin{tabular}{lTTlllr}
\toprule
\textbf{Language}            & \textbf{ISO-639-3} & \textbf{catalog-ref} & \textbf{Genus}                   & 
\textbf{Family}         & \textbf{Macroarea} & \textbf{Size in Bytes} \\
\midrule
Akan                & aka       & ak          & Kwa                     & Niger-Congo    & Africa    & 70,1554        \\
Arabic              & arb       & ar          & Semitic                 & Afro-Asiatic   & Eurasia   & 74,854,900,600   \\
Assamese            & asm       & as          & Indic                   & Indo-European  & Eurasia   & 291,522,098     \\
Bambara             & bam       & bm          & Western Mande           & Mande          & Africa    & 391,747        \\
Basque              & eus       & eu          & Basque                  & Basque         & Eurasia   & 2,360,470,848    \\
Bengali             & ben       & bn          & Indic                   & Indo-European  & Eurasia   & 18,606,823,104   \\
Catalan             & cat       & ca          & Romance                 & Indo-European  & Eurasia   & 17,792,493,289   \\
Chichewa           & nya       & ny          & Bantoid                 & Niger-Congo    & Africa    & 1,187,405       \\
chiShona           & sna       & sn          & Bantoid                 & Niger-Congo    & Africa    & 6,638,639       \\
Chitumbuka         & tum       & tum         & Bantoid                 & Niger-Congo    & Africa    & 170,360        \\
English             & eng       & en          & Germanic                & Indo-European  & Eurasia   & 484,953,009,124  \\
Fon                 & fon       & fon         & Kwa                     & Niger-Congo    & Africa    & 2,478,546       \\
French              & fra       & fr          & Romance                 & Indo-European  & Eurasia   & 208,242,620,434  \\
Gujarati            & guj       & gu          & Indic                   & Indo-European  & Eurasia   & 1,199,986,460    \\
Hindi               & hin       & hi          & Indic                   & Indo-European  & Eurasia   & 24,622,119,985   \\
Igbo                & ibo       & ig          & Igboid                  & Niger-Congo    & Africa    & 14078,521      \\
Indonesian          & ind       & id          & Malayo-Sumbawan         & Austronesian   & Papunesia & 19,972,325,222   \\
isiXhosa               & xho       & xh          & Bantoid                 & Niger-Congo    & Africa    & 14,304,074      \\
isiZulu            & zul       & zu          & Bantoid                 & Niger-Congo    & Africa    & 8,511,561       \\
Kannada             & kan       & kn          & Southern Dravidian      & Dravidian      & Eurasia   & 2,098,453,560    \\
Kikuyu              & kik       & ki          & Bantoid                 & Niger-Congo    & Africa    & 359,615        \\
Kinyarwanda         & kin       & rw          & Bantoid                 & Niger-Congo    & Africa    & 40,428,299      \\
Kirundi             & run       & rn          & Bantoid                 & Niger-Congo    & Africa    & 3,272,550       \\
Lingala             & lin       & ln          & Bantoid                 & Niger-Congo    & Africa    & 1,650,804       \\
Luganda             & lug       & lg          & Bantoid                 & Niger-Congo    & Africa    & 4,568,367       \\
Malayalam           & mal       & ml          & Southern Dravidian      & Dravidian      & Eurasia   & 3,662,571,498    \\
Marathi             & mar       & mr          & Indic                   & Indo-European  & Eurasia   & 1,775,483,122    \\
Nepali              & nep       & ne          & Indic                   & Indo-European  & Eurasia   & 2,551,307,393    \\
Northern Sotho      & nso       & nso         & Bantoid                 & Niger-Congo    & Africa    & 1,764,506       \\
Odia                & ori       & or          & Indic                   & Indo-European  & Eurasia   & 1,157,100,133    \\
Portuguese          & por       & pt          & Romance                 & Indo-European  & Eurasia   & 79,277,543,375   \\
Punjabi             & pan       & pa          & Indic                   & Indo-European  & Eurasia   & 1,572,109,752    \\
Sesotho             & sot       & st          & Bantoid                 & Niger-Congo    & Africa    & 751,034        \\
Setswana            & tsn       & tn          & Bantoid                 & Niger-Congo    & Africa    & 1,502,200       \\
Simplified Chinese  &     ---      & zhs         & Chinese                 & Sino-Tibetan   & Eurasia   & 261,019,433,892  \\
Spanish             & spa       & es          & Romance                 & Indo-European  & Eurasia   & 175,098,365,045  \\
Swahili             & swh       & sw          & Bantoid                 & Niger-Congo    & Africa    & 236,482,543     \\
Tamil               & tam       & ta          & Southern Dravidian      & Dravidian      & Eurasia   & 7,989,206,220    \\
Telugu              & tel       & te          & South-Central Dravidian & Dravidian      & Eurasia   & 2993407,159    \\
Traditional Chinese &      ---     & zht         & Chinese                 & Sino-Tibetan   & Eurasia   & 762,489,150     \\
Twi                 & twi       & tw          & Kwa                     & Niger-Congo    & Africa    & 1,265,041       \\
Urdu                & urd       & ur          & Indic                   & Indo-European  & Eurasia   & 2,781,329,959    \\
Vietnamese          & vie       & vi          & Viet-Muong              & Austro-Asiatic & Eurasia   & 43,709,279,959   \\
Wolof               & wol       & wo          & Wolof                   & Niger-Congo    & Africa    & 3,606,973       \\

Xitsonga            & tso       & ts          & Bantoid                 & Niger-Congo    & Africa    & 707,634        \\
Yoruba              & yor       & yo          & Defoid                  & Niger-Congo    & Africa    & 89,695,835      \\
Programming Languages                & ---      & ---        & ---                    & ---           &       & 174,700,245,772 \\
\bottomrule \\
\end{tabular}%
}
\caption{Linguistic makeup of the ROOTS corpus.}
\label{tab:language_families}
\end{table}
Beyond the corpus itself, the process resulted in the development and release of a number of organizational and technical tools, including those illustrated in \Cref{fig:roots-pipelines}.
The rest of this section will contextualize these efforts by providing a brief summary of the steps taken to compile the corpus. For more detailed documentation of the overall dataset curation process and its outcomes, we refer the reader to~\citet{laurencon2022bigscience}.

\paragraph{Motivation} The disconnect between developers and (in)voluntary users of the technology mentioned in \Cref{sec:background} is particularly apparent in the curation of the datasets that have supported recent large-scale machine learning projects, where intentional ``Data work'' is generally under-valued~\citep{sambasivan2021cascades}. In the context of LLMs, this tendency is exemplified by a range of heuristics-based filtering approaches that prioritize getting as much ``high-quality'' data for as little cost as possible over engaging with the needs---and rights---of data subjects, where quality is commonly defined as maximizing performance on downstream tasks while occasionally removing content deemed offensive by the developers.

While these approaches do yield terabytes of data with comparatively little human effort, compounding biases of the source material (such as CommonCrawl dumps) with those of the filtering method often leads to negative outcomes for marginalized populations. In one case, the use of a block list to remove ``pornographic'' text was shown to also suppress LGBTQ+ and African American English~(AAE) text from a corpus~\citep{dodge2021documenting}. In another, using Reddit outgoing links as an indicator of quality for a seed corpus~\citep{radford2019language} leads to trained models that implicitly prioritize US-centric views in their outputs~\citep{johnson2022ghost}. In yet another project, a filtering approach that relied on a machine learning image-text alignment model was shown to exacerbate its biases in the created multimodal dataset~\citep{birhane2021multimodal}. In addition, this \textit{abstractive} approach to data curation leads to corpora that are difficult to meaningfully document and govern after the fact, as the provenance and authorship of individual items is usually lost in the process (although works such as~\citet{gao2020pile} that prioritize compilations of previously documented individual sources over crawled data provide a step towards addressing these issues~\citep{biderman2022datasheet}).

In the context of the BigScience workshop, and in accordance with its Ethical Charter,\footnote{\href{https://bigscience.huggingface.co/blog/bigscience-ethical-charter}{\texttt{bigscience.huggingface.co/blog/bigscience-ethical-charter}}} we aimed to prioritize human involvement, local expertise, and language expertise in our data curation and documentation process, as outlined in the following sections.

\subsubsection{Data Governance}

Large text corpora comprise text about and created by people: the data subjects. Different people and institutions might legally ``own'' that data, making them data rights-holders. As machine learning developers gather and collate that data into ever-larger datasets to support training larger models, it becomes increasingly important to develop new ways of accounting for the interests of all parties involved -- developers, data subjects, and rights-holders alike.

The BigScience effort aimed to address these needs through a multidisciplinary lens combining technical, legal, and sociological expertise.
The group focused on two main interrelated goals at two different time scales: the design of a structure for long-term international data governance that prioritizes the agency of the data rights-holders, and concrete recommendations for handling the data used directly in the BigScience project.
Progress on the first goal is presented in the work of~\citet{jernite2022data}, which further motivates the needs and requirements of data governance, and outlines the structure needed for a network of data custodians, rights-holders, and other parties to appropriately govern shared data.
The interactions between these actors are designed to account for the privacy, intellectual property, and user rights of the data and algorithm subjects in a way that aims to prioritize local knowledge and expression of guiding values.
In particular, this approach relies on structured agreements between data providers and data hosts\footnote{\href{https://hf.co/spaces/bigscience/data_host_provider_agreement}{\texttt{hf.co/spaces/bigscience/data\_host\_provider\_agreement}}} that specify what the data may be used for.

While we were not able to fully establish an international organization in the comparatively short time between the project start and model training, we worked on integrating lessons from this effort (and conversely adapting it to the practical concerns we were experiencing) in the following main ways: (i)~we sought explicit permission to use the data from specific providers \textbf{within the context of BigScience} whenever possible (such as for the AI2\footnote{\href{https://allenai.org/}{\texttt{allenai.org}}}-managed S2ORC corpus of~\citet{lo2020s2orc} or articles from the French newspaper Le Monde\footnote{\href{https://www.lemonde.fr/}{\texttt{lemonde.fr}}}); (ii)~we kept individual sources separate until the final stages of preprocessing to maintain traceability and handle each according to the needs of its specific context; and (iii)~we adopted a composite release approach for the various data sources that make up the overall corpus to foster reproducibility and follow-up research while respecting these source-dependent needs.
Resources to visualize and access the ROOTS corpus can be found on the Hugging Face Hub organization ``BigScience Data''.\footnote{\href{https://hf.co/bigscience-data}{\texttt{hf.co/bigscience-data}}} The organization hosts several demos (or ``Spaces'') that can be used to gain insights into the full corpus, as well as direct access to the 223 (out of 498) components that we are able to distribute taking into account their licensing status, privacy risks, and agreements with their original custodians. Finally, since we understand that future investigation into the BLOOM models may require full access to the entire corpus, we are also inviting researchers with a relevant research project in mind to join ongoing efforts to analyze the data through a sign-up form.\footnote{\href{https://forms.gle/qyYswbEL5kA23Wu99}{{\texttt{forms.gle/qyYswbEL5kA23Wu99}}}}

\subsubsection{Data Sources}
\label{sec:sourcing}

Given a strategy for data governance, the next step was to determine the composition of the training corpus. This stage was driven by several goals, which sometimes had inherent tensions.
Some of those tensions included building a language model that was accessible to as many people as possible around the world while only including languages for which we had enough expertise to curate a dataset of comparable scale (and to a lesser extent composition) to previous efforts while improving the standards of documentation and respect for data and algorithm subject rights.

\paragraph{Language Choices} These considerations led us to an incremental process for choosing which languages were to be included in the corpus. We started with a list of eight of the world's largest languages by number of speakers for which we did active outreach in the early stages of the project to invite fluent speakers to join the data efforts. Then, on the recommendation of language communities~\citep{nekoto2020participatory} we expanded Swahili in the original selection to the category of Niger-Congo languages, and Hindi and Urdu to Indic languages~\citep{kunchukuttan2020indic}. Finally, we proposed that any group of 3 or more participants fluent in an additional language could add it to the supported list if they would commit to selecting sources and guiding processing choices in the language in order to avoid common issues with corpora selected through automatic language identification without specific language expertise~\citep{caswell2022quality}.

\paragraph{Source Selection} The biggest part of the corpus was curated by workshop participants and research collectives who collectively compiled the ``BigScience Catalogue'': a large list of processed and non-processed sources covering a wide range of languages. This took the form of hackathons that were co-organized by communities such as Machine Learning Tokyo, Masakhane, and LatinX in AI~\citep{mcmillan-major2022documenting}. Complementary to those efforts, other working group participants compiled language-specific resources such as the Arabic-focused Masader repository~\citep{alyafeai2021masader,alyafeai2022masader}. A total of 252 sources were identified through this bottom-up approach, with at least 21 sources per language category. Additionally, in order to increase the geographic coverage of some of our Spanish, Chinese, French, and English sources, participants identified locally relevant websites in their language to be added to the corpus via pseudocrawl, a method to obtain those websites from a Common Crawl snapshot.


\paragraph{GitHub Code} The catalogue was further complemented with a dataset of programming languages collected from the GitHub data collection on Google's BigQuery,\footnote{\href{https://cloud.google.com/blog/topics/public-datasets/github-on-bigquery-analyze-all-the-open-source-code}{{\texttt{cloud.google.com/blog/topics/public-datasets/github-on-bigquery-analyze-all-the-open-\\source-code}}}} which was then deduplicated of exact matches. The choice of languages to include mirrored the design choices introduced by~\citet{alphacode} to train the AlphaCode model.

\paragraph{OSCAR} Both in an effort not to diverge from the standard research practice of using the Web as a source of pretraining data \citep{radford2018improving,raffel2020exploring}, and also to satisfy the data volume needs of our compute budget given the size of BLOOM, we further sourced data from OSCAR version 21.09, corresponding to the February~2021 snapshot of the Common Crawl~\citep{OSCAR,AbadjiOrtizSuarezRomaryetal2021}, which ended up constituting 38\% of the corpus. 

\subsubsection{Data Preprocessing}
\label{sec:data-processing}

\begin{figure}[ht!]
\centering
\includegraphics[width=\textwidth]{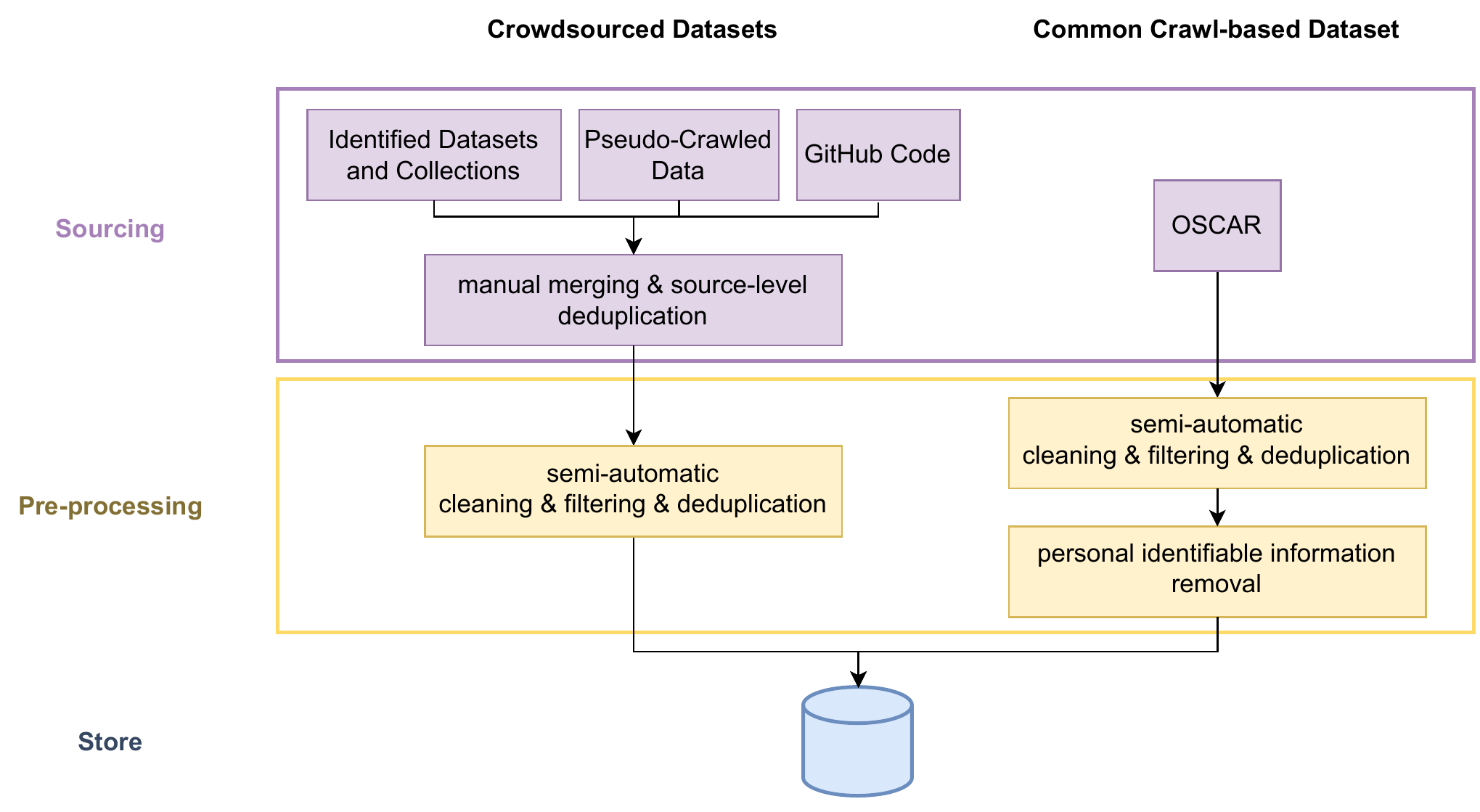}
\caption{Creation Pipeline of the ROOTS Corpus. The purple-colored sourcing stage of the pipeline and the yellow-colored processing stage are described respectively in \Cref{sec:sourcing} and \Cref{sec:data-processing}.}
\label{fig:roots-pipelines}
\end{figure}

After the sources had been identified, data processing involved several steps to handle multiple aspects of data curation.
An overarching view of and processing pipeline to build ROOTS can be seen in \Cref{fig:roots-pipelines}. All tools developed in the process are available on GitHub.\footnote{\href{https://github.com/bigscience-workshop/data-preparation}{\texttt{github.com/bigscience-workshop/data-preparation}}}

\paragraph{Obtaining the Source Data} 
The first step involved obtaining the data for all of the text data sources identified in \Cref{sec:sourcing}, which consisted of a combination of downloading and extracting the text field from a variety of NLP datasets in various formats (including e.g.\ question answering, summarization, or dialogue datasets), scraping and processing large amounts of PDF files from archives~(e.g.\ the French repository of scientific articles\footnote{\href{https://hal.archives-ouvertes.fr/}{\texttt{hal.archives-ouvertes.fr}}}), and extracting and preprocessing text from 192 website entries from the catalogue and another geographically diverse set of 456 websites selected by data working group members. The latter required the development of new tools to extract text from the HTML in the Common Crawl WARC files, which we made available on the main data preparation repository.\footnote{ \href{https://github.com/bigscience-workshop/data-preparation/tree/main/sourcing/cc_pseudo_crawl}{\texttt{github.com/bigscience-workshop/data-preparation/tree/main/sourcing/cc\_pseudo\_crawl}}} We were able to find and extract usable text data from all URLs present in 539 of the websites.

\paragraph{``Quality'' filtering: Text Produced by Humans for Humans} After obtaining the text, we found that most of the sources contained some amount of text that was not natural language, for example preprocessing errors, SEO pages, or spam (including pornographic spam). In order to filter non-natural language, we defined a set of quality indicators, where high-quality text is defined as ``written by humans for humans'', without distinction of content (as we wanted content selection to exclusively be the domain of the more accountable human source selection) or \textit{a priori} judgments of grammaticality. The full list of indicators are described in  \citep{laurencon2022bigscience}. Importantly, the indicators were adapted to the needs of each of the sources in two main ways. First, their parameters such as the thresholds and supporting term lists were selected individually for each language by fluent speakers. Second, we manually went through each individual source to identify which indicators were most likely to identify non-natural language. Both processes were supported by tools to visualize their impact.\footnote{\href{https://hf.co/spaces/huggingface/text-data-filtering}{\texttt{hf.co/spaces/huggingface/text-data-filtering}}}\textsuperscript{,}\footnote{\href{https://hf.co/spaces/bigscience-data/process-pipeline-visualizer}{\texttt{hf.co/spaces/bigscience-data/process-pipeline-visualizer}}}

\paragraph{Deduplication and Privacy Redaction} Finally, we removed near-duplicate documents with two deduplication steps and redacted Personal Identifiable Information (such as social security numbers) that we could identify from the OSCAR version of the corpus---as it was deemed to be the source that presented the highest privacy risks, prompting us to apply regex-based redaction even in cases where the expressions had some false positives.  

\subsubsection{Prompted Datasets}
\label{sec:prompted-data}

\begin{figure}[!ht]
  \centering
  \includegraphics[width=\linewidth]{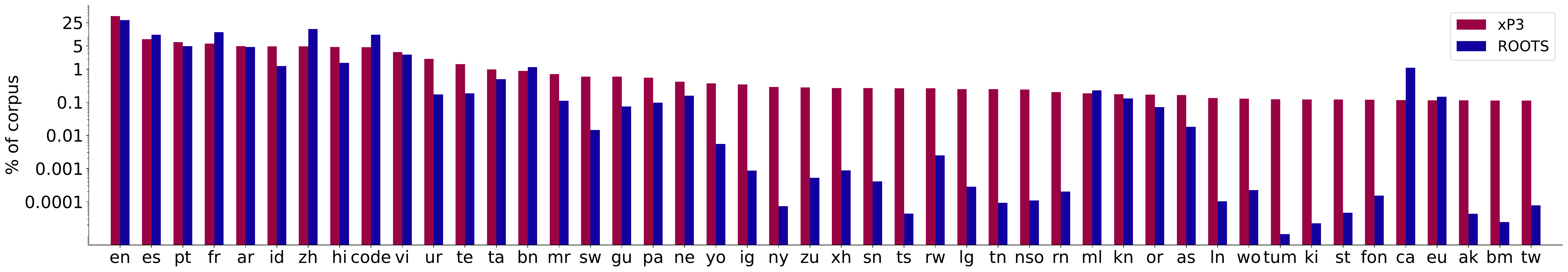}
  \vspace{-4ex}
  \caption{Language distribution of the prompted dataset xP3 closely follows ROOTS.}.
  \label{fig:xp3}
\end{figure}

Multitask prompted finetuning (also referred to as instruction tuning) involves finetuning a pretrained language model on a training mixture composed of a large set of different tasks specified through natural language prompts.
T0~\citep{sanh2022multitask} (developed as part of BigScience) demonstrated that language models finetuned on a multitask mixture of prompted datasets have strong zero-shot task generalization abilities.
Moreover, T0 was shown to outperform language models that are an order of magnitude larger but did not undergo such finetuning. Motivated by these results, we explored using existing natural language datasets to carry out multitask prompted finetuning. 

T0 was trained on a subset of the Public Pool of Prompts (P3), a collection of prompts for various existing and open-source English natural language datasets. This collection of prompts was created through a series of hackathons involving BigScience collaborators and where hackathon participants wrote a total of of 2000+ prompts for 170+ datasets.
Datasets in P3 cover a variety of natural language task including sentiment analysis, question answering, and natural language inference and exclude harmful content or non-natural language such as programming languages. PromptSource~\citep{bach2022promptsource},\footnote{\href{https://github.com/bigscience-workshop/promptsource}{\texttt{github.com/bigscience-workshop/promptsource}}} an open-source toolkit (also developed as part of BigScience) facilitated creating, sharing and using natural language prompts. Full details of the collection process are given in~\citep{sanh2022multitask,bach2022promptsource}.  

After pretraining BLOOM, we applied the same massively multitask finetuning recipe to equip BLOOM with multilingual zero-shot task generalization abilities. We refer to the resulting models as BLOOMZ. To train BLOOMZ, we extended P3 to include new datasets in languages other than English and new tasks, such as translation. This resulted in xP3, a collection of prompts for 83 datasets covering 46 languages and 16 tasks. As highlighted in \Cref{fig:xp3}, xP3 mirrors the language distribution of ROOTS. Tasks in xP3 are both cross-lingual (e.g. translation) and monolingual (e.g. summarization, question answering). We used PromptSource to collect these prompts, adding additional metadata to the prompts, such as input and target languages. To study the importance of multilingual prompts, we also machine-translated English prompts in xP3 to the respective dataset languages to produce a collection called xP3mt. Further details on the prompt collection for xP3 and xP3mt are given in~\cite{muennighoff2022crosslingual}.

\subsection{Model Architecture}
\label{sec:modeling}

This section discusses our design methodology and the architecture of the BLOOM model. 
In-depth studies and experiments can be found in \citet{scao2022what} and \citet{wang2022what}. We first review our design methodology, then motivate our choice of training a causal decoder-only model. Finally, we justify the ways that our model architecture deviates from standard practice.

\subsubsection{Design Methodology}

The design space of possible architectures is immense, making exhaustive exploration impossible.
One option would be to exactly replicate the architecture of an existing large language model.
On the other hand, a great deal of work on improving existing architectures has seen relatively little adoption \citep{narang2021transformer}; adopting some of these recommended practices could yield a significantly better model.
We take a middle ground and focus on model families that have been shown to scale well, and that have reasonable support in publicly available tools and codebases. We ablate components and hyperparameters of the models, seeking to make the best use of our final compute budget.

\paragraph{Experimental Design for Ablations}

One of the main draws of LLMs has been their ability to perform tasks in a ``zero/few-shot'' way: large enough models can perform novel tasks simply from in-context instructions and examples \citep{radford2019language}, without dedicated training on supervised samples. Accordingly, and because finetuning a 100B+ model is unwieldy, we focused our evaluation of architectural decisions on zero-shot generalization, and do not consider transfer learning. Specifically, we measured zero-shot performance on diverse aggregates of tasks: 29 tasks from the EleutherAI Language Model Evaluation Harness (EAI-Eval, \cite{eval-harness}), and 9 tasks from the evaluation set of T0 (T0-Eval, \cite{sanh2022multitask}). There is significant overlap between the two: only one task from T0-Eval (StoryCloze) is not in EAI-Eval, although all prompts between the two are different. See \citet{scao2022what} for a detailed list of tasks and baselines.
We also note that our tasks aggregates share 17 of the 31 tasks of the evaluation of GPT-3 \citep{brown2020language}.

We conducted our ablation experiments using smaller models. We used the 6.7B parameter scale for the pretraining objective ablations~\citep{wang2022what} and the 1.3B scale for the rest including position embeddings, activations, and layer normalization~\citep{scao2022what}.
Recently, \citet{dettmers2022llm} identified a phase transition for models larger than 6.7B, in which the emergence of ``outliers features'' is observed. This questions whether results obtained at the 1.3B scale should be assumed to extrapolate to our final model size.

\paragraph{Out-of-scope Architectures}
We did not consider mixture-of-experts (MoE) \citep{shazeer2017}, due to a lack of widely used GPU-based codebases suitable for training them at scale. Similarly, we also did not consider state-space models \citep{gu2020hippo}. At the time of the design of BLOOM, they consistently underperformed in natural language tasks \citep{gu2021efficiently}. Both of these approaches are promising, and have now demonstrated competitive results--at large scales for MoE~\citep{fedus2022switch, srivastava2022beyond}, and at smaller scale for state-space models with H3~\citep{fu2023hungry}. 

\subsubsection{Architecture and Pretraining Objective}

Although most modern language models are based on the Transformer architecture, there are significant deviations between architectural implementations. Notably, while the original Transformer is based on an encoder-decoder architecture, many popular models have opted for encoder-only (e.g. BERT, \citep{devlin2019bert}) or decoder-only (e.g. GPT, \citep{radford2018improving}) approaches. Currently, all state-of-the-art language models over 100 billion parameters are causal decoder-only models \citep{brown2020language, rae2021scaling, chowdhery2022palm}. 
This is in opposition to the findings of \citet{raffel2020exploring}, in which encoder-decoder models significantly outperform decoder-only models for transfer learning.

Prior to our work, the literature was lacking a systematic evaluation of the zero-shot generalization capabilities of different architectures and pretraining objectives. 
We explored this question in~\citet{wang2022what} where we evaluated encoder-decoder and
decoder-only architectures and their interactions with causal, prefix, and masked language modeling pretraining objectives. 
Our results show that immediately after pretraining, causal decoder-only models performed best -- validating the choice of state-of-the-art LLMs. Furthermore, they can be more efficiently adapted after pretraining to a non-causal architecture and objective--an approach which has been further explored and confirmed by~\citet{tay2022transcending}.

\subsubsection{Modeling Details}

Beyond choosing an architecture and pretraining objective, a number of changes to the original Transformer architecture have been proposed. For example, alternative positional embedding schemes~\citep{su2021roformer, press2021train} or novel activation functions~\citep{shazeer2020glu}. We thus performed a series of experiments to evaluate the benefit of each of these modifications for a causal decoder-only model in~\citet{scao2022what}. We adopted two architectural deviations in BLOOM:

\paragraph{ALiBi Positional Embeddings} Instead of adding positional information to the embedding layer, ALiBi directly attenuates the attention scores based on how far away the keys and queries are \citep{press2021train}. Although ALiBi was initially motivated by its ability to extrapolate to longer sequences, we found it also led to smoother training and better downstream performance even at the original sequence length -- outperforming both learned \citep{vaswani2017attention} and rotary \citep{su2021roformer} embeddings.

\paragraph{Embedding LayerNorm} In preliminary experiments training a 104B parameters model, we experimented with an additional layer normalization immediately after the embedding layer -- as recommended by the \texttt{bitsandbytes}\footnote{\href{https://github.com/TimDettmers/bitsandbytes}{\texttt{github.com/TimDettmers/bitsandbytes}}} library~\citep{dettmers2022llm} with its \texttt{StableEmbedding} layer. We found this significantly improved training stability. Even though we also found it penalizes zero-shot generalization in~\citet{scao2022what}, we train BLOOM with an additional layer normalization after the first embedding layer to avoid training instabilities.
Note the preliminary 104B experiments were conducted in \texttt{float16}, while the final training was in \texttt{bfloat16}. Since then, \texttt{float16} has been attributed as being responsible for many of the observed instabilities in training LLMs \citep{zhang2022opt, zeng2022glm}. It is possible that \texttt{bfloat16} alleviates the need for the embedding LayerNorm.

\begin{figure}[!ht]
  \centering
  \includegraphics[width=0.9\linewidth]{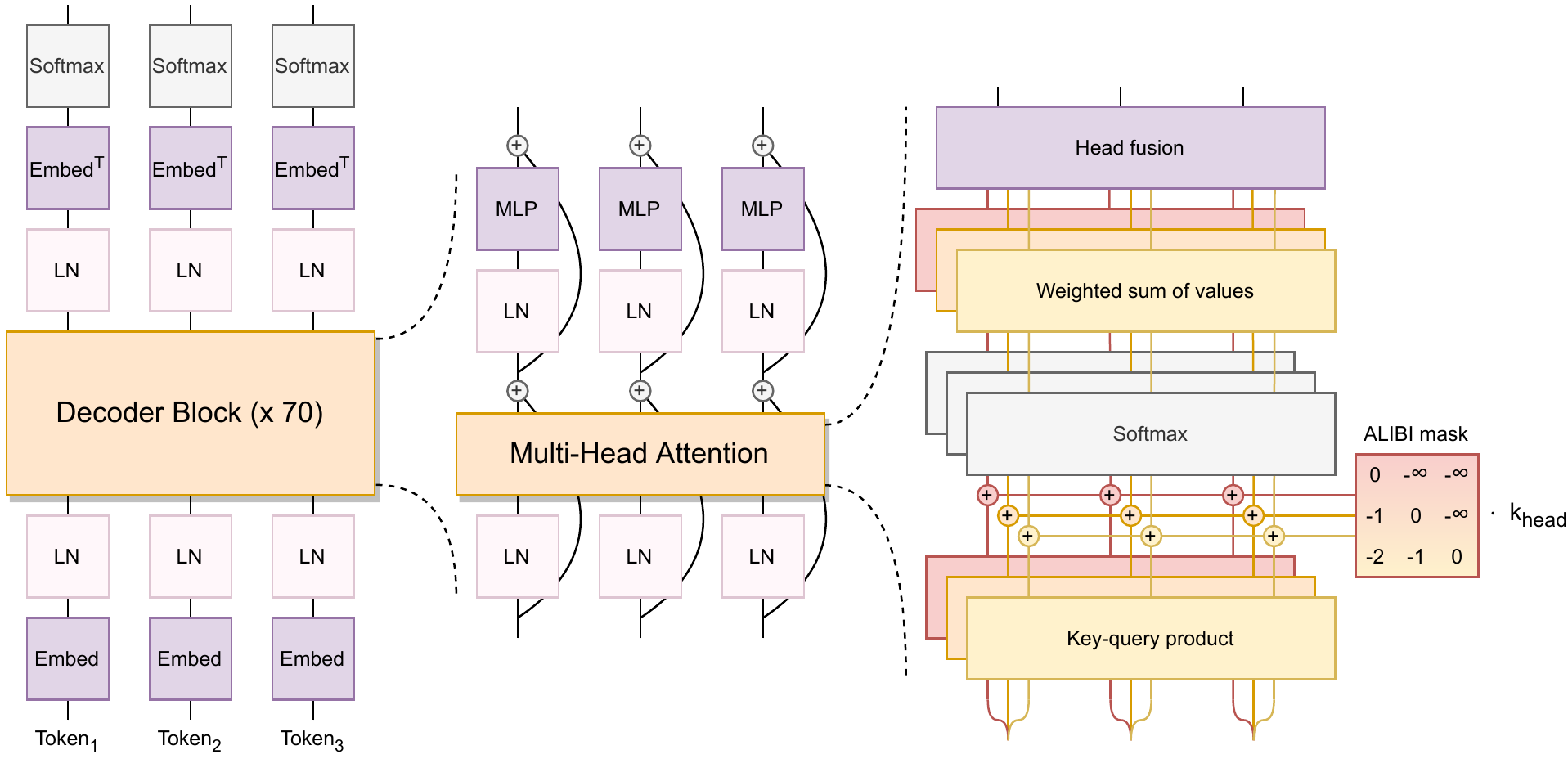}
  \caption{The BLOOM architecture. The $k_{head}$ slope parameters for ALIBI are taken as $2^\frac{-8i}{n}$ with $n$ the number of heads and $i \in {1, 2, ..., n}$.}
  \label{fig:architecture_diagram}
\end{figure}

We represent the full architecture of BLOOM in figure \ref{fig:architecture_diagram} for reference.

\subsection{Tokenization}
\label{sec:tokenization}

The design decisions when training a tokenizer are often neglected in favour of ``default'' settings \citep{mielke2021between}. For instance, OPT~\citep{zhang2022opt} and GPT-3~\citep{brown2020language} both use GPT-2's tokenizer, trained for English. This can be justified by the fact that evaluating the impact of a particular choice on the downstream performance of the model is constrained by the large computational costs of training. However, the diverse nature of BLOOM's training data requires careful design choices to ensure that the tokenizer encodes sentences in a lossless manner.

\paragraph{Validation}
We use the fertility~\citep{acs2019exploring} of our tokenizer compared to existing monolingual tokenizers as a metric for sanity checks. Fertility is defined as the number of subwords created per word or per dataset by the tokenizer, which we measured using subsets of Universal Dependencies 2.9~\citep{nivre-etal-2017-universal} and OSCAR~\citep{OSCAR} in the languages of interest. A very high fertility on a language compared to a monolingual tokenizer may indicate a degradation on the downstream multilingual performance of the model~\citep{rust-etal-2021-good}. Our goal was to not degrade the fertility on each language by more than 10 percentage points when comparing our multilingual tokenizer with monolingual tokenizers in corresponding languages. For all experiments, the Hugging~Face~Tokenizers library~\citep{hftokenizers2019} was used to design and train the tested tokenizers.

\begin{table}[ht]
\begin{center}
\begin{tabular}{lllllll}
\toprule
\textbf{Tokenizer} & \textbf{fr} & \textbf{en} & \textbf{es} & \textbf{zh} & \textbf{hi} & \textbf{ar} \\ 
\midrule
Monolingual   & 1.30 & 1.15 & 1.12 & 1.50 & 1.07 & 1.16 \\ 
BLOOM         & 1.17 {\scriptsize(-11\%)} & 1.15 {\scriptsize(+0\%)} & 1.16 {\scriptsize(+3\%)} & 1.58 {\scriptsize(+5\%)} & 1.18 {\scriptsize(+9\%)} & 1.34 {\scriptsize(+13\%)} \\ 
\bottomrule
\end{tabular}
\caption{Fertilities obtained on Universal Dependencies treebanks on languages with existing monolingual tokenizers. The monolingual tokenizers we used were the ones from CamemBERT~\citep{martin2020camembert}, GPT-2~\citep{radford2019language}, \texttt{DeepESP/gpt2-spanish}, \texttt{bert-base-chinese}, \texttt{monsoon-nlp/hindi-bert} and Arabic BERT~\citep{safaya2020kuisail}, all available on the HuggingFace Hub.}
\label{table:ud_fertilities}
\end{center}
\end{table}

\paragraph{Tokenizer Training Data} We initially used a non-deduplicated subset of ROOTS. However, a qualitative study on the vocabulary of the tokenizer revealed issues in its training data. For instance, in earlier versions of the tokenizer, we found entire URLs stored as tokens caused by several documents containing a high number of duplicates. These issues motivated us to remove duplicated lines in the tokenizer training training data. We then applied the same sampling ratios per language as for the training data.

\paragraph{Vocabulary Size} 

%
A large vocabulary size reduces the risk of over-segmenting some sentences, especially for low-resource languages. We conducted validation experiments using 150k and 250k vocabulary sizes to make comparisons with existing multilingual modeling literature easier~\citep{conneau2020unsupervised,xue2021mt5}. We ultimately settled for a vocabulary of 250k tokens to reach our initial fertility objective compared to monolingual tokenizers. Since the vocabulary size determines the embedding matrix size, it also had to be divisible by 128 for GPU efficiency reasons and by 4 to be able to use Tensor Parallelism. We used a final size of 250,680~vocabulary items with 200 tokens reserved for possible future applications such as removing private information using placeholder tokens.

\paragraph{Byte-level BPE}
The tokenizer is a learned subword tokenizer trained using the Byte Pair Encoding~(BPE) algorithm introduced by \citet{gagebpe}. In order not to lose information during tokenization, the tokenizer creates merges starting from bytes as the smallest units instead of characters \citep{radford2019language}. This way, tokenization never results in unknown tokens because all 256 bytes can be contained in the vocabulary of the tokenizer. In addition, Byte-level~BPE maximizes vocabulary sharing between languages~\citep{wang2020neural}.

\paragraph{Normalization}
 Upstream of the BPE tokenization algorithm, no normalization of the text was performed in order to have the most general model possible. In all cases, we observed that adding unicode normalization such as NFKC did not reduce the fertility by more than 0.8\% on all the languages considered but came at the cost of making the model less general; for example, causing $2^{2}$ and $2 2$ to be encoded in the same way.

\paragraph{Pre-tokenizer}
Our pre-tokenization has two goals: producing a first division of the text (usually using whitespaces and punctuation) and restricting the maximum length of sequences of tokens produced by the BPE algorithm. The pre-tokenization rule used was the following regex: ``\tokenizerregex'' 
 \footnote{\href{https://github.com/bigscience-workshop/bs-tokenizers/blob/c510647dda77a37d83c756bc5a5ce85048ab66c0/tokenizers/train_tokenizer_v3_on_subset.py\#L124}{\texttt{github.com/bigscience-workshop/bs-tokenizers}}}
which splits words apart while preserving all the characters and in particular the sequences of spaces and line breaks that are crucial for programming languages. We do not use English-centric splits common in other tokenizers (e.g. splitting around \texttt{'nt} or \texttt{'ll}). We also didn't use splits on numbers and digits, which caused issues in Arabic and code.




\subsection{Engineering}
\label{sec:engineering}

\subsubsection{Hardware}

The model was trained on Jean Zay,\footnote{\href{http://www.idris.fr/eng/jean-zay/jean-zay-presentation-eng.html}{\texttt{idris.fr/eng/jean-zay/jean-zay-presentation-eng.html}}} a French government-funded supercomputer owned by GENCI and operated at IDRIS, the national computing center for the French National Center for Scientific Research (CNRS).
Training BLOOM took about 3.5 months to complete and consumed 1,082,990 compute hours. Training was conducted on 48 nodes, each having 8 NVIDIA A100 80GB GPUs (a total of 384 GPUs); due to possible hardware failures during training, we also maintained a reserve of 4 spare nodes. The nodes were equipped with 2x AMD EPYC 7543 32-Core CPUs and 512 GB of RAM, while the storage was handled by mix of full flash and hard disk drives using a SpectrumScale (GPFS) parallel file system shared between all nodes and users of the supercomputer. 4 NVLink GPU-to-GPU interconnects per node enabled intra-node communications while 4 Omni-Path 100 Gbps links per node, arranged in an enhanced hypercube 8D global topology, were used for inter-node communications.

\subsubsection{Framework}
BLOOM was trained using Megatron-DeepSpeed\footnote{\href{https://github.com/bigscience-workshop/Megatron-DeepSpeed}{\texttt{github.com/bigscience-workshop/Megatron-DeepSpeed}}}~\citep{smith2022using}, a framework for large-scale distributed training. It consists of two parts: Megatron-LM\footnote{\href{https://github.com/NVIDIA/Megatron-LM}{\texttt{github.com/NVIDIA/Megatron-LM}}}~\citep{shoeybi2019megatron} provides the Transformer implementation, tensor parallelism, and data loading primitives, whereas DeepSpeed\footnote{\href{https://github.com/microsoft/DeepSpeed}{\texttt{github.com/microsoft/DeepSpeed}}}~\citep{rasley2020deepspeed} provides the ZeRO optimizer, model pipelining, and general distributed training components. This framework allows us to train efficiently with \textit{3D parallelism} (\citealp{narayanan2021efficient}, shown in~\autoref{fig:3d-parallelism}), a fusion of three complementary approaches to distributed training. These approaches are described below:

\begin{figure}[!ht]
  \centering
  \includegraphics[width=\linewidth]{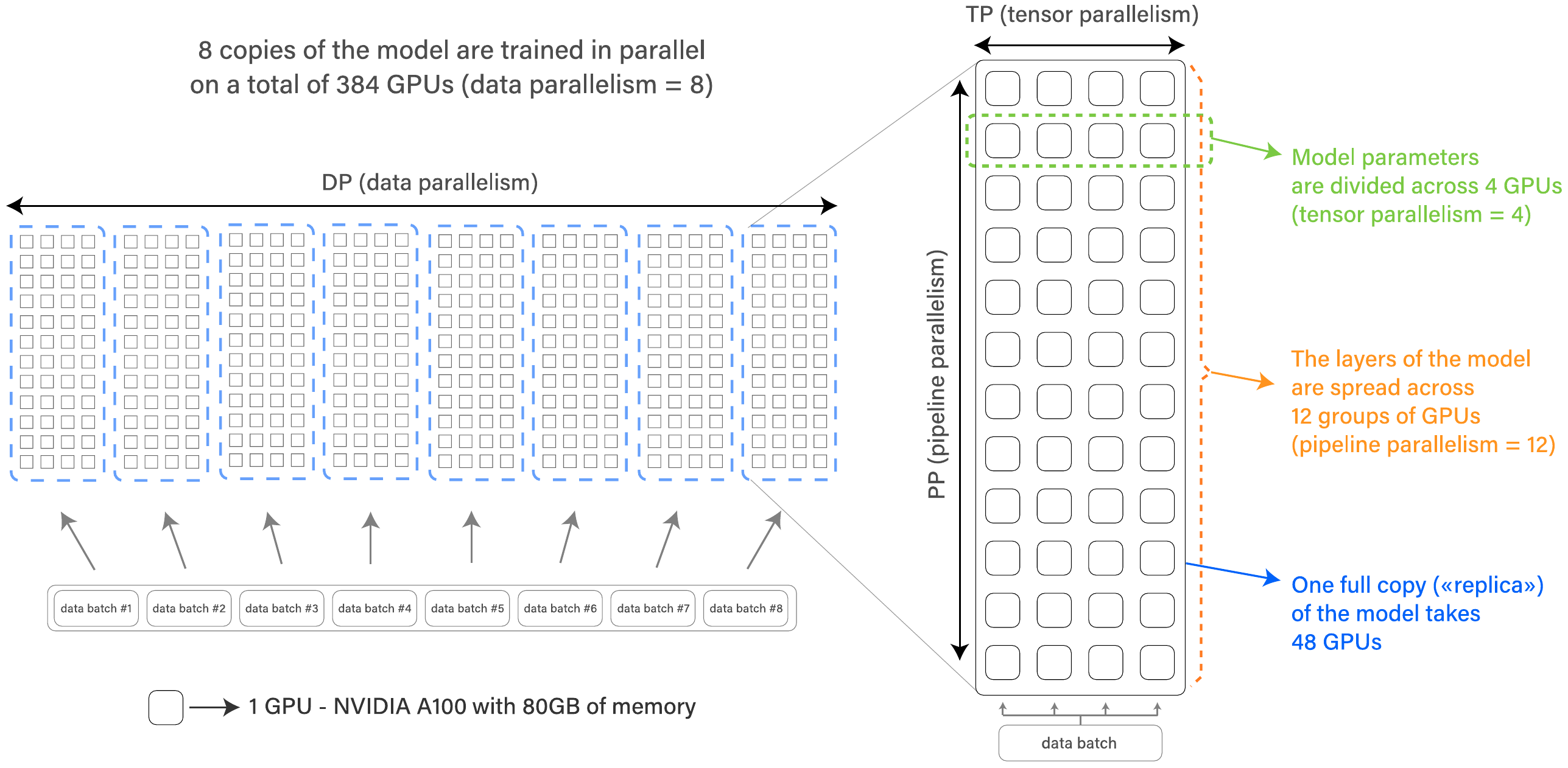}
  \caption{DP+PP+TP combination leads to 3D parallelism.}
  \label{fig:3d-parallelism}
\end{figure}

\begin{description}
\item[Data parallelism (DP)] replicates the model multiple times, with each replica placed on a different device and fed a slice of the data. The processing is done in parallel and all model replicas are synchronized at the end of each training step.
\item[Tensor parallelism (TP)] partitions individual layers of the model across multiple devices. This way, instead of having the whole activation or gradient tensor reside on a single GPU, we place shards of this tensor on separate GPUs. This technique is sometimes called horizontal parallelism or intra-layer model parallelism.
\item[Pipeline parallelism (PP)] splits up the model's layers across multiple GPUs, so that only a fraction of the layers of the model are placed on each GPU. This is sometimes called vertical parallelism.
\end{description}
Finally, the Zero Redundancy Optimizer (ZeRO; \citealp{rajbhandari2020zero}) allows different processes to only hold a fraction of data (parameters, gradients, and optimizer states) required for a training step. We used ZeRO stage 1, meaning that only the optimizer states are sharded in this manner.

The four components described above are combined together to allow scaling to hundreds of GPUs with extremely high GPU utilization. We were able to achieve 156 TFLOPs in our fastest configuration with A100 GPUs, attaining our objective of half of the theoretical peak performance of 312 TFLOPs (in \texttt{float32} or \texttt{bfloat16}).

\subsubsection{Floating Point Format}

In earlier experiments with 104B-parameter models on NVIDIA V100 GPUs, we observed numerical instabilities that caused irreversible training divergences.
We hypothesize that these instabilities stem from our initial use of IEEE \texttt{float16} --- a 16-bit floating point format with a very limited dynamic range that can cause overflows.
The NVIDIA A100 GPUs that we ultimately had access to support the \texttt{bfloat16} format~\citep{bf16blog,kalamkar2019study}, which has the same dynamic range as \texttt{float32}.
On the other hand, \texttt{bfloat16} still has much lower precision, which motivated our use of mixed-precision training~\citep{micikevicius2018mixed}. This technique performs certain precision-sensitive operations such as gradient accumulation and softmax in \texttt{float32} precision and the rest of operations in lower precision, allowing us to achieve a balance of high performance and training stability.
Ultimately, we performed final training in \texttt{bfloat16} mixed precision, which proved to solve the instability problem (in line with previous observation by \citealp{smith2022using}).

\subsubsection{Fused CUDA Kernels}

In general, GPUs cannot retrieve data to perform computations on and perform these computations at the same time. Moreover, the compute performance of modern GPUs is much higher than the speed of memory transfer required for every operation (often called \textit{a kernel} in GPU programming).
Kernel fusion~\citep{kernel_fusion} is an approach for optimizing GPU-based computations by performing several consecutive operations in only one kernel call. This approach offers a way to minimize data transfers: intermediary results stay in the GPU register instead of being copied into VRAM, saving overhead. 

We used several custom fused CUDA kernels provided by Megatron-LM. First, we used an optimized kernel to perform LayerNorm, as well as kernels to fuse various combinations of the scaling, masking, and softmax operations. The addition of a bias term is also fused with the GeLU activation using the JIT functionality of PyTorch. As an example consequence of the use of fused kernels, adding the bias term in the GeLU operation adds no additional time, as the operation is memory-bound: the additional computation is negligible compared to data transfers between GPU VRAM and registers, so fusing both operations essentially halves their runtime.

\subsubsection{Additional Challenges}

Scaling to 384 GPUs required two final changes: disabling asynchronous CUDA kernel launches (for ease of debugging and to prevent deadlocks) and splitting parameter groups into smaller subgroups (to avoid excessive CPU memory allocations). 

During training, we faced issues with hardware failures: on average, 1--2 GPU failures occurred each week. As backup nodes were available and automatically used, and checkpoints were saved every three hours, this did not affect training throughput significantly. A PyTorch deadlock bug in the data loader and disk space issues led to 5--10h downtimes. Given the relative sparsity of engineering issues, and since there was only one loss spike, which the model swiftly recovered from, human intervention was less necessary than in comparable projects~\citep{zhang2022opt}.
Full details of our experience with training BLOOM and a detailed report of all issues we faced are publicly available.\footnote{\href{https://github.com/bigscience-workshop/bigscience/blob/master/train/tr11-176B-ml/chronicles.md}{\texttt{github.com/bigscience-workshop/bigscience/blob/master/train/tr11-176B-ml/chronicles.md}}}


\subsection{Training}
\label{sec:training}

\begin{table}[ht]
\begin{center}
\resizebox{\textwidth}{!}{
    \begin{tabular}{l|ccccc|c}
    Hyperparameter ($\downarrow$) & BLOOM-560M & BLOOM-1.1B & BLOOM-1.7B & BLOOM-3B & BLOOM-7.1B & BLOOM \\ 
    \midrule
    \multicolumn{7}{c}{\emph{Architecture hyperparameters}} \\
    \midrule
    Parameters & 559M & 1,065M & 1,722M & 3,003M & 7,069M & 176,247M \\ 
    Precision & \multicolumn{5}{c|}{\texttt{float16}} & \texttt{bfloat16} \\ 
    Layers   & 24 & 24 & 24 & 30 & 30 & 70 \\ 
    Hidden dim. & 1024 & 1536 & 2048 & 2560 & 4096 & 14336 \\ 
    Attention heads & 16 & 16 & 16 & 32 & 32 & 112 \\
    Vocab size & \multicolumn{5}{c|}{250,680} & 250,680 \\ 
    Sequence length & \multicolumn{5}{c|}{2048} & 2048 \\
    Activation & \multicolumn{5}{c|}{\texttt{GELU}} & \texttt{GELU} \\
    Position emb. & \multicolumn{5}{c|}{\texttt{Alibi}} & \texttt{Alibi} \\ 
    Tied emb. & \multicolumn{5}{c|}{\texttt{True}} & \texttt{True} \\ 
    \midrule
    \multicolumn{7}{c}{\emph{Pretraining hyperparameters}} \\
    \midrule
    Global Batch Size  & 256 & 256 & 512 & 512 & 512 & 2048 \\ 
    Learning rate & 3.0e-4 & 2.5e-4 & 2e-4 & 1.6e-4 & 1.2e-4 &  6e-5  \\ 
    Total tokens & \multicolumn{5}{c|}{341B} & 366B \\ 
    Warmup tokens & \multicolumn{5}{c|}{375M} & 375M \\ 
    Decay tokens & \multicolumn{5}{c|}{410B} & 410B \\ 
    Decay style & \multicolumn{5}{c|}{\texttt{cosine}} & \texttt{cosine} \\ 
    Min. learning rate & \multicolumn{5}{c|}{1e-5} & 6e-6 \\ 
    Adam $(\beta_1, \beta_2)$ & \multicolumn{5}{c|}{(0.9, 0.95)} & (0.9, 0.95) \\     
    Weight decay & \multicolumn{5}{c|}{1e-1} & 1e-1 \\ 
    Gradient clipping & \multicolumn{5}{c|}{1.0} & 1.0 \\
    \midrule
    \multicolumn{7}{c}{\emph{Multitask finetuning hyperparameters}} \\
    \midrule
    Global Batch Size  & 1024 & 1024 & 2048 & 2048 & 2048 & 2048 \\ 
    Learning rate & 2.0e-5 & 2.0e-5 & 2.0e-5 & 2.0e-5 & 2.0e-5 &  2.0e-5  \\ 
    Total tokens & \multicolumn{5}{c|}{13B} & 13B \\ 
    Warmup tokens & \multicolumn{5}{c|}{0} & 0 \\ 
    Decay style & \multicolumn{5}{c|}{\texttt{constant}} & \texttt{constant}
    \\     
    Weight decay & \multicolumn{5}{c|}{1e-4} & 1e-4 \\ 
    \end{tabular}
}
\caption{BLOOM \& BLOOMZ Training Hyperparameters.}
\label{table:hp}
\end{center}
\end{table}

\paragraph{Pretrained Models} We train six size variants of BLOOM with respective hyperparameters detailed in Table \ref{table:hp}. Architecture and training hyperparameters come from our experimental results \citep{scao2022what} and prior work on training large language models \citep{brown2020language,kaplan2020scaling}. Model depth and width for the non-176B models roughly follow previous literature \citep{brown2020language, zhang2022opt}, deviating for 3B and 7.1B in order only to fit the models more easily on our training setup. Embedding parameter sizes are larger for BLOOM owing to the larger multilingual vocabulary, but scaling literature discounts embedding operations \citep{kaplan2020scaling}. During the development process at the 104B parameters scale, we experimented with different values of Adam $\beta$ parameters, weight decay and gradient clipping to target stability, but did not find it helpful. For all models, we use a cosine learning rate decay schedule \citep{loschchilovcosinedecay} over 410B tokens, taken as an upper bound for the length of training if compute permitted, and warmup for 375M tokens. We use weight decay, gradient clipping, and no dropout. The ROOTS dataset contains around 341 billion tokens of text, so we aimed to train all models for the equivalent amount of tokens. However, in light of revised scaling laws published during training \citep{hoffmann2022training}, we decided to train the large models for an additional 25 billion tokens on repeated data. As warmup tokens + decay tokens were larger than the total number of tokens, the end of learning rate decay was never reached.

\paragraph{Multitask Finetuning} Finetuned BLOOMZ models~\citep{muennighoff2022crosslingual} maintain the same architecture hyperparameters as BLOOM models. The finetuning hyperparameters are loosely based on T0 \citep{sanh2022multitask} and FLAN \citep{wei2021finetuned}. Learning rates are determined by doubling the minimum learning rate of the respective pretrained model and then rounding. Global batch sizes are multiplied by four for small variants to increase throughput. While the models are finetuned for 13 billion tokens, the best checkpoint is chosen according to a separate validation set. We found performance to plateau after 1 -- 6 billion tokens of finetuning.

\paragraph{Contrastive Finetuning} We also perform contrastive finetuning of the 1.3 and 7.1 billion parameter BLOOM models using the SGPT Bi-Encoder recipe~\citep{muennighoff2022sgpt} to train models that produce high-quality text embeddings. We created SGPT-BLOOM-7.1B-msmarco\footnote{\href{https://hf.co/bigscience/sgpt-bloom-7b1-msmarco}{\texttt{hf.co/bigscience/sgpt-bloom-7b1-msmarco}}} geared towards multilingual information retrieval and SGPT-BLOOM-1.7B-nli\footnote{\href{https://hf.co/bigscience-data/sgpt-bloom-1b7-nli}{\texttt{hf.co/bigscience-data/sgpt-bloom-1b7-nli}}} for multilingual semantic textual similarity (STS). However, recent benchmarking has found these models to also generalize to various other embedding tasks, such as bitext mining, reranking or feature extraction for downstream classification~\citep{muennighoff2022mteb}.

\subsubsection{Carbon Footprint}

While most attempts to estimate the carbon footprint of language models have shed light on the emissions produced due to energy consumed during model training (e.g.~\citealp{patterson2021carbon, strubell2019energy}), other sources of emissions are also important to consider. In our efforts to estimate the carbon emissions of BLOOM, we were inspired by the Life Cycle Assessment (LCA) approach~\citep{klopffer1997life} and aimed to consider aspects such as the emissions of equipment manufacturing, intermediate model training, and deployment. According to our estimates, the carbon emissions from BLOOM training add up to approximately 81 tons of \carboneq, of which 14\% were generated by the equipment manufacturing process (11 tons), 30\% by the energy consumed during training (25 tons) and 55\% by idle consumption of the equipment and computing cluster used for training (45 tons). 

\begin{table}[ht]
\begin{center}\small
\begin{tabular}{lrrr}
\toprule
{\textbf{\begin{tabular}[c]{@{}c@{}} Model \\ name \end{tabular}}}  &  \multicolumn{1}{l}{\textbf{\begin{tabular}[c]{@{}c@{}}Number of \\ parameters \end{tabular}}}  & \multicolumn{1}{c}{\textbf{\begin{tabular}[c]{@{}c@{}} Power \\ consumption \end{tabular}}} & \multicolumn{1}{c} {\textbf{\begin{tabular}[c]{@{}c@{}} \carboneq~\\ emissions  \end{tabular}}} \\ 
\midrule
GPT-3 & 175B  & 1,287 MWh & \textit{502 tons} \\ 
Gopher & 280B  & \textit{1,066 MWh} & \textit{352 tons} \\
OPT & 175B & \textit{324 MWh} &  70 tons \\ 
BLOOM & 176B  & 433 MWh & 25 tons   \\ 
\bottomrule
\end{tabular}
\caption{Comparison of carbon emissions between BLOOM and similar LLMs. Numbers in \textit{italics} have been inferred based on data provided in the papers describing the models.}
\label{table:co2}
\end{center}
\end{table}

Comparing the carbon emissions of BLOOM training to other similar models (see Table~\ref{table:co2}), reveals that while the energy consumption of BLOOM is slightly higher than OPT~\citep{zhang2022opt} (433 Mwh compared to OPT's 324 MWh), its emissions are approximately 2/3 less (25 tons versus 70 tons). This is thanks to the low carbon intensity of the energy grid used for training BLOOM, which emits 57~\carbonintensity, compared to 231~\carbonintensity~ for the grid used for OPT training. Specifically, France's national energy grid (which is used by Jean Zay) is largely powered by nuclear energy, which is low-carbon compared to grids powered by energy sources such as coal and natural gas. While the sustainability of nuclear energy is debated, it is one of the least carbon-intensive sources of energy that is currently available. Both BLOOM and OPT incurred significantly less carbon emissions than GPT-3 (as reported by~\citep{patterson2021carbon}), which can be attributed to several factors including more efficient hardware as well as less carbon-intensive energy sources.

We also pursued further exploration of the carbon footprint of (1) the computation carried out on Jean Zay within the scope of the Big Science workshop, and (2) running the BLOOM model API in real time. In terms of the footprint of the totality of the computation, we estimate that the final BLOOM training represents approximately 37\% of the overall emissions, with other processes such as intermediate training runs and model evaluation adding up to the other 63\%. This is slightly less than the estimate made by the authors of the OPT paper, who stated that the total carbon footprint of their model is roughly 2 times higher due to experimentation, baselines and ablation~\citep{zhang2022opt}. Our ongoing exploration of the carbon emissions of the BLOOM API have estimated that the real-time deployment of the model on a GCP instance with 16 GPUs running in the \texttt{us-central1} region results in approximately 20 kg of~\carboneq{} emitted per day of deployment (or 0.83 kg per hour). This figure is not representative of all deployment use-cases, and will vary depending on the hardware used as well as the specifics of model implementation (e.g.\ whether batching is used) and the number of requests the model receives. Further information regarding BLOOM's carbon footprint can be found in~\cite{luccioni2022estimating}.

\subsection{Release}

Openness has been central to the development of BLOOM and we wanted to ensure it is easily available for the community to use. As such, we worked on producing documentation as a Model Card \citep{ModelCard} and a new license addressing specific goals of the project.

\paragraph{Model Card}
Following best practices for releasing machine learning models, the BLOOM model has been released along with a detailed Model Card\footnote{\href{https://hf.co/bigscience/bloom}{\texttt{hf.co/bigscience/bloom}}}~\citep{ModelCard} describing its technical specifications, details on training, intended-use, out-of-scope uses as well as the model's limitations.
Participants across working groups worked together to produce the final Model Card and similar cards for each checkpoint. The work was collaborative, primarily composed ``live'' by thinking through and discussing each section, then further dividing into subsections based on the categorizations and distinctions participants naturally ended up creating throughout discussions.

\paragraph{Licensing}
Being mindful of the potentially harmful use-cases that BLOOM could enable, we chose to strike a balance between unrestricted open-access and responsible-use by including behavioral-use clauses~\citep{LicensingPaper} to limit the application of the model towards potentially harmful use-cases. Such clauses are routinely being included in a growing class of ``Responsible AI Licenses (RAIL)''\footnote{\href{http://licenses.ai}{\texttt{licenses.ai}}} that the community has been adopting when releasing their models.\footnote{\href{https://the-turing-way.netlify.app/reproducible-research/licensing/licensing-ml.html}{\texttt{the-turing-way.netlify.app/reproducible-research/licensing/licensing-ml.html}}} 
A distinguishing aspect of the RAIL license developed for BLOOM is that it separates licensing of the ``source code'' and ``model'', as referenced by its trained parameters. It further includes detailed definitions of ``use'' and ``derived works'' of the model to ensure that anticipated downstream use by prompting, finetuning, distillation, use of logits and probability distributions are explicitly identified. The license contains $13$ behavioral-use restrictions that have been identified based on the intended uses and limitations described in the BLOOM Model Card, as well as the BigScience ethical charter. The license offers the model at no charge and users are free to use the model as long as they comply with the terms (including usage restrictions). The source code for BLOOM has been made available under an Apache 2.0 open source license.

\section{Evaluation}
\label{sec:evaluation}

Our evaluations focus on zero-shot and few-shot settings. Our goal is to present an accurate picture of how BLOOM compares to existing LLMs in settings that most realistically reflect the way the models are likely to be used in practice. Because of the scale of these models, prompt-based adaptation and few-shot ``in-context learning'' are currently more common than finetuning. Thus, we report results on a range of tasks - SuperGLUE \ref{sec:eval:superglue}, machine translation \ref{sec:eval:mt}, summarization \ref{sec:eval:wikilingua} - and languages in zero-shot and one-shot prompt-based settings, as well as after multitask finetuning (\Cref{sec:eval:bloomz}). We also perform code generation \ref{sec:eval:code}, use BLOOM-derived text embeddings for representation tasks \ref{sec:sgptbloom} and interpret BLOOM's generalization abilities from the perspective of multilingual probing (\Cref{sec:eval:probing}).

\subsection{Experimental Design}
\label{sec:eval:design}

\subsubsection{Prompts} Based on recent research on the impact of prompting on language model performance, we decided to build a language model evaluation suite that allowed us to vary both the basic task data as well as the prompting that is used to contextualize the task. Our prompts were developed prior to BLOOM's release, and did not undergo any \textit{a priori} refinement using models. That is, the prompts we use in our evaluation are ones that humans believed were a reasonable way to solicit the desired task behavior from a language model. Our goal for designing prompts in this way is to simulate realistic zero-shot or one-shot results that a new user could expect from BLOOM. This is in contrast to presenting best-case performances that might result from multiple rounds of trial-and-error on prompt design. We choose to report the former because the latter is harder to reproduce systematically, is arguably a less representative picture of how the model works in the average setting, and is not representative of true zero-shot learning where no labeled data is available. 

We generate multiple prompts per task using \texttt{promptsource}~\citep{bach2022promptsource}. We follow the procedure used by \citet{sanh2022multitask}, in which prompt generation is crowdsourced, and thus we see substantial variety in length and style across prompts. To improve quality and clarity, multiple peer reviews were performed on each prompt for artifacts and consistency.

Table~\ref{tab:prompt-examples} shows examples of the resulting prompts used for the WMT'14 task. We also generate prompts for many tasks that are not included in this paper due to resource constraints. All of our prompts for all tasks (both those analyzed in the paper and those not yet analyzed) are publicly available.\footnote{\href{https://github.com/bigscience-workshop/promptsource/tree/eval-hackathon}{\texttt{github.com/bigscience-workshop/promptsource/tree/eval-hackathon}}}

\begin{table*}[ht!]
    \centering\small
    \resizebox{\textwidth}{!}{
    \begin{tabular}{lll}
    \toprule
    Prompt name & Prompt & Target \\
    \midrule
       a\_good\_translation-source+target & Given the following source text: [source sentence], a good L2 translation is: & [target sentence]\\
       gpt3-target & What is the L2 translation of the sentence: [source sentence]? & [target sentence] \\ 
      version-target & if the original version says [source sentence]; then the L2 version should say: & [target sentence] \\
      xglm-source+target &  L1: [source sentence] = L2: & [target sentence]\\
       \bottomrule
    \end{tabular}}
    \caption{Four prompts for the WMT'14 dataset \citep{bojar-etal-2014-findings} for MT evaluation. Above, ``L1'' and ``L2'' are~ replaced with language names (e.g.~``Bengali'' and ``Russian'').}
    \label{tab:prompt-examples}
\end{table*}

\subsubsection{Infrastructure} 
Our framework extends EleutherAI's Language Model Evaluation Harness \citep{eval-harness} by integrating it with the \texttt{promptsource}~\citep{bach2022promptsource} library described in \Cref{sec:prompted-data}. We release our Prompted Language Model Evaluation Harness 
as an open source library for people to use.
We use this framework in order to run the experiments and aggregate results.

\subsubsection{Datasets} 

\paragraph{SuperGLUE} We use a subset of the SuperGLUE \citep{wang-et-al-2019-superglue} evaluation suite of classification tasks, specifically: Ax-b, Ax-g, BoolQ, CB, WiC, WSC, and RTE tasks.
We excluded the remaining tasks because they require an order of magntiude more compute to run than all of these tasks we consider combined.
These tasks are English-only, and are thus included to facilitate comparison with prior work, which has primarily focused on English-only models. We also note that performance on these tasks has not yet been widely reported using zero- and one-shot prompt-based setting. T0 \citep{sanh2022multitask} is the first exception, but that model is instruction-tuned and thus not directly comparable to models like BLOOM and OPT. For each task, we select a random sample of five prompts from \texttt{promptsource} and evaluate all models on that set of prompts.
As with other prompting tasks in Evaluation Harness~\citep{eval-harness}, the prediction of a model for a given prompt is measured using the maximum log likelihood among a set of specified candidate label strings associated with the prompt. 

\paragraph{Machine Translation (MT)} We evaluate BLOOM on three datasets (using ISO-639-1 codes to refer to languages): WMT14 en$\leftrightarrow$fr and en$\leftrightarrow$hi \citep{bojar-etal-2014-findings}, Flores-101 \citep{goyal-etal-2022-flores} and DiaBLa \citep{bawden-et-al-2020-diabla}. We evaluate using the \texttt{sacrebleu} \citep{post-2018-call} implementation of BLEU \citep{papineni-etal-2002-bleu}, using default tokenisation for WMT and DiaBLa and \texttt{spm-flores-101} for Flores.\footnote{BLEU+case:mixed+numrefs.1+smooth.exp+\{13a,tok:spm-flores\}+version:2.2.1} We use greedy decoding with generation proceeding until the EOS token, or additionally \texttt{\textbackslash n\#\#\#\textbackslash n} for the 1-shot case. The maximum generation length was set per dataset to be in line with what is typically used in the literature; specifically, 64 tokens for WMT14 and 512 tokens for Flores-101 and DiaBla.
Task-specific experimental design details are below.

\paragraph{Summarization} We evaluate summarization on the WikiLingua \citep{ladhak-etal-2020-wikilingua} dataset. WikiLingua is a multilingual summarization dataset comprising WikiHow article and step-by-step summary pairs. Pairs are aligned across multiple languages, with translation of source and summary further reviewed by an international translation team. One-shot conditional natural language generation has typically not been reported by models with size comparable to BLOOM. PaLM \citep{chowdhery2022palm} is the first exception, and reports scores on WikiLingua; however, only the model's ability to summarize in English was examined (-> en). By contrast, we opted to test BLOOM's inherent multilingual ability by assessing the abstractive summarization in the source language (e.g.~vi -> vi). We focus on the nine 
languages (Arabic, English, Spanish, French, Hindi, Indonesian, Portuguese, Vietnamese and Chinese) which were amongst those targeted as part of the BigScience effort. 

Natural language generation is notoriously challenging to evaluate, with multilingual generation compounding this challenge due to a lack of metric support. Following the suggestions by \citet{gehrmann_cracked}, we report ROUGE-2, ROUGE-L \citep{lin-2004-rouge},\footnote{For ROUGE, we used the Python implementation at\\ \href{https://github.com/google-research/google-research/tree/master/rouge}{\texttt{github.com/google-research/google-research/rouge}}, commit \texttt{f935042}.} and Levenshtein distance. One important modification to ROUGE is using the SentencePiece tokenizer \citep{kudo-richardson-2018-sentencepiece} built from the Flores-101 dataset \citep{goyal-etal-2022-flores}.
A naive approach would use a tokenizer based on English, but using a multilingual tokenizer improves the capacity to measure the fidelity of multilingual generations. To minimize inference time of the model we use the subsamples from the updated GEM benchmark \citep{gemv2} (3000 uniformly sampled test examples).
The authors note that there is minimal difference when comparing model performance between the subsamples and the full test sets.
For decoding and generation, we use the same procedure as described above for MT.

\subsubsection{Baseline Models}

We use the following baseline models where appropriate (e.g.\ in settings where they support the language of the evaluation dataset): 
\begin{itemize}
    \item mGPT \citep{shliazhko2022mgpt}, GPT-style models trained on 60 languages from Wikipedia and Common Crawl
    \item GPT-Neo \citep{black2021gpt}, GPT-J-6B \citep{wang2021gpt}, and GPT-NeoX \citep{black2022gpt}, a family of GPT-style models trained on The Pile \citep{gao2020pile}
    \item T0 \citep{sanh2022multitask}, a variant of T5 \citep{raffel2020exploring} that underwent multitask prompted finetuning on datasets from P3 \citep{bach2022promptsource}
    \item OPT \citep{zhang2022opt}, a family of GPT-style model trained on a mixture of datasets including those from RoBERTa \cite{liu2019roberta} and The Pile \citep{gao2020pile}
    \item XGLM \citep{lin2021xglm}, a GPT-style multilingual model trained on a variant of CC100 \citep{conneau2020unsupervised}
    \item M2M \citep{fan-etal-2021-beyond}, a massively multilingual model trained to translate between 100 languages
    \item AlexaTM \citep{soltan-etal-2022-alexatm}, an encoder-decoder model trained on a mixture of masked and causal language modeling on data from Wikipedia and mC4 \citep{xue2021mt5}
    \item mTk-Instruct \citep{wang2022benchmarking}, a variant of T5 that underwent multitask prompted finetuning on datasets from Super-NaturalInstructions
    \item Codex \citep{chen2021evaluating}, a family of GPT models finetuned on code from GitHub
    \item GPT-fr \citep{simoulin:hal-03265900}, a GPT-style model trained on French text
\end{itemize}


\subsection{SuperGLUE}
\label{sec:eval:superglue}

\begin{figure}[!htbp]
    \centering
    \includegraphics[width=\linewidth]{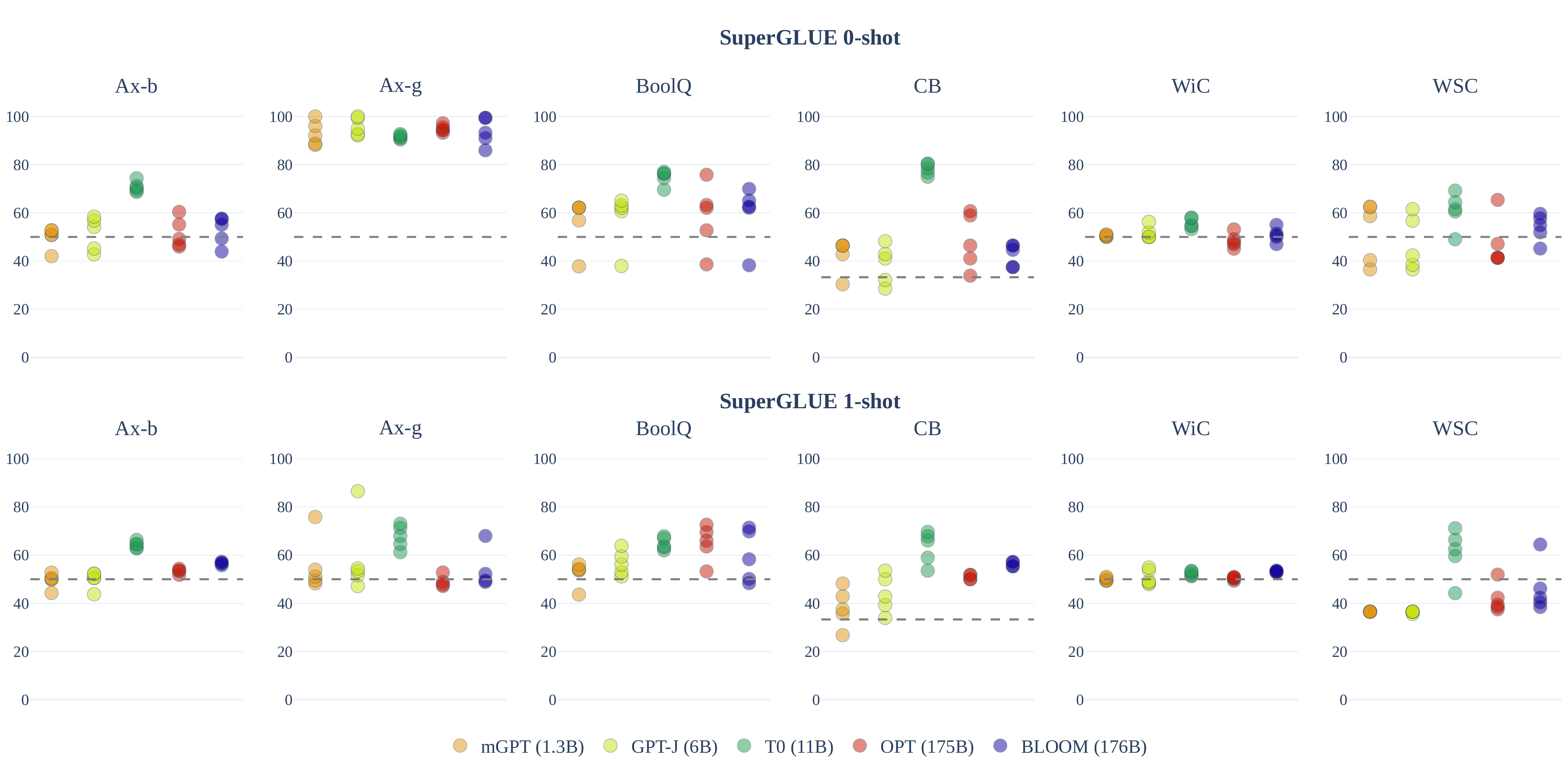}
    \caption{Performance of various LLMs on subset of tasks from SuperGLUE benchmark in zero- and one-shot prompt-based setting.}
    \label{fig:superglue}
\end{figure}

Figure \ref{fig:superglue} shows zero- and one-shot performance on SuperGLUE. In both settings, on entailment tasks (BoolQ and CB), performance is well above random chance for BLOOM, T0, OPT, and GPT-J. On other tasks, while the best prompts do better, the average performance across prompts hovers around chance, suggesting that the success of individual prompts is primarily statistical variation. There is some signal for BLOOM in the diagnostic (Ax-b and Ax-g) datasets. The exception is the T0 model, which shows strong performance. However, this model is finetuned in the multitask setting (similar to BLOOMZ, see \Cref{sec:eval:bloomz}) in order to improve performance in zero-shot prompting settings, and thus is not directly comparable to the other models shown here. 

As models go from zero-shot to one-shot, variability is reduced across all prompts and models and performance slightly and inconsistently increases. Notably, BLOOM sees more of an increase in performance than comparable models when going from zero-shot to one-shot, as it is generally behind OPT in the zero-shot setting but matches or improves on it in the one-shot setting, even though it has only partly been trained on English. This may be because a multilingual language model gains more certainty in the language of input and output with a longer context.

\begin{figure}[ht!]
    \centering
    \includegraphics[width=\linewidth]{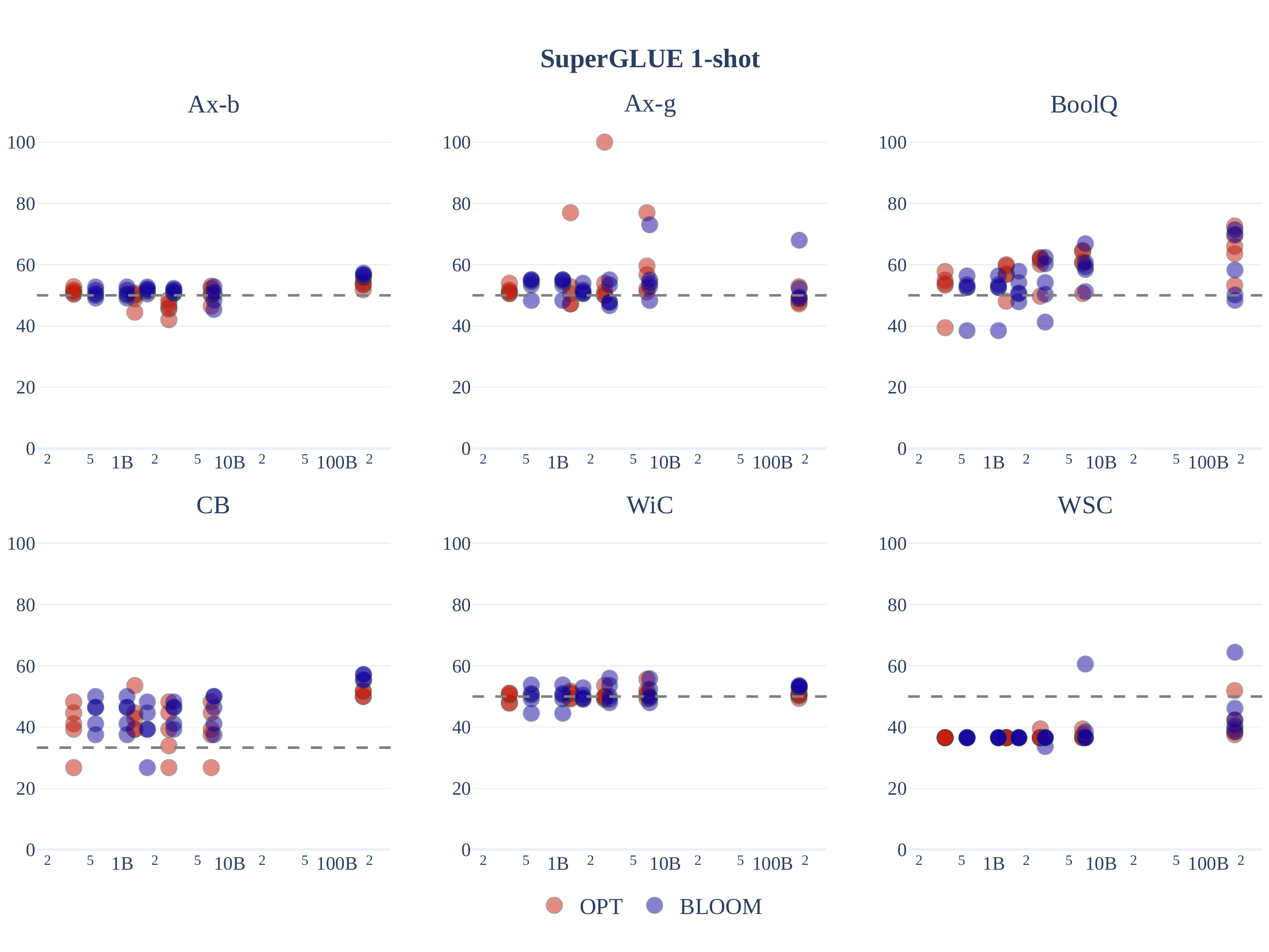}
    \caption{Comparison of the scaling of BLOOM versus OPT on each SuperGLUE one-shot task. Each point represents the average accuracy of a model within the BLOOM or OPT family of models on one of the five task prompts. The number of parameters on the x-axis is presented in log-scale.}
    \label{fig:superglue-scaling}
\end{figure}

We perform an additional analysis comparing BLOOM models across model sizes.  As a baseline, we also measure the average one-shot accuracy of OPT models of similar sizes (350M parameters to 175B parameters).\footnote{We do not evaluate OPT-66B because of the lack of a similarly-sized BLOOM model.} Figure \ref{fig:superglue-scaling} shows the accuracy of each prompt on each task across model scales. Both OPT and BLOOM model families improve very slightly with scale, with only models over 2 billion parameters showing signal, and there is no consistent difference between families across all tasks. In the 1-shot setting, BLOOM-176B is ahead of OPT-175B on Ax-b, CB, WSC and WiC, and matches it on the other tasks, suggesting that multilinguality does not limit performance of BLOOM on English-only tasks in the zero-shot setting.

\subsection{Machine Translation}
\label{sec:eval:mt}

In addition to the results presented here, a more detailed analysis of BLOOM's MT quality can be found in \citep{bawden-yvon-2023-mt-bloom}.

\subsubsection{WMT}
\label{sec:eval:mt:wmt}

WMT results for BLOOM-176B in the zero-shot and 1-shot setting are given in Table~\ref{tab:WMT-14enfr-byprompts-0shot}.
The best prompts tend to be the more verbose ones; the ``version-target'' prompt is consistently better and the ``gpt3-target'' and ``xglm-source+target'' prompts have very poor performance, especially for zero-shot. In the one-shot setting, BLOOM can, with the right prompt, perform competent translation, although it is behind dedicated (supervised) models such as M2M-100 (43.8 BLEU for English$\rightarrow$French and 40.4 for French$\rightarrow$English, compared to 34.2 and 35.4 BLEU for BLOOM).
The two major problems observed, particularly in the zero-shot setting, are (i)~over-generation and (ii)~not producing the correct language (an obvious prerequisite for a good translation). Both of these aspects are greatly improved as the number of few-shot examples is increased. 

\begin{table}[ht!]
\centering\small
\addtolength{\tabcolsep}{-2.5pt}
\begin{tabular}{lrr|rr|rr|rr}
\toprule
Prompt & \multicolumn{2}{c}{en $\rightarrow$ fr} & \multicolumn{2}{c}{fr $\rightarrow$ en } & \multicolumn{2}{c}{en $\rightarrow$ hi} & \multicolumn{2}{c}{hi $\rightarrow$ en} \\
\midrule
Shots  & 0 & 1 & 0 & 1 & 0 & 1 & 0 & 1 \\
\midrule
a\_good\_translation-source+target & 15.38 & 36.39 & 14.15 & 36.56 & 1.90 & 14.49 & 10.19 & 24.60  \\
gpt3-target     & 7.90 & 32.55 & 12.73 & 33.14 & 0.26 & 6.51 & 0.66 & 9.98 \\
version-target  & 21.96 & 34.22 & 26.79 & 35.42 & 1.96 & 13.95 & 11.48 & 25.80 \\
xglm-source+target     & 14.91 & 27.83 & 15.52 & 34.51 & 6.80 & 13.62 & 12.05 & 25.04 \\
\bottomrule
\end{tabular}
\caption{WMT'14 zero- and one-shot results (BLEU) for BLOOM-176B. The prompts used are described in Table~\ref{tab:prompt-examples}.}
\label{tab:WMT-14enfr-byprompts-0shot}
\end{table}

\subsubsection{DiaBLa}

\begin{table}[!ht]
    \centering\small
    \begin{tabular}{llrrrr}
    \toprule
    & & \multicolumn{2}{c}{en$\rightarrow$fr} & \multicolumn{2}{c}{fr$\rightarrow$en} \\
    1-shot context & Truncate & BLEU & COMET & BLEU & COMET \\
    \toprule
    \multirow{2}{*}{Rand.}   & $\times$ &  5.7 & 0.342 & 12.1 & 0.614 \\
    &  $\checkmark$ & 37.6 & \textbf{0.634} & 41.4 & \textbf{0.757} \\
    \midrule
    \multirow{2}{*}{Prev.}   & $\times$ & 6.1 & 0.328 & 12.3 & 0.617 \\
    & $\checkmark$ & \textbf{38.5} & 0.614 & \textbf{41.6} & 0.751 \\
    \bottomrule
    \end{tabular}
    \caption{DiaBLa 1-shot results (BLEU) for the ``xglm-source+target'' prompt when using the previous or a random sentence as the 1-shot example (with and without truncation of outputs). In bold the best results for each direction.}
    \label{tab:diabla-context-results}
\end{table}

Table~\ref{tab:diabla-context-results} shows results testing the use of linguistic context with DiaBLa, a parallel dataset of informal bilingual dialogues. In a 1-shot context and using the ``xglm-source+target'' prompt, we compare the effect of using a random test set example as the 1-shot example versus using the previous dialogue utterance. In light of the overgeneration issues seen and in order to evaluate the quality of the prediction independently of overgeneration, we report results for both original outputs and after applying a custom truncation function.\footnote{The truncation rule is specific to the ``xglm-source+target'' prompt and the fact that overgeneration consists of repeating the prompt pattern. Anything after a first newline or the regular expression pattern \texttt{= .+?:} is discarded.} The automatic results are inconclusive, with little difference between scores (BLEU scores are higher for previous context but COMET scores are lower). Despite these results, there is evidence in the predictions themselves that the model is able to use the context of the 1-shot example to make translation choices. See \citep{bawden-yvon-2023-mt-bloom} for examples and further analysis.




\subsubsection{Flores}

\begin{table}[!ht]
 \centering\footnotesize
 \begin{subtable}[t]{0.48\textwidth}
 \scalebox{0.86}{
  \begin{tabular}{llrrrrr}
\toprule
Src$\downarrow$ & Trg$\rightarrow$ & en & bn & hi & sw & yo \\
\midrule
\multirow{2}{*}{en} & BLOOM & -- & 24.6 & 27.2 & 20.5 & 2.6 \\
 & M2M & -- & 23.0 & 28.1 & 26.9 & 2.2 \\
\midrule
\multirow{2}{*}{bn} & BLOOM & 29.9 & -- & 16.3 & -- & -- \\
 & M2M & 22.9 & -- & 21.8 & -- & -- \\
\midrule
\multirow{2}{*}{hi} & BLOOM & 35.1 & 23.8 & -- & -- & -- \\
 & M2M & 27.9 & 21.8 & -- & -- & -- \\
\midrule
\multirow{2}{*}{sw} & BLOOM & 37.4 & -- & -- & -- & 1.3 \\
 & M2M & 30.4 & -- & -- & -- & 1.3 \\
\midrule
\multirow{2}{*}{yo} & BLOOM & 4.1 & -- & -- & 0.9 & -- \\
 & M2M & 4.2 & -- & -- & 1.9 & -- \\
  \bottomrule
\end{tabular}}
\caption{Low-resource languages}
\label{tab:flores101_summary:high-low}
\end{subtable}
\hfill
 \begin{subtable}[t]{0.48\textwidth}
 \centering\footnotesize
  \scalebox{0.86}{
 \begin{tabular}{llrrrrrr}
\toprule
Src$\downarrow$ & Trg$\rightarrow$ & ca & es & fr & gl & it & pt \\
\midrule
\multirow{2}{*}{ca} & BLOOM & -- & 28.9 & 33.8 & 19.2 & 19.8 & 33.0 \\
 & M2M & -- & 25.2 & 35.1 & 33.4 & 25.5 & 35.2 \\
\midrule
\multirow{2}{*}{es} & BLOOM & 31.2 & -- & 24.8 & 23.3 & 16.5 & 29.1 \\
 & M2M & 23.1 & -- & 29.3 & 27.5 & 23.9 & 28.1 \\
\midrule
\multirow{2}{*}{fr} & BLOOM & 37.2 & 27.5 & -- & 24.9 & 24.0 & 38.9 \\
 & M2M & 28.7 & 25.6 & -- & 32.8 & 28.6 & 37.8 \\
\midrule
\multirow{2}{*}{gl} & BLOOM & 37.5 & 27.1 & 33.8 & -- & 18.3 & 32.2 \\
 & M2M & 30.1 & 27.6 & 37.1 & -- & 26.9 & 34.8 \\
\midrule
\multirow{2}{*}{it} & BLOOM & 31.0 & 25.4 & 31.4 & 20.2 & -- & 29.2 \\
 & M2M & 25.2 & 29.2 & 34.4 & 29.2 & -- & 31.5 \\
\midrule
\multirow{2}{*}{pt} & BLOOM & 39.6 & 28.1 & 40.3 & 27.1 & 20.1 & -- \\
 & M2M & 30.7 & 26.9 & 40.2 & 33.8 & 28.1 & -- \\
    \bottomrule
  \end{tabular}}
\caption{Romance languages}
\label{tab:flores101_summary:same-fami}
\end{subtable}

\vspace{1em}

\begin{subtable}[b]{0.48\textwidth}
 \centering\footnotesize
\scalebox{0.85}{
\begin{tabular}{llrrrrr}
\toprule
Src $\downarrow$ & Trg $\rightarrow$ & ar & en & es & fr & zh \\
\midrule
\multirow{3}{*}{ar} & BLOOM & -- & 40.3 & 23.3 & 33.1 & 17.7 \\
 & M2M & -- & 25.5 & 16.7 & 25.7 & 13.1 \\
 & AlexaTM & -- & 41.8 & 23.2 & 35.5 & -- \\
\midrule
\multirow{3}{*}{en} & BLOOM & 28.2 & -- & 29.4 & 45.0 & 26.7 \\
 & M2M & 17.9 & -- & 25.6 & 42.0 & 19.3 \\
 & AlexaTM & 32.0 & -- & 31.0 & 50.7 & -- \\
\midrule
\multirow{3}{*}{es} & BLOOM & 18.8 & 32.7 & -- & 24.8 & 20.9 \\
 & M2M & 12.1 & 25.1 & -- & 29.3 & 14.9 \\
 & AlexaTM & 20.8 & 34.6 & -- & 33.4 & -- \\
\midrule
\multirow{3}{*}{fr} & BLOOM & 23.4 & 45.6 & 27.5 & -- & 23.2 \\
 & M2M & 15.4 & 37.2 & 25.6 & -- & 17.6 \\
 & AlexaTM & 24.7 & 47.1 & 26.3 & -- & --\\
\midrule
\multirow{3}{*}{zh} & BLOOM & 15.0 & 30.5 & 20.5 & 26.0 & -- \\
 & M2M & 11.55 & 20.9 & 16.9 & 24.3 & -- \\
 & AlexaTM & -- & -- & -- & -- & --\\
\bottomrule
\end{tabular}}
\caption{High-resource language pairs.}
\label{tab:flores101_summary:high-high}
\end{subtable}
\hfill
\begin{subtable}[b]{0.48\textwidth}
\centering\footnotesize
\scalebox{0.85}{
\begin{tabular}{llrrrrr}
\toprule
Src $\downarrow$ & Trg $\rightarrow$ & en & fr & hi & id & vi \\
\midrule
\multirow{2}{*}{en} & BLOOM & -- & 45.0 & 27.2 & 39.0 & 28.5 \\
 & M2M & -- & 42.0 & 28.1 & 37.3 & 35.1 \\
\midrule
\multirow{2}{*}{fr} & BLOOM & 45.6 & -- & 18.5 & 31.4 & 32.8 \\
 & M2M & 37.2 & -- & 22.9 & 29.1 & 30.3 \\
\midrule
\multirow{2}{*}{hi} & BLOOM & 35.1 & 27.6 & -- & -- & -- \\
 & M2M & 27.9 & 25.9 & -- & -- & -- \\
\midrule
\multirow{2}{*}{id} & BLOOM & 43.2 & 30.4 & -- & -- & -- \\
 & M2M & 33.7 & 30.8 & -- & -- & -- \\
\midrule
\multirow{2}{*}{vi} & BLOOM & 38.7 & 26.8 & -- & -- & -- \\
 & M2M & 29.5 & 25.8 & -- & -- & -- \\
  \bottomrule
\end{tabular}}
\caption{High$\rightarrow$mid-resource language pairs.}
\label{tab:flores101:high-mid}
\end{subtable}
\caption{\label{tab:flores-results}1-shot MT results (spBLEU) on the Flores-101 devtest set. }
\end{table}

In the 1-shot setting, we test several language directions in the Flores-101 \citep{goyal-etal-2022-flores} devtest set using the ``xglm-source+target'' prompt \citep{lin2021xglm}. 
The 1-shot example is randomly taken from the dev set.
We separate out results for low-resource language pairs (\Cref{tab:flores101_summary:high-low}), between related languages of the Romance language family (\Cref{tab:flores101_summary:same-fami}), high-resource language pairs (\Cref{tab:flores101_summary:high-high}) and high-to-mid-resource language pairs (\Cref{tab:flores101:high-mid}). 
Languages are classified as low-, mid- and high-resource depending on their representation in ROOTS.
We compare to supervised results from the M2M-100 model \citep{fan-etal-2021-beyond} with 615M parameters, for which scores are computed by \cite{goyal-etal-2022-flores}. Additionally, we compare to 
32-shot AlexaTM results for high-resource language pairs \citep{soltan-etal-2022-alexatm}. 
Results are good across the board for both translation between high-resource languages and from high- to mid-resource languages, suggesting BLOOM's good multilingual capacity, even across scripts (here between Latin (or extended Latin), Chinese, Arabic and Devanagari scripts). Compared to the supervised M2M-100 model, results are often comparable and sometimes better in this 1-shot setting, and results are comparable in many cases to those of AlexaTM (even though AlexTM results are for 32-shot).

The translation quality for many of the low-resource languages is good, comparable to or even slightly better than the supervised M2M model.
However, results are very poor between Swahili and Yoruba, languages that are present but under-represented in BLOOM's training data ($<$50k tokens each). This contrasts with the results for translation between Romance (and therefore related) languages, where results are good across-the-board, including for translation from Galician (glg), a language not included in the training data, but which shares many similarities with the other Romance languages, in particular with Portuguese (por). This however does question BLOOM's quality on those under-represented low-resource languages included in training.

\subsection{Summarization}
\label{sec:eval:wikilingua}

\begin{figure*}[ht!]
    \centering
    \includegraphics[width=\linewidth]{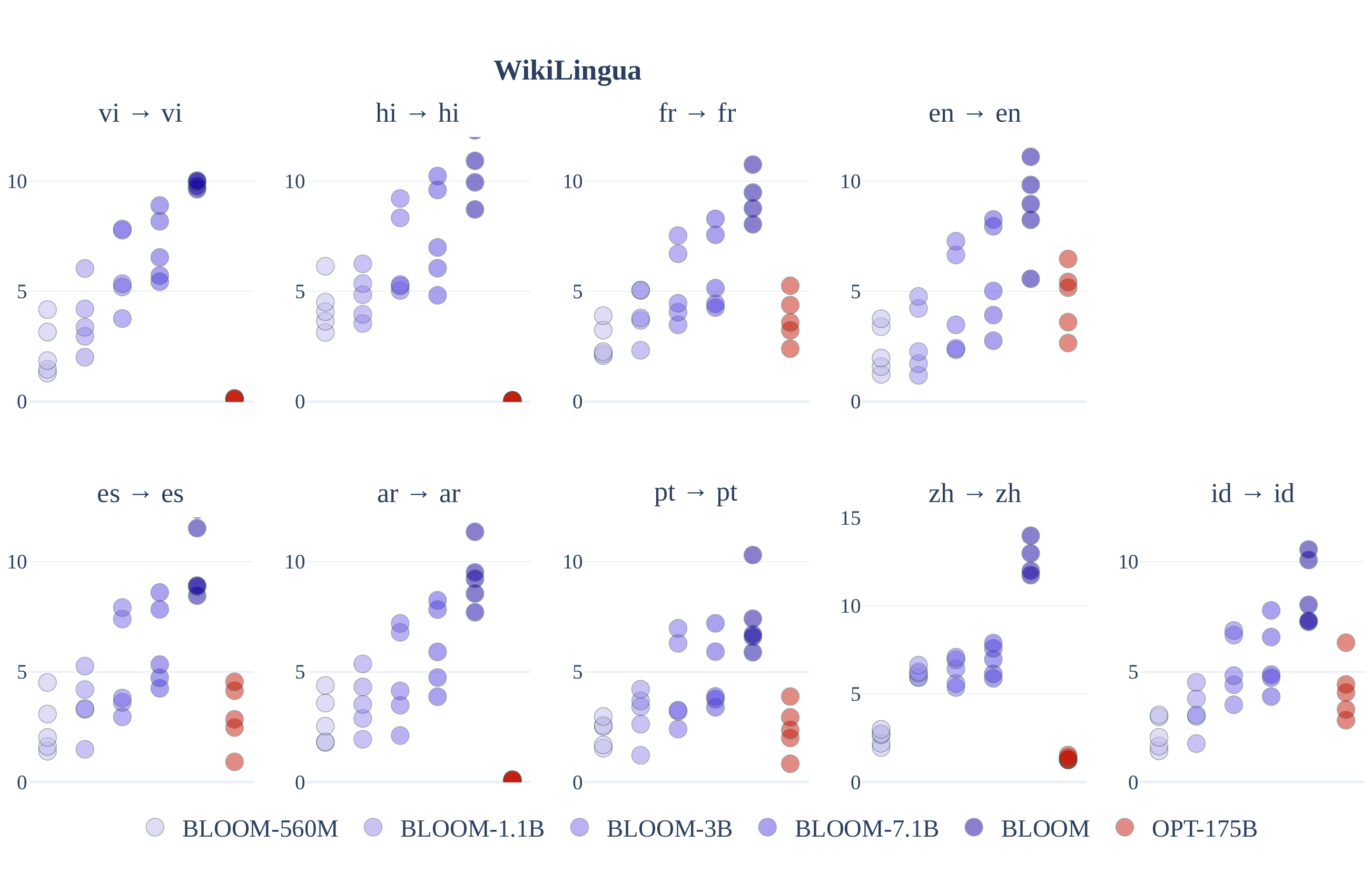}
    \caption{WikiLingua One-shot Results. Each plot represents a different language with per-prompt ROUGE-2 F-measure scores.}
    \label{fig:wikilingua}
\end{figure*}

Figure \ref{fig:wikilingua} shows one-shot results for BLOOM models alongside OPT-175B for comparison. Each point represents a per-prompt score. The key takeaways are that BLOOM attains higher performance on multilingual summarization than OPT and that performance increases as the parameter count of the model increases. We suspect this is due to BLOOM's multilingual-focused training.

As discussed in Section \ref{sec:eval:design}, we report ROUGE-2 scores for the sake of comparability with prior work, and because there is a lack of alternatives for generation evaluation. 
However, we qualitatively observe that in many cases, the ROUGE-2 score understates the quality of the summaries generated by the systems.

\subsection{Code Generation}
\label{sec:eval:code}

\begin{table}[ht!]
    \begin{center}
    \begin{small}
    \begin{sc}
    \begin{tabular}{lccc}
    \toprule
    & \multicolumn{3}{c}{pass@$k$} \\
    & $k=1$ & $k=10$ & $k=100$ \\
    \midrule
    GPT-Neo 1.3B & 4.79\% & 7.47\% & 16.30\% \\
    GPT-Neo 2.7B & 6.41\% & 11.27\% & 21.37\% \\
    GPT-J 6B & 11.62\% & 15.74\% & 27.74\% \\
    GPT-NeoX 20B & 15.4\% & 25.6\% & 41.2\% \\
    \midrule
    Codex-300M & 13.17\% & 20.37\% & 36.27\% \\
    Codex-679M & 16.22\% & 25.7\% & 40.95\% \\
    Codex-2.5B & 21.36\% & 35.42\% & 59.5\% \\
    Codex-12B & 28.81\% & 46.81\% & 72.31\% \\
    \midrule
    BLOOM-560M & 0.82\% & 3.02\% & 5.91\% \\
    BLOOM-1.1B & 2.48\% & 5.93\% & 9.62\% \\
    BLOOM-1.7B & 4.03\% & 7.45\% & 12.75\% \\
    BLOOM-3B & 6.48\% & 11.35\% & 20.43\% \\
    BLOOM-7.1B & 7.73\% & 17.38\% & 29.47\% \\
    BLOOM & 15.52\% & 32.20\% & 55.45\% \\
    \midrule
    BLOOMZ-560M & 2.18 \% & 4.11\% & 9.00\% \\
    BLOOMZ-1.1B & 2.63\% & 6.22\% & 11.68\% \\
    BLOOMZ-1.7B & 4.38\% & 8.73\% & 16.09\% \\
    BLOOMZ-3B & 6.29\% & 11.94\% & 19.06\% \\
    BLOOMZ-7.1B & 8.06\% & 15.03\% & 27.49\% \\
    BLOOMZ & 12.06\% & 26.53\% & 48.44\% \\
    \bottomrule
    \end{tabular}
    \caption{
        Performance on HumanEval \citep{chen2021evaluating}. Non-BLOOM results come from prior work \citep{chen2021evaluating,fried2022incoder}. The Codex model is a language model that was finetuned on code, while the GPT models \citep{black2021gpt,wang2021gpt,black2022gpt} are trained on a mix of code and text like BLOOM.
    }
    \label{tab:humaneval}
    \end{sc}
    \end{small}
    \end{center}
\end{table}

The BLOOM pretraining corpus, ROOTS, consists of around 11\% of code. In Table~\ref{tab:humaneval}, we report benchmarking results of BLOOM on HumanEval \citep{chen2021evaluating}. We find the performance of pretrained BLOOM models to be similar to that of the similar-sized GPT models trained on the Pile \citep{gao2020pile}. The Pile contains English data and around 13\% of code (GitHub + StackExchange), which is similar to the code data sources and proportions in ROOTS. The Codex models, which have solely been finetuned on code, are significantly stronger than other models. Multitask finetuned BLOOMZ models do not improve significantly over BLOOM models. We hypothesize this is due to the finetuning dataset, xP3, not containing significant amounts of pure code completion. Rather, xP3 contains code-related tasks, such as estimating the time complexity of a given Python code snippet. Additional analysis is provided in~\cite{muennighoff2022crosslingual}.

\subsection{HELM benchmark}

\begin{figure*}[h!]
    \centering
    \includegraphics[width=\linewidth]{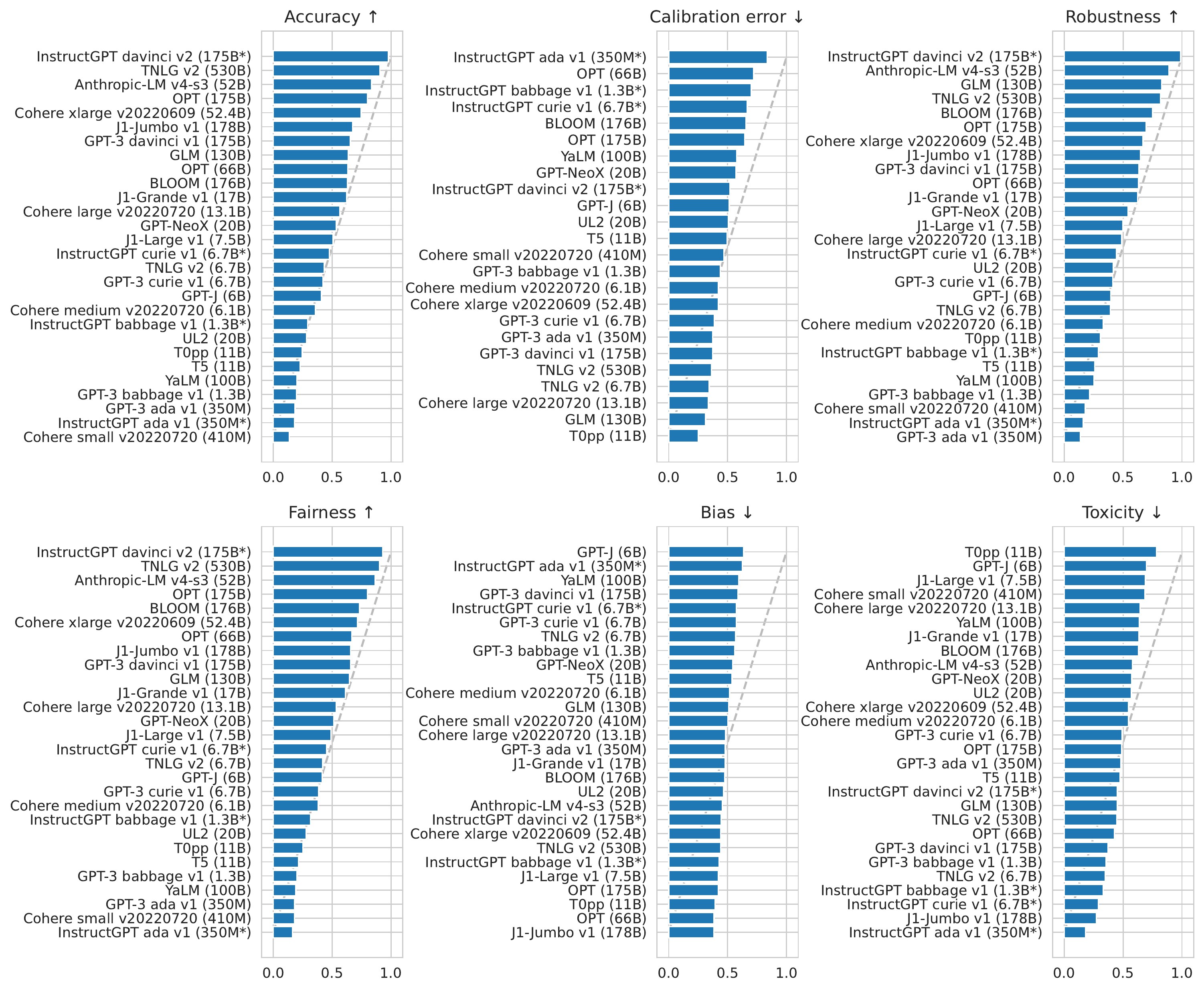}
    \caption{Results for a wide variety of language models on the 5-shot HELM benchmark. Taken from \cite{HELM}}
    \label{fig:helm}
\end{figure*}

For completeness, we reproduce here evaluations from the HELM benchmark \citep{HELM}, which ran 5-shot evaluations of a variety of language models on English-only tasks. Despite the multilingual training, BLOOM is roughly on par in accuracy with previous-generation English-only models, such as GPT3-davinci v1 and J1-Grande v1, but behind more recent monolingual models such as InstructGPT davinci v2, Turing NLG v2, Anthropic-LM v4-s3, or OPT. Like other large language models of this size, it is not very well calibrated, but quite robust. Finally, on this benchmark, it is one of the best models for fairness, slightly more toxic than average in English, and average for bias.

\subsection{Multitask Finetuning}
\label{sec:eval:bloomz}

\begin{figure*}[ht!]
    \centering
    \includegraphics[width=\linewidth]{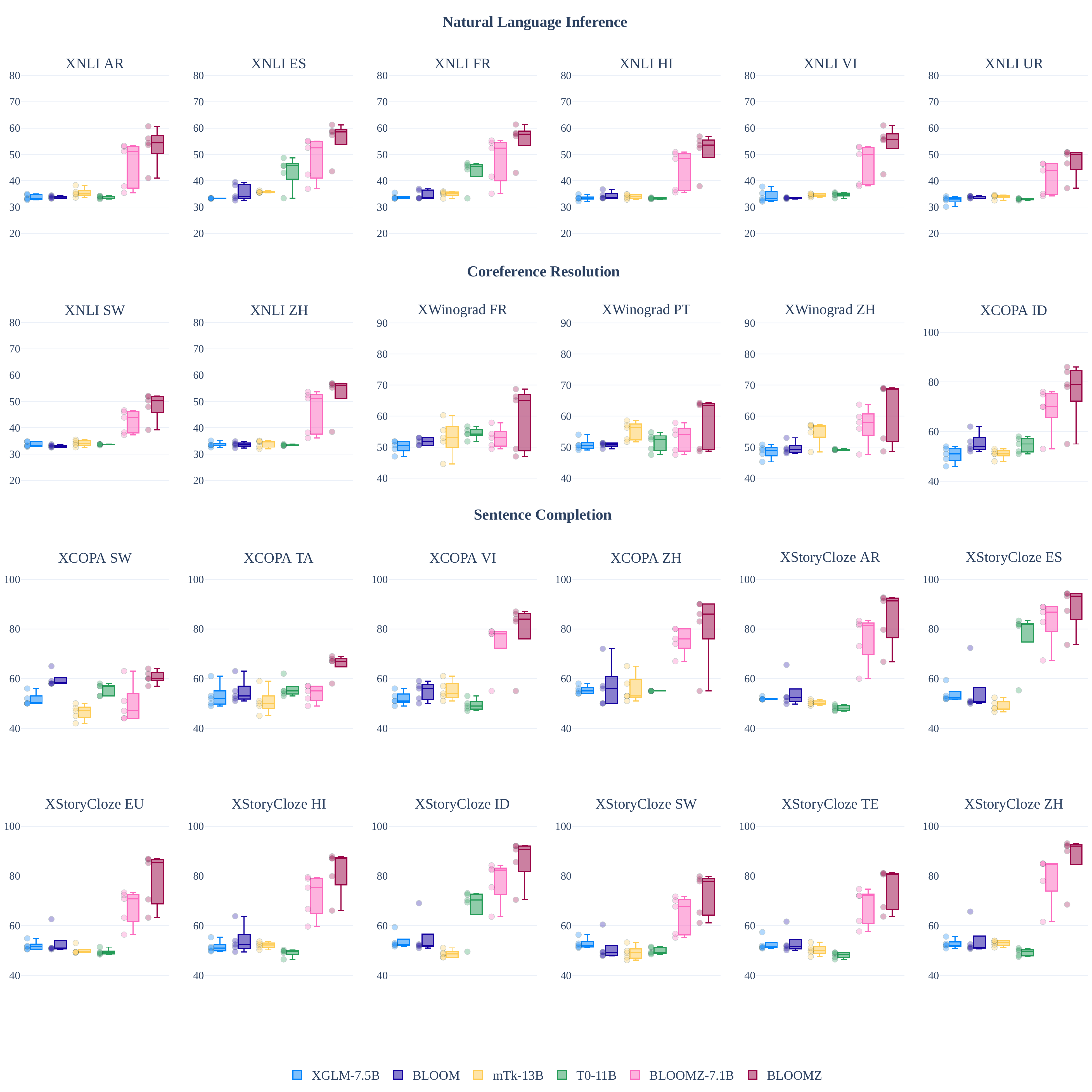}
    \caption{BLOOMZ zero-shot task generalization. Five untuned prompts are evaluated for each dataset and plotted. T0 is monolingual (English) while other models are multilingual. T0 performance may be hurt by its inability to tokenize some non-English texts.}
    \label{fig:taskgen}
\end{figure*}

Building on recent work on multitask finetuning \citep{sanh2022multitask,wei2021finetuned,wang2022what} we explore using \textit{multilingual} multitask finetuning to improve the zero-shot performance of the BLOOM model. We conducted multilingual multitask finetuning of BLOOM models using the xP3 corpus outlined in Section \ref{sec:prompted-data}. We find that zero-shot performance significantly increases. In Figure~\ref{fig:taskgen}, we compare the zero-shot performance of pretrained BLOOM and XGLM models with multitask finetuned BLOOMZ, T0 and mTk-Instruct~\citep{wang2022benchmarking}. BLOOM and XGLM performances are near the random baselines of 33\% for NLI (XNLI) and 50\% for coreference resolution (XWinograd) and sentence completion (XCOPA and XStoryCloze). After going through multilingual multitask finetuning (BLOOMZ), zero-shot performance significantly improves on the depicted held-out tasks. Despite also being multitask finetuned, T0 performs badly on the multilingual datasets shown due to it being a monolingual English model. Additional results provided in~\cite{muennighoff2022crosslingual}, however, show that models finetuned on xP3 also outperform T0 on English datasets when controlling for size and architecture. This is likely due to T0's finetuning dataset (P3) containing less diverse datasets and prompts than xP3. Multitask finetuning performance has been shown to correlate with the amount of datasets and prompts~\citep{chung2022scaling}.

\subsection{Embeddings}
\label{sec:sgptbloom}

\begin{table*}[th!]
\centering
\scriptsize
\resizebox{\textwidth}{!}{%
\begin{tabular}{lccccccc}
\toprule
{} & ST5-XL & LASER2 & MiniLM-L12 \footnote{\url{https://hf.co/sentence-transformers/paraphrase-multilingual-MiniLM-L12-v2}} & MPNet\footnote{\url{https://hf.co/sentence-transformers/https://huggingface.co/sentence-transformers/paraphrase-multilingual-mpnet-base-v2}} & LaBSE & SGPT-BLOOM-1.7B & SGPT-BLOOM-7.1B \\
\midrule
\multicolumn{8}{c}{\emph{Embedding classification performance on MASSIVE~\citep{fitzgerald2022massive} scored using accuracy}} \\
\midrule
Arabic (ar) & 4.18 & 37.16 & 51.43 & 45.14 & 50.86 & 54.59 & \textbf{59.25} \\
Bengali (bn) & 2.60 & 42.51 & 48.79 & 35.34 & 58.22 & 57.76 & \textbf{61.59} \\
English (en) & \textbf{72.09} & 47.91 & 69.32 & 66.84 & 61.46 & 66.69 & 69.67 \\
Spanish (es) & 57.97 & 45.44 & 64.43 & 59.66 & 58.32 & 61.77 & \textbf{66.35} \\
French (fr) & 60.99 & 46.13 & 64.82 & 60.25 & 60.47 & 64.58 & \textbf{66.95} \\
Hindi (hi) & 3.02 & 40.20 & 62.77 & 58.37 & 59.40 & 60.74 & \textbf{63.54} \\
Indonesian (id) & 41.53 & 45.81 & \textbf{65.43} & 59.85 & 61.12 & 60.07 & 64.06 \\
Kannada (kn) & 2.79 & 4.32 & 50.63 & 40.98 & \textbf{56.24} & 48.56 & 53.54 \\
Malayalam (ml) & 2.98 & 41.33 & 54.34 & 42.41 & 57.91 & 55.10 & \textbf{58.27} \\
Portuguese (pt) & 57.95 & 48.55 & 64.89 & 61.27 & 60.16 & 62.52 & \textbf{66.69} \\
Swahili (sw) & 30.60 & 31.89 & 31.95 & 29.57 & \textbf{51.62} & 43.90 & 49.81 \\
Tamil (ta) & 1.79 & 29.63 & 50.17 & 36.77 & 55.04 & 52.66 & \textbf{56.40} \\
Telugu (te) & 2.26 & 36.03 & 52.82 & 40.72 & \textbf{58.32} & 49.32 & 54.71 \\
Urdu (ur) & 2.70 & 26.11 & 56.37 & 52.80 & 56.70 & 51.00 & \textbf{56.75} \\
Vietnamese (vi) & 21.47 & 44.33 & 59.68 & 56.61 & 56.67 & 59.85 & \textbf{64.53} \\
\midrule
\multicolumn{8}{c}{\emph{Semantic textual similarity on STS22~\citep{madabushi2022semeval} scored using spearman correlation of cosine similarities}} \\
\midrule
Arabic (ar) & 29.60 & 42.57 & 52.19 & 46.20 & 57.67 & 48.64 & \textbf{58.67} \\
English (en) & 64.32 & 39.76 & 63.06 & 61.72 & 60.97 & 61.45 & \textbf{66.13} \\
Spanish (es) & 58.16 & 54.92 & 59.91 & 56.56 & 63.18 & 61.81 & \textbf{65.41} \\
French (fr) & 77.49 & 58.61 & 74.30 & 70.55 & 77.95 & 73.18 & \textbf{80.38} \\
Chinese (zh) & 33.55 & 49.41 & 61.75 & 58.75 & 63.02 & 58.53 & \textbf{66.78} \\
\bottomrule
\end{tabular}}
\caption{Performance of BLOOM models finetuned for sentence embeddings on classification and STS datasets from MTEB~\citep{muennighoff2022crosslingual}.}
\label{tab:bloom:emb}
\end{table*}

In Section \ref{sec:training}, we have outlined the contrastive finetuning procedure for creating SGPT-BLOOM text embedding models. In Table \ref{tab:bloom:emb}, we report benchmarking results on two multilingual datasets from the Massive Text Embedding Benchmark (MTEB,~\citealp{muennighoff2022mteb}). We find that SGPT-BLOOM-7.1B-msmarco\footnote{\href{https://hf.co/bigscience/sgpt-bloom-7b1-msmarco}{\texttt{hf.co/bigscience/sgpt-bloom-7b1-msmarco}}} provides state-of-the-art performance on several classification and semantic textual similarity splits. However, with 7.1 billion parameters it is an order of magnitude larger than models like the displayed multilingual MiniLM\footnote{\href{https://hf.co/sentence-transformers/paraphrase-multilingual-MiniLM-L12-v2}{\texttt{hf.co/sentence-transformers/paraphrase-multilingual-MiniLM-L12-v2}}} and MPNet\footnote{\href{https://hf.co/sentence-transformers/paraphrase-multilingual-mpnet-base-v2}{\texttt{hf.co/sentence-transformers/paraphrase-multilingual-mpnet-base-v2}}}. SGPT-BLOOM-1.7B-nli\footnote{\href{https://hf.co/bigscience-data/sgpt-bloom-1b7-nli}{\texttt{hf.co/bigscience/sgpt-bloom-1b7-nli}}} performs significantly worse, likely due to less parameters and its finetuning being shorter (NLI is a much smaller dataset than MS-MARCO). Apart from the BLOOM models, ST5-XL\footnote{\href{https://hf.co/sentence-transformers/sentence-t5-xl}{\texttt{hf.co/sentence-transformers/sentence-t5-xl}}} is the largest model with 1.2 billion parameters. However, as an English-only model its performance on non-English languages is poor. The languages displayed are part of the BLOOM pretraining corpus. Performance on more languages and datasets can be inspected on the MTEB leaderboard\footnote{\href{https://hf.co/spaces/mteb/leaderboard}{\texttt{hf.co/spaces/mteb/leaderboard}}}.


\subsection{Multilingual Probing}
\label{sec:eval:probing}
Probing has emerged as a significant evaluation paradigm to analyze and interpret the inner workings of LLMs~\citep{ettinger-etal-2016-probing,adi:2017:ICLR,belinkov-etal-2017-neural,hupkes2018visualisation,tenney2018you,belinkov-glass-2019-analysis,teehan-etal-2022-emergent}, although it comes with certain shortcomings \citep{belinkov-2022-probing}. Examination of the LLM embeddings can help shed light on the generalizing abilities of the model apart from its training objective loss or downstream task evaluation, which is especially beneficial for examining languages lacking annotated datasets or benchmarks.

\subsubsection{Method}
\label{subsec:eval:probing}
For interpreting BLOOM's multilingual generalizing abilities, we utilize the ``Universal Probing'' framework\footnote{\href{https://github.com/bigscience-workshop/massive-probing-framework}{\texttt{github.com/bigscience-workshop/massive-probing-framework}}} for systematic probing analysis in $104$ languages and $80$ morphosyntactic features~\citep{serikov2022universal}. The framework provides SentEval-style~\citep{conneau-etal-2018-cram} probing setup and datasets for each language available in Universal Dependencies (UD;~\citealp{nivre-etal-2016-universal}). We consider the following $17$ languages from $7$ language families present in BLOOM's pretraining corpus (\Cref{sec:dataset}) and UD treebanks: Arabic (Afro-Asiatic), Bambara (Mande), Basque (language isolate), Bengali, Catalan, English, French, Hindi, Marathi, Portuguese, Spanish, Urdu (Indo-European), Chinese (Sino-Tibetan), Indonesian (Austronesian), Tamil (Dravidian), Wolof, Yoruba (Niger-Congo). Our setup covers $38$ morphosyntactic features in total, which represent language-specific linguistic information. We provide a dataset sample in~\autoref{tab:probing_task_ex}.

\begin{table}[ht!]
    \centering
    \resizebox{\textwidth}{!}{%
    \begin{tabular}{llp{0.8\textwidth}}
    \toprule
        Language & Label & Sentence \\
    \midrule
        English & Sing & The \textbf{scheme} makes money through sponsorship and advertising . \\
        & Plur & Still , there are \textbf{questions} left unanswered .\\
        Spanish & Sing & Eligi\textbf{o} no ir tras un tercer período en el siguiente ciclo de elecciones .\\
        & Plur & Todavía quedan \textbf{preguntas} sin responder .\\
    \bottomrule
    \end{tabular}}
    \caption{Examples of the Number task in English and Spanish. The subject number indicator is highlighted in \textbf{bold}. The task is to predict if the sentence includes a singular subject number (upper sentence) and a plural subject number (bottom sentence).}
    \label{tab:probing_task_ex}
\end{table}

The probing procedure is conducted as follows. First, we compute \texttt{<s>}-pooled representations of the input sentence at each layer of the 1.7B-parameter BLOOM variant (``BLOOM 1B7'') and BLOOM (with 176B parameters). Second, we train a binary logistic regression classifier to predict a presence of a morphosyntactic feature in the sentence. Logistic regression is chosen due to its higher selectivity as opposed to non-linear probing classifiers~\citep{hewitt-liang-2019-designing}. We use the original UD training, validation, and test splits here. Third, the probing performance is evaluated by $F_1$ weighted score due to target class imbalance for most probing tasks. The results are averaged across three runs with different random seeds.

\paragraph{Baselines} We compare the probing performance with random guessing and logistic regression classifiers trained on the following TF-IDF features~\citep{salton1973specification}: word unigrams, character N-grams, BPE\footnote{BertTokenizer: \href{https://hf.co/bert-base-multilingual-cased}{\texttt{hf.co/bert-base-multilingual-cased}}} token N-grams, and  SentencePiece\footnote{XLMRobertaTokenizer: \href{https://hf.co/xlm-roberta-base}{\texttt{hf.co/xlm-roberta-base}}} (SP;~\citealp{kudo-richardson-2018-sentencepiece}) token N-grams. We use the N-gram range $\in [1; 4]$ and limit the TF-IDF vocabularies to top-$250$k features. 

\paragraph{Correlation} We run statistical tests to analyze correlations between the probing performance and linguistic, dataset, and model configuration criteria:
\begin{itemize}
    \item Language script: the results are divided into two groups by the language script -- Latin and others (Devanagari, Tamil, and Arabic). Here, we use the non-parametric test Mann-Whitney U~\citep{mann1947controlling}.
    \item Language family: the results are divided into $7$ groups by the language family. We apply the ANOVA to analyze the variance between the groups.
    \item Probing and pretraining dataset size: we run the Pearson correlation coefficient test \citep{pearson1895vii} to compute correlation between the probing performance and these data configuration criteria.
    \item Effect of the model size: the results are divided into two groups by the BLOOM version. Here, we use the Mann-Whitney U test to see if there is a correlation between the number of parameters and the probing results.
\end{itemize}

\begin{table*}[t!]
\centering
\scriptsize
\resizebox{\textwidth}{!}{%
\begin{tabular}{lccccccc}
\toprule
{} &               BLOOM-1B7 &                   BLOOM &    Random &        TF-IDF (Char) &           TF-IDF (Word) &            TF-IDF (BPE) &             TF-IDF (SP) \\
\midrule
Arabic     &  \textbf{0.66} \tiny{$\pm$0.27} &  0.64 \tiny{$\pm$0.27} & 0.49 \tiny{$\pm$0.013} & 0.41 \tiny{$\pm$0.44} &   0.4 \tiny{$\pm$0.44} &  0.41 \tiny{$\pm$0.44} &  0.41 \tiny{$\pm$0.44} \\
Bambara    &  \textbf{0.64} \tiny{$\pm$0.16} &  0.59 \tiny{$\pm$0.16} &  0.45 \tiny{$\pm$0.1}  & 0.52 \tiny{$\pm$0.46} &  0.45 \tiny{$\pm$0.47} &  0.48 \tiny{$\pm$0.49} &  0.49 \tiny{$\pm$0.49} \\
Basque     &  \textbf{0.68} \tiny{$\pm$0.19} & 0.62 \tiny{$\pm$0.19} & 0.49 \tiny{$\pm$0.03}  &  0.41 \tiny{$\pm$0.43} &  0.44 \tiny{$\pm$0.46} &  0.48 \tiny{$\pm$0.44} &  0.41 \tiny{$\pm$0.46} \\
Bengali    &  0.42 \tiny{$\pm$0.15} &  0.45 \tiny{$\pm$0.12} & 0.35 \tiny{$\pm$0.23}  & 0.63 \tiny{$\pm$0.48} &  0.37 \tiny{$\pm$0.44} &  0.41 \tiny{$\pm$0.32} &  \textbf{0.76} \tiny{$\pm$0.28} \\
Catalan    &  \textbf{0.65} \tiny{$\pm$0.25} &  0.61 \tiny{$\pm$0.26} & 0.34 \tiny{$\pm$0.01}  & 0.24 \tiny{$\pm$0.38} &  0.24 \tiny{$\pm$0.39} &  0.24 \tiny{$\pm$0.39} &  0.24 \tiny{$\pm$0.39} \\
Chinese    &  \textbf{0.66} \tiny{$\pm$0.25} &  0.50 \tiny{$\pm$0.21} & 0.55 \tiny{$\pm$0.03}  & 0.03 \tiny{$\pm$0.05} & 0.11 \tiny{$\pm$0.28} &  0.04 \tiny{$\pm$0.06} &  0.03 \tiny{$\pm$0.05} \\
English    &  \textbf{0.57} \tiny{$\pm$0.24} &  \textbf{0.57} \tiny{$\pm$0.24} & 0.43 \tiny{$\pm$0.03} & 0.45 \tiny{$\pm$0.43} &  0.46 \tiny{$\pm$0.43} &  0.45 \tiny{$\pm$0.43} &  0.44 \tiny{$\pm$0.44} \\
French     &  \textbf{0.61} \tiny{$\pm$0.23} &  0.57 \tiny{$\pm$0.22} & 0.44 \tiny{$\pm$0.02} & 0.32 \tiny{$\pm$0.43} &  0.32 \tiny{$\pm$0.43} &  0.32 \tiny{$\pm$0.43} &  0.33 \tiny{$\pm$0.44} \\
Hindi      & \textbf{0.63} \tiny{$\pm$0.23} &   0.6 \tiny{$\pm$0.25} & 0.48 \tiny{$\pm$0.03}   & 0.53 \tiny{$\pm$0.46} &  0.55 \tiny{$\pm$0.47} &  0.53 \tiny{$\pm$0.46} &  0.53 \tiny{$\pm$0.46} \\
Indonesian &  \textbf{0.65} \tiny{$\pm$0.27} &   0.6 \tiny{$\pm$0.27} & 0.48 \tiny{$\pm$0.05} & 0.41 \tiny{$\pm$0.46} &  0.43 \tiny{$\pm$0.45} &  0.41 \tiny{$\pm$0.46} &  0.45 \tiny{$\pm$0.45} \\
Marathi    &  \textbf{0.57} \tiny{$\pm$0.25} &  0.48 \tiny{$\pm$0.24} &  0.32 \tiny{$\pm$0.09}  & 0.44 \tiny{$\pm$0.47} &  0.46 \tiny{$\pm$0.46} &  0.44 \tiny{$\pm$0.47} &  0.44 \tiny{$\pm$0.47} \\
Portugese  &  \textbf{0.67} \tiny{$\pm$0.23} &  0.63 \tiny{$\pm$0.26} & 0.4 \tiny{$\pm$0.03}  &  0.48 \tiny{$\pm$0.48} &  0.49 \tiny{$\pm$0.48} &  0.48 \tiny{$\pm$0.48} &  0.48 \tiny{$\pm$0.48} \\
Spanish    &  \textbf{0.66} \tiny{$\pm$0.24} &  0.65 \tiny{$\pm$0.24} & 0.42 \tiny{$\pm$0.02} & 0.35 \tiny{$\pm$0.42} &  0.35 \tiny{$\pm$0.44} &  0.36 \tiny{$\pm$0.44} &  0.36 \tiny{$\pm$0.43} \\
Tamil      &  \textbf{0.57} \tiny{$\pm$0.25} &  0.51 \tiny{$\pm$0.27} & 0.43 \tiny{$\pm$0.05} & 0.51 \tiny{$\pm$0.44} &  0.53 \tiny{$\pm$0.44} &   0.5 \tiny{$\pm$0.44} &   0.5 \tiny{$\pm$0.44} \\
Urdu       &  \textbf{0.75} \tiny{$\pm$0.21} &   0.70 \tiny{$\pm$0.24} & 0.43 \tiny{$\pm$0.02} & 0.39 \tiny{$\pm$0.48} &  0.39 \tiny{$\pm$0.47} &  0.39 \tiny{$\pm$0.48} &  0.39 \tiny{$\pm$0.48} \\
Wolof      & \textbf{0.51} \tiny{$\pm$0.32} &  0.47 \tiny{$\pm$0.32} & 0.41 \tiny{$\pm$0.02} & 0.26 \tiny{$\pm$0.39} &  0.25 \tiny{$\pm$0.39} &   0.3 \tiny{$\pm$0.43} &  0.27 \tiny{$\pm$0.39} \\
Yoruba     &  \textbf{0.48} \tiny{$\pm$0.07} &  0.36 \tiny{$\pm$0.07} & 0.43 \tiny{$\pm$0.06} & 0.33 \tiny{$\pm$0.45} &  0.09 \tiny{$\pm$0.05} &  0.16 \tiny{$\pm$0.11} &  0.09 \tiny{$\pm$0.05} \\
\bottomrule
\end{tabular}}
\caption{Probing performance ($F_1$ averaged by layers) of the BLOOM-based classifiers and count-based baselines. The results are averaged over probing tasks, and three experiment runs within each language. Standard deviation is determined by the results along the language tasks.}
\label{tab:bloom:probing}
\end{table*}
\subsubsection{Results}

\begin{figure*}[t!]
    \centering
    \begin{subfigure}[b]{0.49\linewidth}
    \centering
    \includegraphics[width=\linewidth]{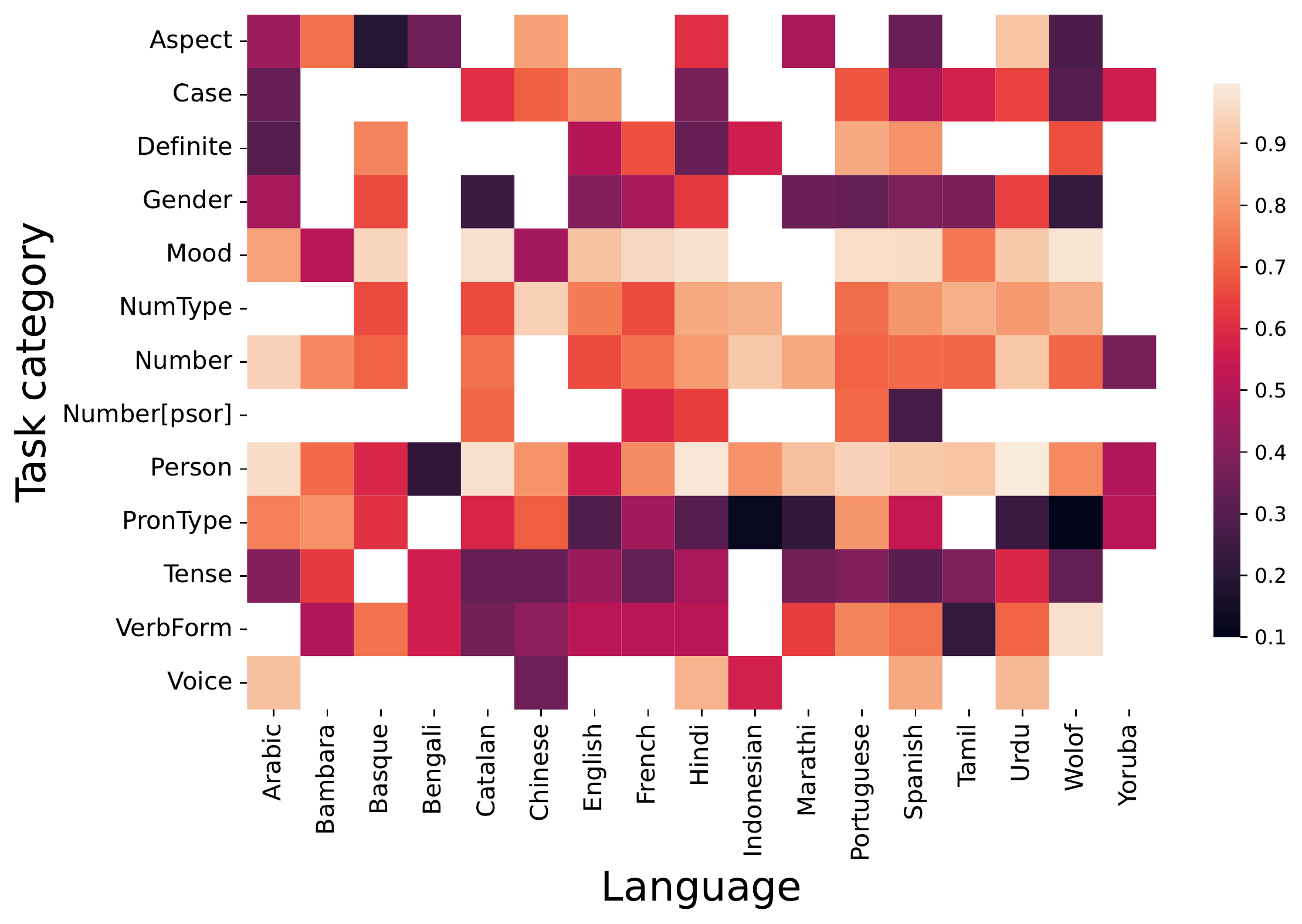}
    \caption{BLOOM-1B7}
    \label{fig:1b7-probing}
    \end{subfigure}
    \begin{subfigure}[b]{0.49\linewidth}
    \centering
    \includegraphics[width=\linewidth]{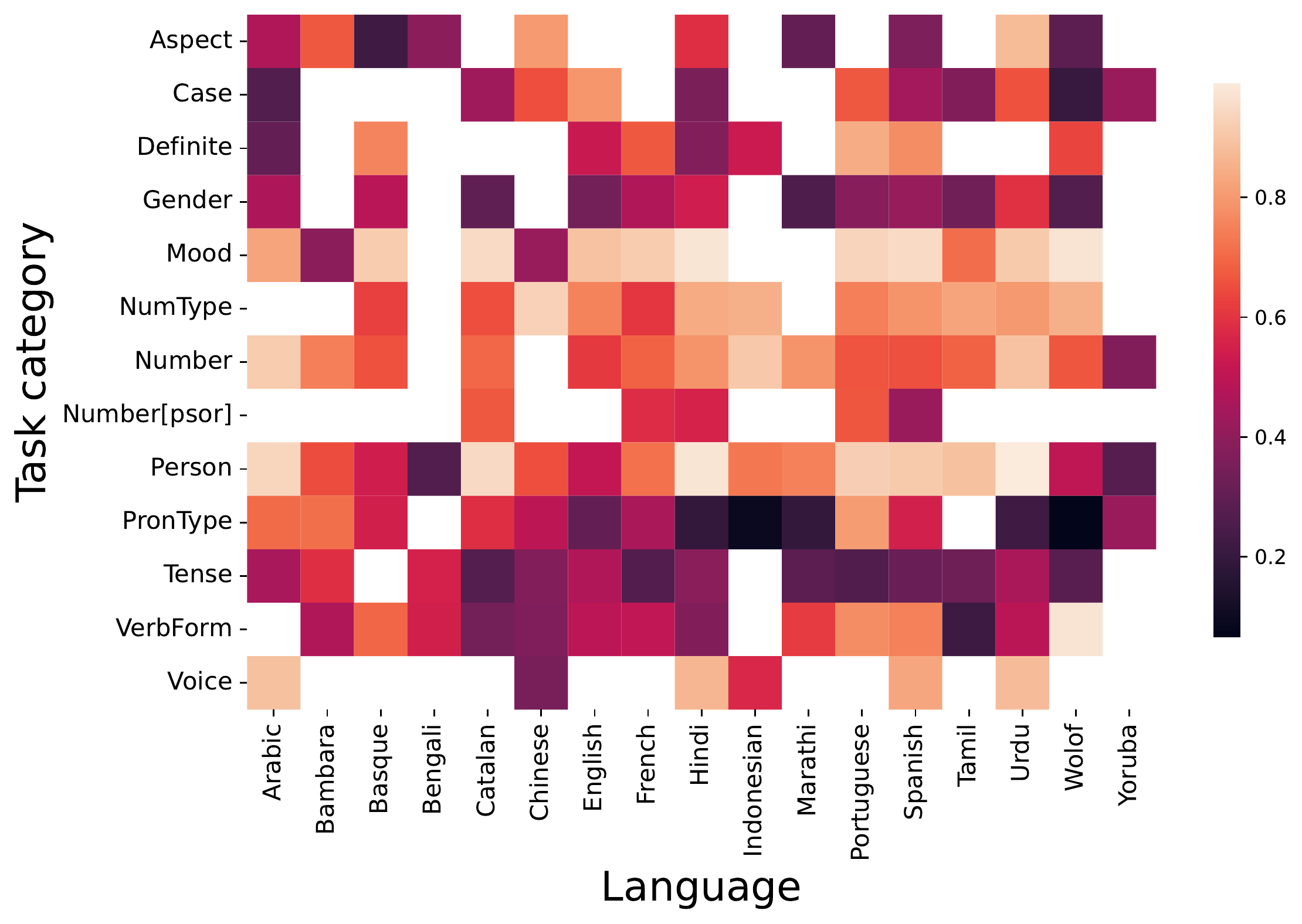}
    \caption{BLOOM}
    \label{fig:bloom-probing}
    \end{subfigure}    
    \caption{Probing classifiers' results by language and task category. White squares denote that the morphosyntactic category is not represented in the language.
    }
    \label{fig:probing_results}
\end{figure*}

\paragraph{Probing} ~\autoref{tab:bloom:probing} presents the results of probing experiments averaged over the probing tasks and experiment runs within each language. The overall pattern is that BLOOM-1B7 performs on par or better than BLOOM, and both LLMs outperform the count-based baselines. In particular, the LLMs achieve more robust performance on Arabic, Basque, and Indo-European languages (e.g., Catalan, French, Hindi, Portuguese, Spanish, and Urdu), while Bengali, Wolof, and Yoruba receive the lowest scores. We attribute this behavior to the transfer abilities: BLOOM infers linguistic properties better for the closely related languages that comprise a significant amount of data. For example, the performance on any Romance language is better than in English, and the results in Indic languages are close to those in high-resource languages.

~\autoref{fig:probing_results} presents the language-wise probing performance results for morphosyntactic features represented at least in $5$ languages. The probing performance of both LLMs is similar despite the difference in size. We find that the LLMs infer Mood and Person well with no regard for language. Number, NumType (numeral type), and Voice are moderately inferred in most languages. The models generally show worse qualities in the other categories, indicating that they do not encode such morphological information. The possible explanation of such difference in performance may be the diversity of possible values of these categories. For example, Mood and Person share similar values across the presented languages, while the set of Case values is highly dependent on the language.

\paragraph{Correlation}
\begin{table*}[t!]
\centering
\small
\begin{tabular}{cccc}
\toprule
Criterion & Model & Test & p-value \\
\midrule
\multirow{2}{*}{Language script} & \begin{tabular}{@{}c@{}} BLOOM \\ BLOOM-1B7\end{tabular} & \multirow{2}{*}{Mann-Whitney U} & \begin{tabular}{@{}c@{}}0.72 \\ 0.13\end{tabular}  \\
\midrule
\multirow{2}{*}{Language family} & \begin{tabular}{@{}c@{}} BLOOM \\ BLOOM-1B7\end{tabular} & \multirow{2}{*}{ANOVA} & \begin{tabular}{@{}c@{}}<0.01 \\ <0.01\end{tabular}  \\
\midrule
\multirow{2}{*}{Probing dataset size} &  \begin{tabular}{@{}c@{}} BLOOM \\ BLOOM-1B7 \end{tabular} & \multirow{2}{*}{Pearson} & \begin{tabular}{@{}c@{}}0.63 \\ 0.02\end{tabular}  \\
\midrule
\multirow{2}{*}{Pretraining dataset size} & \begin{tabular}{@{}c@{}} BLOOM \\ BLOOM-1B7 \end{tabular} & \multirow{2}{*}{Pearson} & \begin{tabular}{@{}c@{}}0.46 \\ <0.01\end{tabular}  \\
\midrule
Difference between versions & BLOOM \& BLOOM-1B7 & Mann-Whitney U & <0.01 \\

\bottomrule
\end{tabular}
\caption{Results of statistical tests and correlation analysis between probing performance and linguistic, dataset, and model configuration criteria.}
\label{tab:correlation_probing}
\end{table*}

The correlation analysis results support conclusions on the probing performance and reveals contributing factors (see~\autoref{tab:correlation_probing}). Both models show similar results on the languages with different language scripts. Results of BLOOM-1B7 are highly correlated with language family, probing dataset size, and pretraining dataset size. According to the results of Mann-Whithey U test, BLOOM-1B7 shows significantly better results ($p < 0.01$) than BLOOM. However, BLOOM shows more stable performance on different languages in spite of the amount of data it has seen during pretraining. This might indicate the better generalization abilities of the model with more parameters. 

\paragraph{Discussion}
It should be noted that the following questions remain for further research:
\begin{enumerate}
    \item \textbf{Generalizing abilities.} BLOOM-1B7 is leading in the average performance of morphosyntactic feature classification for the languages in~\autoref{tab:bloom:probing}. The BLOOM results are lower, which can be interpreted as a worse grammatical generalization over the aforecited languages. However, the BLOOM-1B7's probing correlation results with factors like pretraining dataset size are more prominent, which makes it potentially less generalizing on the under-resourced languages than the bigger version.
    \item \textbf{Multilingual abilities.}
    A separate research interest implies considering languages that are not explicitly included in the pretraining corpus of the models. Expanding the set of languages for probing will allow for a typological interpretation and a deeper analysis of the most learnable and hard-to-learn linguistic features on a more considerable scope.
    \item \textbf{Under-resourced language evaluation.} The under-resourced languages of the Indic and Niger-Congo families included in the pretraining corpus in smaller shares represent a separate subject for future probing. We also plan to investigate the results of high-resourced and under-resourced languages to reveal possible linguistic insights in these two groups.
    \item \textbf{Different layers and training dynamics.} The analysis has focused on averaged representations of all layers and at the end of training. Analyzing different layers may reveal how morpho-syntactic representations are built during processing. Similarly, investigating how properties are acquired over the course of pre-training  \citep{choshen-etal-2022-grammar,zhang-etal-2021-need,voloshina2022neural} is a viable direction for research. 
\end{enumerate}

\subsection{Bias}


As a preliminary study into the biases learned by BLOOM, we present evaluation on the \texttt{multilingual CrowS-Pairs} dataset, which combines a revised version of the CrowS-Pairs dataset developed by~\cite{nangia-etal-2020-crows} together with the French version of CrowS-Pairs introduced by~\cite{neveol-etal-2022-french}. 
One challenge of this evaluation was to adapt a dataset originally intended for masked language models to autoregressive language models such as BLOOM. CrowS-Pairs relies on minimal pairs to compare a stereotyped statement and a non-stereotyped statement (e.g. ``\textit{Women} can't drive.'' is a gender stereotype while ``\textit{Men} can't drive'' is not). The two statements differ only by the social category targeted by the stereotype and that social category is present in the stereotyped statement and not in the non-stereotyped statement. The evaluation aims at assessing systematic preference of models for stereotyped statements. The original ``metric score'' compared pseudo-log-likelihood of sentences in a pair to determine which sentence received a higher score from a masked language model. Prompts were designed to require the model to select one of the statements based on the ``likely'' and ``realistic'' nature of the situations described.

\begin{figure}[!ht]
\centering
\includegraphics[width=\textwidth]{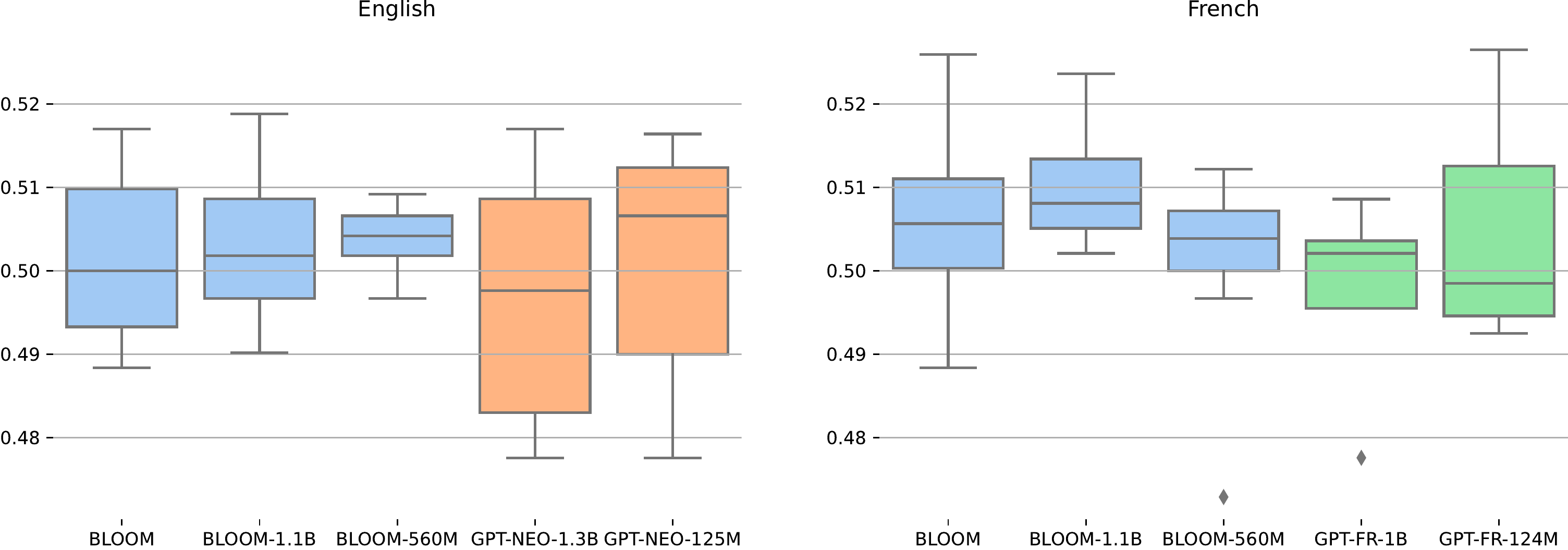}
\caption{Overall accuracy of BLOOM on \texttt{crowS-Pairs} per prompt for English and French. Results on the two smallest BLOOM models and monolingual GPT models of comparable size are also shown.}
\label{fig:crowspairs_enfr}
\end{figure}

Figure \ref{fig:crowspairs_enfr} shows that BLOOM's overall prompt accuracy was close to .50, which suggests an overall absence of bias. We note that the scores in English and French are very close, suggesting similar overall behavior of the model on both languages.
We also show results on mono-lingual autoregressive models --- GPT-Neo~\citep{black2021gpt} and GPT-FR~\citep{simoulin:hal-03265900} for English and French, respectively.



Table \ref{tab:crowspairs_disaggregated} presents the results per bias type in the \texttt{CrowS-Pairs} dataset. The results are quite homogeneous over the categories, which contrasts with previous studies on masked language models, which suggested models were prone to bias in specific categories, which differed between models tested. Nonetheless, accuracy significantly differs from 50 (T-test, p < .05) overall for both languages, as well as for a number of bias categories, as shown per asterisks in the table.  




 \begin{table}
 \centering
 \begin{tabular}{lll|l}
 \toprule
 Bias type              & support &English & French \\
 \midrule
 ethnicity \/ color     & 460   & 50.05     & 50.48* \\
 gender                 & 321   & 51.17*    & 51.24* \\
 socioeconomic status   &196    & 51.05*    & 52.22*\\
 nationality            &253    & 49.25*    & 48.49* \\
 religion               &115    & 53.82*     & 53.01*\\
 age                    &90     & 49.35     & 50.13 \\
 sexual orientation     &91     & 50.00     & 49.9 \\
 physical appearance    &72     & 48.20     & 49.67\\
 disability             &66     & 48.49*    & 49.16* \\
 other                  & 13    & 50.18     & 42.1* \\
 \midrule
 All & 1,677 & 49.78* & 50.61* \\
 \bottomrule
 \end{tabular}
 \caption{BLOOM accuracy results on \texttt{crowS-Pairs} bias categories averaged over eight runs for English and French. Significance for the one sample T-test ($p<.05$) is indicated with *.}
 \label{tab:crowspairs_disaggregated}
 \end{table}

\paragraph{Limitations}
\cite{blodgett-etal-2021-stereotyping} discuss validity issues with the original CrowS-Pairs corpus. The CrowS-Pairs version used here differs from the original by addressing some of the issues pointed out by \citet{blodgett-etal-2021-stereotyping} and by constructing $~200$ additional sentence pairs based on stereotypes collected from French speakers.
In a recent evaluation of bias in masked language models in English and French, results obtained on the revised dataset were not significantly different from those obtained on the original dataset~\cite{neveol-etal-2022-french}. However, its original validation does not naturally apply here, and comparison to other CrowS-Pairs results is more difficult. 
For a stronger assessment of bias, results obtained with CrowS-Pairs should be compared with other measures of bias, and also assessed for all languages in the model. However, as noted by \cite{talat2022you}, very little material (corpora, measures) is available for multilingual bias assessment. 

Although our examinations suggest a limited presence of bias in the model, they cannot cover the breadth of possible usage scenarios. One such scenario where models may have a larger impact is on linguistic diversity and language variation encountered.
As the training resources for BLOOM are carefully curated, they may also capture some language variations to a larger degree than other models. This also impacts the ability of trained models to equitably represent different variations. Such differences can aid in the propagation and legitimization of some language variants over others.
Our evaluation of biases in the model are further limited to the situations, languages and language variants that are covered by \texttt{multilingual CrowS-Pairs}.
We therefore expect a distinction between our findings using CrowS-Pairs and wider model use \citep[for a more detailed exploration on such differences, see][]{raji_AI_2021}.

\section{Conclusion}
In this work, we present BLOOM, a 176B-parameter open-access multilingual language model.
BLOOM was created by BigScience, a collaboration of hundreds of researchers, and was trained on the French government-funded Jean Zay supercomputer for 3.5 months.
In this paper, we chronicled the development of BLOOM, from the creation of its training dataset ROOTS to the design of its architecture and tokenizer.
We also discuss evaluation results of BLOOM and other large language models, finding it has competitive performance that improves after multitask finetuning.

We hope that the release of a powerful multilingual language model unlocks new applications and research directions for large language models.
Further, we hope that documenting our experience will help the machine learning research community organize new large-scale collaborative projects similar to BigScience. Besides enabling results that are impossible for any individual research group to achieve, this form of organization will also allow more people with different backgrounds to share their ideas and participate in the development of major advances in the field.

\section{Contributions}
\label{sec:contributions}

Authors are assigned to each authorship category according to which aspects of the project they contributed to.
Many authors appear under multiple categories because they contributed to the project in more than one way.
Author order in all categories is alphabetical by first name, except for ``Major Contributors'' where authors are shuffled randomly apart from Teven Le Scao, who is intentionally listed first and ``Organization'' where Thomas Wolf is intentionally listed last.
A description of each category follows.
For finer-grained contribution details, please see the papers mentioned under each category.
\begin{description}
\item[Major Contributors] lists individuals without whom BLOOM would not have happened and/or who spent more than 20\% of their time on the BigScience effort as a whole.
\item[Dataset] lists individuals who contributed to data sourcing, organization, and processing efforts, including the authors of \citet{laurencon2022bigscience}, \citet{mcmillan-major2022documenting}, and \citet{jernite2022data}.
\item[Tokenization] lists individuals who built the BLOOM tokenizer and authors of \citet{mielke2021between}.
\item[Prompt Engineering] lists individuals who wrote, edited, and reviewed prompt templates for the datasets we consider as well as authors of \citet{sanh2022multitask}, \citet{bach2022promptsource}, and \citet{muennighoff2022crosslingual}.
\item[Architecture and Objective] lists individuals who ran experiments to help determine BLOOM's model architecture and training objective, including authors of \citet{wang2022what} and \citet{scao2022what}.
\item[Engineering] lists individuals who contributed to code and infrastructure to  train BLOOM on the Jean Zay supercomputer.
\item[Evaluation and interpretability] lists individuals who helped evaluate the BLOOM model as well as authors of \citet{talat2022you}.
\item[Broader Impacts] lists authors of the ethical charter, license, and model card, in addition to individuals who studied privacy issues, social impacts, and BLOOM's carbon footprint.
\item[Applications] lists members of working groups focused on applications of BLOOM, including authors of \citet{bigbio}, \citet{fries2022dataset}, and \citet{de-toni-etal-2022-entities}.
\item[Organization] lists individuals who coordinated the BigScience effort and authors of \citet{bigscience2022org}.
\end{description}

\begin{acks}
The BigScience Workshop was granted access to the HPC resources of the Institut du développement et des ressources en informatique scientifique (IDRIS) du Centre national de la recherche scientifique (CNRS) under the allocation 2021-A0101012475 made by the Grand équipement national de calcul intensif (GENCI). Model training ran on the Jean-Zay supercomputer of GENCI at IDRIS, and we thank the IDRIS team for their responsive support throughout the project, in particular Rémi Lacroix.

Roman Castagné, Thomas Wang, Beno\^\i t Sagot and Rachel Bawden's contributions were funded by Beno\^\i t Sagot's and Rachel Bawden's chairs in the PRAIRIE institute funded by the French national agency ANR as part of the ``Investissements d’avenir'' programme under the reference ANR-19-P3IA-0001.
Aur\'elie N\'ev\'eol's contribution was supported by ANR under grant GEM
ANR-19-CE38-0012. 
Oskar van der Wal's contributions were financed by the Dutch Research Council (NWO) as part of Open Competition Digitalisation-SSH with project number 406.DI.19.059.

The BigScience Workshop would also like to acknowledge the support and financing of the following organizations, organization members and affiliations of some of the participants: ESPCI and LAMSADE (Dauphine Universit\'e, PSL, CNRS) for Alexandre Allauzen; MELODI team at IRIT/University of Toulouse for Farah Benamara, Chlo\'e Braud, Philippe Muller, and V\'eronique Moriceau; IRISA LinkMedia team IMATAG/CNRS for Vincent Claveau and Antoine Chaffin; Universit\'e de Lorraine ATILF UMR 7118 CNRS / UL for Mathieu Constant; University of Paris for Beno\^\i t Crabb\'e, Marie Candito and Antoine Simoulin; GdR TAL (CNRS) for B\'eatrice Daille; CNRS DR1 INSERM UMR1093 UBFC Dijon for Peter Ford Dominey; Aix-Marseille University UTLN CNRS LIS/UMR7220 for Beno\^\i t Favre and Fr\'ed\'eric B\'echet; CEA LASTI for Bertrand Delezoide, Olivier Ferret, Adrian Popescu and Julien Tourille; Sorbonne Universit\'e LORIA for Karen Fort; CNRS DR1 LORIA UMR7503 Nancy for Claire Gardent and Christophe Cerisara; MAS Laboratory of Ecole Centrale Paris for C\'eline Hudelot, RCLN/LIPN UMR 7030 University Sorbonne-Paris-Nord/CNRS for Joseph Le Roux and Nadi Tomeh, Universit\'e de Paris and Necker - Enfants Malades hospital for Antoine Neuraz and Ivan Lerner, Universit\'e Paris Saclay LISN CNRS UMR9105 for Aur\'elie N\'ev\'eol, Anne-Laure Ligozat, Caio Corro, Francois Yvon; Inria, Univ. Bordeaux and Ensta ParisTech for Pierre-Yves Oudeyer, C\'edric Colas, Grgur Kovac, Tristan Karch; Inria Paris for Beno\^\i t Sagot, Djam\'e Seddah, Pedro Ortiz; University Toulouse CNRS for Ludovic Tanguy, Sorbonne Universit\'e, LIMICS (Sorbonne Universit\'e, Inserm, Univ. Sorbonne Paris Nord) for Xavier Tannier; I3S Laboratory, CNRS, INRIA, Universit\'e Cote d’Azur for Serena Villata and Elena Cabrio; Airbus, Central Research \& Technology for Guillaume Alleon, Alexandre Arnold, and Catherine Kobus; Cloud Temple for Jean-Michel Dussoux; Illuin Technology for Robert Vesoul, Gautier Viaud, Martin d’Hoffschmidt, and Wacim Belblidia; Levia.ai for 
Romain Riviere; LightOn for Igor Carron, Laurent Daudet, Iacopo Poli, and Julien Launay; Nabla for Alexandre Lebrun, Martin Raison, and Samuel Humeau; Naver Labs Europe for Matthias Gall\'e and Laurent Besacier; Orange Labs for G\'eraldine Damnati, Johannes Heinecke, and Frederic Herledan; OVHcloud for Jean-Louis Queguiner and Guillaume Salou; ReciTAL for Thomas Scialom, Gilles Moyse, and Jacopo Staiano; Renault Group for Vincent Feuillard, Joan Andr\'e, Francois-Paul Servant, Raphael Sourty, and Ayhan Uyanik; SYSTRAN for Jean Senellart, Josep Crego, Elise Michon, Guillaume Klein, Dakun Zhang, and Natalia Segal; Ubisoft for Guillaume Gaudron. Leipzig University and the Center for Scalable Data Analytics and Artificial Intelligence (ScaDS.AI) in Leipzig for Christopher Akiki.

Hugging Face provided storage for the entirety of the project, as well as compute for development and part of training the smaller BLOOM models. Many of the evaluations in this paper were made possible by compute resources donated by CoreWeave and EleutherAI.
\end{acks}

\vskip 0.2in
\bibliography{sample}

\begin{thebibliography}{171}
\providecommand{\natexlab}[1]{#1}
\providecommand{\url}[1]{\texttt{#1}}
\expandafter\ifx\csname urlstyle\endcsname\relax
  \providecommand{\doi}[1]{doi: #1}\else
  \providecommand{\doi}{doi: \begingroup \urlstyle{rm}\Url}\fi

\bibitem[Abadji et~al.(2021)Abadji, Su{\'a}rez, Romary, and
  Sagot]{AbadjiOrtizSuarezRomaryetal2021}
Julien Abadji, Pedro Javier~Ortiz Su{\'a}rez, Laurent Romary, and Beno{\^\i}t
  Sagot.
\newblock Ungoliant: An optimized pipeline for the generation of a very
  large-scale multilingual web corpus.
\newblock In Harald L{\"u}ngen, Marc Kupietz, Piotr Bański, Adrien Barbaresi,
  Simon Clematide, and Ines Pisetta, editors, \emph{Proceedings of the Workshop
  on Challenges in the Management of Large Corpora (CMLC-9)}, pages 1--9,
  Limerick, Ireland, 2021. Leibniz-Institut f{\"u}r Deutsche Sprache.
\newblock \doi{10.14618/ids-pub-10468}.
\newblock URL \url{https://nbn-resolving.org/urn:nbn:de:bsz:mh39-104688}.

\bibitem[{\' A}cs(2019)]{acs2019exploring}
Judit {\' A}cs.
\newblock Exploring bert's vocabulary, 2019.
\newblock URL
  \url{http://juditacs.github.io/2019/02/19/bert-tokenization-stats.html}.

\bibitem[Adi et~al.(2017)Adi, Kermany, Belinkov, Lavi, and
  Goldberg]{adi:2017:ICLR}
Yossi Adi, Einat Kermany, Yonatan Belinkov, Ofer Lavi, and Yoav Goldberg.
\newblock Fine-grained analysis of sentence embeddings using auxiliary
  prediction tasks.
\newblock In \emph{International Conference on Learning Representations
  (ICLR)}, April 2017.

\bibitem[Akiki et~al.(2022)Akiki, Pistilli, Mieskes, Gallé, Wolf, Ilić, and
  Jernite]{bigscience2022org}
Christopher Akiki, Giada Pistilli, Margot Mieskes, Matthias Gallé, Thomas
  Wolf, Suzana Ilić, and Yacine Jernite.
\newblock {BigScience: A Case Study in the Social Construction of a
  Multilingual Large Language Model}, 2022.
\newblock URL \url{https://arxiv.org/abs/2212.04960}.

\bibitem[Al-Rfou et~al.(2019)Al-Rfou, Choe, Constant, Guo, and
  Jones]{al2019character}
Rami Al-Rfou, Dokook Choe, Noah Constant, Mandy Guo, and Llion Jones.
\newblock Character-level language modeling with deeper self-attention.
\newblock In \emph{Proceedings of the AAAI conference on artificial
  intelligence}, 2019.

\bibitem[Altaher et~al.(2022)Altaher, Fadel, Alotaibi, Alyazidi, Al{-}Mutairi,
  Aldhbuiub, Mosaibah, Rezk, Alhendi, Shal, Alghamdi, AlShaibani, Zakraoui,
  Mohammed, Gaanoun, Elmadani, Ghaleb, Tazi, Alharbi, Masoud, and
  Alyafeai]{alyafeai2022masader}
Yousef Altaher, Ali Fadel, Mazen Alotaibi, Mazen Alyazidi, Mishari
  Al{-}Mutairi, Mutlaq Aldhbuiub, Abdulrahman Mosaibah, Abdelrahman Rezk,
  Abdulrazzaq Alhendi, Mazen~Abo Shal, Emad~A. Alghamdi, Maged~Saeed
  AlShaibani, Jezia Zakraoui, Wafaa Mohammed, Kamel Gaanoun, Khalid~N.
  Elmadani, Mustafa Ghaleb, Nouamane Tazi, Raed Alharbi, Maraim Masoud, and
  Zaid Alyafeai.
\newblock Masader plus: {A} new interface for exploring +500 arabic {NLP}
  datasets.
\newblock \emph{CoRR}, abs/2208.00932, 2022.
\newblock \doi{10.48550/arXiv.2208.00932}.
\newblock URL \url{https://doi.org/10.48550/arXiv.2208.00932}.

\bibitem[Alyafeai et~al.(2021)Alyafeai, Masoud, Ghaleb, and
  AlShaibani]{alyafeai2021masader}
Zaid Alyafeai, Maraim Masoud, Mustafa Ghaleb, and Maged~Saeed AlShaibani.
\newblock Masader: Metadata sourcing for arabic text and speech data resources.
\newblock \emph{CoRR}, abs/2110.06744, 2021.
\newblock URL \url{https://arxiv.org/abs/2110.06744}.

\bibitem[Bach et~al.(2022)Bach, Sanh, Yong, Webson, Raffel, Nayak, Sharma, Kim,
  Bari, Fevry, Alyafeai, Dey, Santilli, Sun, Ben-david, Xu, Chhablani, Wang,
  Fries, Al-shaibani, Sharma, Thakker, Almubarak, Tang, Radev, Jiang, and
  Rush]{bach2022promptsource}
Stephen Bach, Victor Sanh, Zheng~Xin Yong, Albert Webson, Colin Raffel,
  Nihal~V. Nayak, Abheesht Sharma, Taewoon Kim, M~Saiful Bari, Thibault Fevry,
  Zaid Alyafeai, Manan Dey, Andrea Santilli, Zhiqing Sun, Srulik Ben-david,
  Canwen Xu, Gunjan Chhablani, Han Wang, Jason Fries, Maged Al-shaibani, Shanya
  Sharma, Urmish Thakker, Khalid Almubarak, Xiangru Tang, Dragomir Radev, Mike
  Tian-jian Jiang, and Alexander Rush.
\newblock {P}rompt{S}ource: An integrated development environment and
  repository for natural language prompts.
\newblock In \emph{Proceedings of the 60th Annual Meeting of the Association
  for Computational Linguistics: System Demonstrations}, pages 93--104, Dublin,
  Ireland, May 2022. Association for Computational Linguistics.
\newblock \doi{10.18653/v1/2022.acl-demo.9}.
\newblock URL \url{https://aclanthology.org/2022.acl-demo.9}.

\bibitem[Bannour et~al.(2021)Bannour, Ghannay, N{\'e}v{\'e}ol, and
  Ligozat]{bannour-etal-2021-evaluating}
Nesrine Bannour, Sahar Ghannay, Aur{\'e}lie N{\'e}v{\'e}ol, and Anne-Laure
  Ligozat.
\newblock Evaluating the carbon footprint of {NLP} methods: a survey and
  analysis of existing tools.
\newblock In \emph{Proceedings of the Second Workshop on Simple and Efficient
  Natural Language Processing}, pages 11--21, Virtual, November 2021.
  Association for Computational Linguistics.
\newblock \doi{10.18653/v1/2021.sustainlp-1.2}.
\newblock URL \url{https://aclanthology.org/2021.sustainlp-1.2}.

\bibitem[Bawden and Yvon(2023)]{bawden-yvon-2023-mt-bloom}
Rachel Bawden and Fran{\c{c}}ois Yvon.
\newblock Investigating the translation performance of a large multilingual
  language model: the case of {BLOOM}.
\newblock \emph{CoRR}, abs/2303.01911, 2023.
\newblock \doi{10.48550/arXiv.2303.01911}.
\newblock URL \url{https://doi.org/10.48550/arXiv.2303.01911}.

\bibitem[Bawden et~al.(2020)Bawden, Bilinski, Lavergne, and
  Rosset]{bawden-et-al-2020-diabla}
Rachel Bawden, Eric Bilinski, Thomas Lavergne, and Sophie Rosset.
\newblock {DiaBLa: A Corpus of Bilingual Spontaneous Written Dialogues for
  Machine Translation}.
\newblock \emph{Language Resources and Evaluation}, pages 635--660, 2020.
\newblock \doi{10.1007/s10579-020-09514-4}.
\newblock URL \url{https://doi.org/10.1007/s10579-020-09514-4}.

\bibitem[Belinkov(2022)]{belinkov-2022-probing}
Yonatan Belinkov.
\newblock Probing classifiers: Promises, shortcomings, and advances.
\newblock \emph{Computational Linguistics}, 48\penalty0 (1):\penalty0 207--219,
  March 2022.
\newblock \doi{10.1162/coli_a_00422}.
\newblock URL \url{https://aclanthology.org/2022.cl-1.7}.

\bibitem[Belinkov and Glass(2019)]{belinkov-glass-2019-analysis}
Yonatan Belinkov and James Glass.
\newblock Analysis methods in neural language processing: A survey.
\newblock \emph{Transactions of the Association for Computational Linguistics},
  7:\penalty0 49--72, March 2019.
\newblock \doi{10.1162/tacl_a_00254}.
\newblock URL \url{https://www.aclweb.org/anthology/Q19-1004}.

\bibitem[Belinkov et~al.(2017)Belinkov, Durrani, Dalvi, Sajjad, and
  Glass]{belinkov-etal-2017-neural}
Yonatan Belinkov, Nadir Durrani, Fahim Dalvi, Hassan Sajjad, and James Glass.
\newblock What do neural machine translation models learn about morphology?
\newblock In \emph{Proceedings of the 55th Annual Meeting of the Association
  for Computational Linguistics (Volume 1: Long Papers)}, pages 861--872,
  Vancouver, Canada, July 2017. Association for Computational Linguistics.
\newblock \doi{10.18653/v1/P17-1080}.
\newblock URL \url{https://www.aclweb.org/anthology/P17-1080}.

\bibitem[Bender et~al.(2021)Bender, Gebru, McMillan-Major, and
  Shmitchell]{bender2021dangers}
Emily~M Bender, Timnit Gebru, Angelina McMillan-Major, and Shmargaret
  Shmitchell.
\newblock On the dangers of stochastic parrots: Can language models be too big?
\newblock In \emph{Proceedings of the 2021 ACM Conference on Fairness,
  Accountability, and Transparency}, pages 610--623, 2021.

\bibitem[Bengio et~al.(2000)Bengio, Ducharme, and Vincent]{bengio2000neural}
Yoshua Bengio, R{\'e}jean Ducharme, and Pascal Vincent.
\newblock A neural probabilistic language model.
\newblock \emph{Advances in Neural Information Processing Systems}, 2000.

\bibitem[Biderman et~al.(2022)Biderman, Bicheno, and
  Gao]{biderman2022datasheet}
Stella Biderman, Kieran Bicheno, and Leo Gao.
\newblock Datasheet for the pile.
\newblock \emph{arXiv preprint arXiv:2201.07311}, 2022.

\bibitem[{BigScience Workshop}(2022)]{bloom_model}
{BigScience Workshop}.
\newblock {BLOOM} (revision 4ab0472), 2022.
\newblock URL \url{https://huggingface.co/bigscience/bloom}.

\bibitem[Birhane et~al.(2021)Birhane, Prabhu, and
  Kahembwe]{birhane2021multimodal}
Abeba Birhane, Vinay~Uday Prabhu, and Emmanuel Kahembwe.
\newblock Multimodal datasets: misogyny, pornography, and malignant
  stereotypes.
\newblock \emph{ArXiv}, abs/2110.01963, 2021.

\bibitem[Birhane et~al.(2022)Birhane, Kalluri, Card, Agnew, Dotan, and
  Bao]{birhane2022values}
Abeba Birhane, Pratyusha Kalluri, Dallas Card, William Agnew, Ravit Dotan, and
  Michelle Bao.
\newblock The values encoded in machine learning research.
\newblock In \emph{2022 ACM Conference on Fairness, Accountability, and
  Transparency}, FAccT '22, page 173–184, New York, NY, USA, 2022.
  Association for Computing Machinery.
\newblock ISBN 9781450393522.
\newblock \doi{10.1145/3531146.3533083}.
\newblock URL \url{https://doi.org/10.1145/3531146.3533083}.

\bibitem[Black et~al.()Black, Gao, Wang, Leahy, and Biderman]{black2021gpt}
Sid Black, Leo Gao, Phil Wang, Connor Leahy, and Stella Biderman.
\newblock Gpt-neo: Large scale autoregressive language modeling with
  mesh-tensorflow, march 2021.
\newblock \emph{URL https://doi. org/10.5281/zenodo}, 5297715.

\bibitem[Black et~al.(2022)Black, Biderman, Hallahan, Anthony, Gao, Golding,
  He, Leahy, McDonell, Phang, et~al.]{black2022gpt}
Sid Black, Stella Biderman, Eric Hallahan, Quentin Anthony, Leo Gao, Laurence
  Golding, Horace He, Connor Leahy, Kyle McDonell, Jason Phang, et~al.
\newblock {GPT-NeoX-20B}: An open-source autoregressive language model.
\newblock \emph{arXiv preprint arXiv:2204.06745}, 2022.

\bibitem[Blodgett et~al.(2021)Blodgett, Lopez, Olteanu, Sim, and
  Wallach]{blodgett-etal-2021-stereotyping}
Su~Lin Blodgett, Gilsinia Lopez, Alexandra Olteanu, Robert Sim, and Hanna
  Wallach.
\newblock Stereotyping {N}orwegian salmon: An inventory of pitfalls in fairness
  benchmark datasets.
\newblock In \emph{Proceedings of the 59th Annual Meeting of the Association
  for Computational Linguistics and the 11th International Joint Conference on
  Natural Language Processing (Volume 1: Long Papers)}, pages 1004--1015,
  Online, August 2021. Association for Computational Linguistics.
\newblock \doi{10.18653/v1/2021.acl-long.81}.
\newblock URL \url{https://aclanthology.org/2021.acl-long.81}.

\bibitem[Bojar et~al.(2014)Bojar, Buck, Federmann, Haddow, Koehn, Leveling,
  Monz, Pecina, Post, Saint-Amand, Soricut, Specia, and
  Tamchyna]{bojar-etal-2014-findings}
Ond{\v{r}}ej Bojar, Christian Buck, Christian Federmann, Barry Haddow, Philipp
  Koehn, Johannes Leveling, Christof Monz, Pavel Pecina, Matt Post, Herve
  Saint-Amand, Radu Soricut, Lucia Specia, and Ale{\v{s}} Tamchyna.
\newblock Findings of the 2014 workshop on statistical machine translation.
\newblock In \emph{Proceedings of the Ninth Workshop on Statistical Machine
  Translation}, pages 12--58, Baltimore, Maryland, USA, June 2014. Association
  for Computational Linguistics.
\newblock \doi{10.3115/v1/W14-3302}.
\newblock URL \url{https://aclanthology.org/W14-3302}.

\bibitem[Brennen(2018)]{brennen2018hype}
J.~Scott Brennen.
\newblock An industry-led debate: how uk media cover artificial intelligence,
  2018.

\bibitem[Brennen et~al.(2022)Brennen, Howard, and Nielsen]{brennen2022expect}
J~Scott Brennen, Philip~N Howard, and Rasmus~K Nielsen.
\newblock What to expect when you’re expecting robots: Futures, expectations,
  and pseudo-artificial general intelligence in uk news.
\newblock \emph{Journalism}, 23\penalty0 (1):\penalty0 22--38, 2022.
\newblock \doi{10.1177/1464884920947535}.
\newblock URL \url{https://doi.org/10.1177/1464884920947535}.

\bibitem[Brown et~al.(2020)Brown, Mann, Ryder, Subbiah, Kaplan, Dhariwal,
  Neelakantan, Shyam, Sastry, Askell, Agarwal, Herbert-Voss, Krueger, Henighan,
  Child, Ramesh, Ziegler, Wu, Winter, Hesse, Chen, Sigler, Litwin, Gray, Chess,
  Clark, Berner, McCandlish, Radford, Sutskever, and Amodei]{brown2020language}
Tom Brown, Benjamin Mann, Nick Ryder, Melanie Subbiah, Jared~D Kaplan, Prafulla
  Dhariwal, Arvind Neelakantan, Pranav Shyam, Girish Sastry, Amanda Askell,
  Sandhini Agarwal, Ariel Herbert-Voss, Gretchen Krueger, Tom Henighan, Rewon
  Child, Aditya Ramesh, Daniel Ziegler, Jeffrey Wu, Clemens Winter, Chris
  Hesse, Mark Chen, Eric Sigler, Mateusz Litwin, Scott Gray, Benjamin Chess,
  Jack Clark, Christopher Berner, Sam McCandlish, Alec Radford, Ilya Sutskever,
  and Dario Amodei.
\newblock Language models are few-shot learners.
\newblock \emph{Advances in Neural Information Processing Systems}, 2020.

\bibitem[Caswell et~al.(2022)Caswell, Kreutzer, Wang, Wahab, van Esch,
  Ulzii-Orshikh, Tapo, Subramani, Sokolov, Sikasote, Setyawan, Sarin, Samb,
  Sagot, Rivera, Gonzales, Papadimitriou, Osei, Suarez, Orife, Ogueji,
  Niyongabo, Nguyen, Muller, Muller, Muhammad, Muhammad, Mnyakeni, Mirzakhalov,
  Matangira, Leong, Lawson, Kudugunta, Jernite, Jenny, Firat, Dossou, Dlamini,
  de~Silva, cCabuk Balli, Biderman, Battisti, Baruwa, Bapna, Baljekar, Azime,
  Awokoya, Ataman, Ahia, Ahia, Agrawal, and Adeyemi]{caswell2022quality}
Isaac Caswell, Julia Kreutzer, Lisa Wang, Ahsan Wahab, Daan van Esch,
  Nasanbayar Ulzii-Orshikh, Allahsera~Auguste Tapo, Nishant Subramani, Artem
  Sokolov, Claytone Sikasote, Monang Setyawan, Supheakmungkol Sarin, Sokhar
  Samb, Beno{\^\i}t Sagot, Clara Rivera, Annette~Rios Gonzales, Isabel
  Papadimitriou, Salomey Osei, Pedro~Ortiz Suarez, Iroro Orife, Kelechi Ogueji,
  Rubungo~Andre Niyongabo, Toan~Q. Nguyen, Mathias Muller, Andre~Matthias
  Muller, Shamsuddeen~Hassan Muhammad, Nanda~Firdausi Muhammad, Ayanda
  Mnyakeni, Jamshidbek Mirzakhalov, Tapiwanashe Matangira, Colin Leong, Nze
  Lawson, Sneha Kudugunta, Yacine Jernite, M.~Jenny, Orhan Firat, Bonaventure
  F.~P. Dossou, Sakhile Dlamini, Nisansa de~Silva, Sakine cCabuk Balli,
  Stella~Rose Biderman, Alessia Battisti, Ahmed Baruwa, Ankur Bapna, Pallavi~N.
  Baljekar, Israel~Abebe Azime, Ayodele Awokoya, Duygu Ataman, Orevaoghene
  Ahia, Oghenefego Ahia, Sweta Agrawal, and Mofetoluwa Adeyemi.
\newblock Quality at a glance: An audit of web-crawled multilingual datasets.
\newblock \emph{Transactions of the Association for Computational Linguistics},
  10:\penalty0 50--72, 2022.

\bibitem[Chen et~al.(2021)Chen, Tworek, Jun, Yuan, Pinto, Kaplan, Edwards,
  Burda, Joseph, Brockman, et~al.]{chen2021evaluating}
Mark Chen, Jerry Tworek, Heewoo Jun, Qiming Yuan, Henrique Ponde de~Oliveira
  Pinto, Jared Kaplan, Harri Edwards, Yuri Burda, Nicholas Joseph, Greg
  Brockman, et~al.
\newblock Evaluating large language models trained on code.
\newblock \emph{arXiv preprint arXiv:2107.03374}, 2021.

\bibitem[Choshen et~al.(2022)Choshen, Hacohen, Weinshall, and
  Abend]{choshen-etal-2022-grammar}
Leshem Choshen, Guy Hacohen, Daphna Weinshall, and Omri Abend.
\newblock The grammar-learning trajectories of neural language models.
\newblock In \emph{Proceedings of the 60th Annual Meeting of the Association
  for Computational Linguistics (Volume 1: Long Papers)}, pages 8281--8297,
  Dublin, Ireland, May 2022. Association for Computational Linguistics.
\newblock \doi{10.18653/v1/2022.acl-long.568}.
\newblock URL \url{https://aclanthology.org/2022.acl-long.568}.

\bibitem[Chowdhery et~al.(2022)Chowdhery, Narang, Devlin, Bosma, Mishra,
  Roberts, Barham, Chung, Sutton, Gehrmann, Schuh, Shi, Tsvyashchenko, Maynez,
  Abhishek~Rao, Tay, Shazeer, Prabhakaran, Reif, Du, Hutchinson, Pope,
  Bradbury, Austin, Isard, Gur-Ari, Yin, Duke, Levskaya, Ghemawat, Dev,
  Michalewski, Garcia, Misra, Robinson, Fedus, Zhou, Ippolito, Luan, Lim, Zoph,
  Spiridonov, Sepassi, Dohan, Agrawal, Omernick, Dai, Pillai, Pellat,
  Lewkowycz, Moreira, Child, Polozov, Lee, Zhou, Wang, Saeta, Diaz, Firat,
  Catasta, Wei, Meier-Hellstern, Eck, Dean, Petrov, and
  Fiedel]{chowdhery2022palm}
Aakanksha Chowdhery, Sharan Narang, Jacob Devlin, Maarten Bosma, Gaurav Mishra,
  Adam Roberts, Paul Barham, Hyung~Won Chung, Charles Sutton, Sebastian
  Gehrmann, Parker Schuh, Kensen Shi, Sasha Tsvyashchenko, Joshua Maynez,
  Parker~Barnes Abhishek~Rao, Yi~Tay, Noam Shazeer, Vinodkumar Prabhakaran,
  Emily Reif, Nan Du, Ben Hutchinson, Reiner Pope, James Bradbury, Jacob
  Austin, Michael Isard, Guy Gur-Ari, Pengcheng Yin, Toju Duke, Anselm
  Levskaya, Sanjay Ghemawat, Sunipa Dev, Henryk Michalewski, Xavier Garcia,
  Vedant Misra, Kevin Robinson, Liam Fedus, Denny Zhou, Daphne Ippolito, David
  Luan, Hyeontaek Lim, Barret Zoph, Alexander Spiridonov, Ryan Sepassi, David
  Dohan, Shivani Agrawal, Mark Omernick, Andrew~M. Dai,
  Thanumalayan~Sankaranarayana Pillai, Marie Pellat, Aitor Lewkowycz, Erica
  Moreira, Rewon Child, Oleksandr Polozov, Katherine Lee, Zongwei Zhou, Xuezhi
  Wang, Brennan Saeta, Mark Diaz, Orhan Firat, Michele Catasta, Jason Wei,
  Kathy Meier-Hellstern, Douglas Eck, Jeff Dean, Slav Petrov, and Noah Fiedel.
\newblock Palm: Scaling language modeling with pathways.
\newblock \emph{arXiv preprint arXiv:2204.02311}, 2022.

\bibitem[Chung et~al.(2022)Chung, Hou, Longpre, Zoph, Tay, Fedus, Li, Wang,
  Dehghani, Brahma, et~al.]{chung2022scaling}
Hyung~Won Chung, Le~Hou, Shayne Longpre, Barret Zoph, Yi~Tay, William Fedus,
  Eric Li, Xuezhi Wang, Mostafa Dehghani, Siddhartha Brahma, et~al.
\newblock Scaling instruction-finetuned language models.
\newblock \emph{arXiv preprint arXiv:2210.11416}, 2022.

\bibitem[Collobert et~al.(2011)Collobert, Weston, Bottou, Karlen, Kavukcuoglu,
  and Kuksa]{collobert2011natural}
Ronan Collobert, Jason Weston, L{\'e}on Bottou, Michael Karlen, Koray
  Kavukcuoglu, and Pavel Kuksa.
\newblock Natural language processing (almost) from scratch.
\newblock \emph{Journal of machine learning research}, 12, 2011.

\bibitem[Conneau et~al.(2018)Conneau, Kruszewski, Lample, Barrault, and
  Baroni]{conneau-etal-2018-cram}
Alexis Conneau, German Kruszewski, Guillaume Lample, Lo{\"\i}c Barrault, and
  Marco Baroni.
\newblock What you can cram into a single {\$}{\&}!{\#}* vector: Probing
  sentence embeddings for linguistic properties.
\newblock In \emph{Proceedings of the 56th Annual Meeting of the Association
  for Computational Linguistics (Volume 1: Long Papers)}, pages 2126--2136,
  Melbourne, Australia, July 2018. Association for Computational Linguistics.
\newblock \doi{10.18653/v1/P18-1198}.
\newblock URL \url{https://aclanthology.org/P18-1198}.

\bibitem[Conneau et~al.(2020)Conneau, Khandelwal, Goyal, Chaudhary, Wenzek,
  Guzm{\'a}n, Grave, Ott, Zettlemoyer, and Stoyanov]{conneau2020unsupervised}
Alexis Conneau, Kartikay Khandelwal, Naman Goyal, Vishrav Chaudhary, Guillaume
  Wenzek, Francisco Guzm{\'a}n, Edouard Grave, Myle Ott, Luke Zettlemoyer, and
  Veselin Stoyanov.
\newblock Unsupervised cross-lingual representation learning at scale.
\newblock In \emph{Proceedings of the 58th Annual Meeting of the Association
  for Computational Linguistics}, pages 8440--8451, Online, July 2020.
  Association for Computational Linguistics.
\newblock \doi{10.18653/v1/2020.acl-main.747}.
\newblock URL \url{https://aclanthology.org/2020.acl-main.747}.

\bibitem[Contractor et~al.(2022)Contractor, McDuff, Haines, Lee, Hines, Hecht,
  Vincent, and Li]{LicensingPaper}
Danish Contractor, Daniel McDuff, Julia~Katherine Haines, Jenny Lee,
  Christopher Hines, Brent Hecht, Nicholas Vincent, and Hanlin Li.
\newblock Behavioral use licensing for responsible ai.
\newblock In \emph{2022 ACM Conference on Fairness, Accountability, and
  Transparency}, FAccT '22, page 778–788, New York, NY, USA, 2022.
  Association for Computing Machinery.
\newblock ISBN 9781450393522.
\newblock \doi{10.1145/3531146.3533143}.
\newblock URL \url{https://doi.org/10.1145/3531146.3533143}.

\bibitem[De~Toni et~al.(2022)De~Toni, Akiki, De~La~Rosa, Fourrier, Manjavacas,
  Schweter, and Van~Strien]{de-toni-etal-2022-entities}
Francesco De~Toni, Christopher Akiki, Javier De~La~Rosa, Cl{\'e}mentine
  Fourrier, Enrique Manjavacas, Stefan Schweter, and Daniel Van~Strien.
\newblock Entities, dates, and languages: Zero-shot on historical texts with
  t0.
\newblock In \emph{Proceedings of BigScience Episode {\#}5 -- Workshop on
  Challenges {\&} Perspectives in Creating Large Language Models}, pages
  75--83, virtual+Dublin, May 2022. Association for Computational Linguistics.
\newblock \doi{10.18653/v1/2022.bigscience-1.7}.
\newblock URL \url{https://aclanthology.org/2022.bigscience-1.7}.

\bibitem[Dettmers et~al.(2022)Dettmers, Lewis, Belkada, and
  Zettlemoyer]{dettmers2022llm}
Tim Dettmers, Mike Lewis, Younes Belkada, and Luke Zettlemoyer.
\newblock {LLM}.int8(): 8-bit matrix multiplication for transformers at scale.
\newblock \emph{arXiv preprint arXiv:2208.07339}, 2022.

\bibitem[Devlin et~al.(2019)Devlin, Chang, Lee, and Toutanova]{devlin2019bert}
Jacob Devlin, Ming-Wei Chang, Kenton Lee, and Kristina Toutanova.
\newblock {BERT}: Pre-training of deep bidirectional transformers for language
  understanding.
\newblock In \emph{Conference of the North American Chapter of the Association
  for Computational Linguistics}, 2019.

\bibitem[Dodge et~al.(2021)Dodge, Sap, Marasovi{\'c}, Agnew, Ilharco,
  Groeneveld, Mitchell, and Gardner]{dodge2021documenting}
Jesse Dodge, Maarten Sap, Ana Marasovi{\'c}, William Agnew, Gabriel Ilharco,
  Dirk Groeneveld, Margaret Mitchell, and Matt Gardner.
\newblock Documenting large webtext corpora: A case study on the colossal clean
  crawled corpus.
\newblock In \emph{Conference on Empirical Methods in Natural Language
  Processing}, 2021.

\bibitem[Ettinger et~al.(2016)Ettinger, Elgohary, and
  Resnik]{ettinger-etal-2016-probing}
Allyson Ettinger, Ahmed Elgohary, and Philip Resnik.
\newblock Probing for semantic evidence of composition by means of simple
  classification tasks.
\newblock In \emph{Proceedings of the 1st Workshop on Evaluating Vector-Space
  Representations for {NLP}}, pages 134--139, Berlin, Germany, August 2016.
  Association for Computational Linguistics.
\newblock \doi{10.18653/v1/W16-2524}.
\newblock URL \url{https://www.aclweb.org/anthology/W16-2524}.

\bibitem[Fan et~al.(2021)Fan, Bhosale, Schwenk, Ma, El-Kishky, Goyal, Baines,
  Celebi, Wenzek, Chaudhary, Goyal, Birch, Liptchinsky, Edunov, Auli, and
  Joulin]{fan-etal-2021-beyond}
Angela Fan, Shruti Bhosale, Holger Schwenk, Zhiyi Ma, Ahmed El-Kishky,
  Siddharth Goyal, Mandeep Baines, Onur Celebi, Guillaume Wenzek, Vishrav
  Chaudhary, Naman Goyal, Tom Birch, Vitaliy Liptchinsky, Sergey Edunov,
  Michael Auli, and Armand Joulin.
\newblock Beyond {English-Centric} multilingual machine translation.
\newblock \emph{Journal of Machine Learning Research}, 22\penalty0
  (107):\penalty0 1--48, 2021.
\newblock URL \url{http://jmlr.org/papers/v22/20-1307.html}.

\bibitem[Fedus et~al.(2022)Fedus, Zoph, and Shazeer]{fedus2022switch}
William Fedus, Barret Zoph, and Noam Shazeer.
\newblock Switch transformers: Scaling to trillion parameter models with simple
  and efficient sparsity.
\newblock \emph{Journal of Machine Learning Research}, 23\penalty0
  (120):\penalty0 1--39, 2022.

\bibitem[FitzGerald et~al.(2022)FitzGerald, Hench, Peris, Mackie, Rottmann,
  Sanchez, Nash, Urbach, Kakarala, Singh, Ranganath, Crist, Britan, Leeuwis,
  Tur, and Natarajan]{fitzgerald2022massive}
Jack FitzGerald, Christopher Hench, Charith Peris, Scott Mackie, Kay Rottmann,
  Ana Sanchez, Aaron Nash, Liam Urbach, Vishesh Kakarala, Richa Singh, Swetha
  Ranganath, Laurie Crist, Misha Britan, Wouter Leeuwis, Gokhan Tur, and Prem
  Natarajan.
\newblock Massive: A 1m-example multilingual natural language understanding
  dataset with 51 typologically-diverse languages, 2022.
\newblock URL \url{https://arxiv.org/abs/2204.08582}.

\bibitem[Fried et~al.(2022)Fried, Aghajanyan, Lin, Wang, Wallace, Shi, Zhong,
  Yih, Zettlemoyer, and Lewis]{fried2022incoder}
Daniel Fried, Armen Aghajanyan, Jessy Lin, Sida Wang, Eric Wallace, Freda Shi,
  Ruiqi Zhong, Wen-tau Yih, Luke Zettlemoyer, and Mike Lewis.
\newblock Incoder: A generative model for code infilling and synthesis.
\newblock \emph{arXiv preprint arXiv:2204.05999}, 2022.

\bibitem[Fries et~al.(2022{\natexlab{a}})Fries, Seelam, Altay, Weber, Kang,
  Datta, Su, Garda, Wang, Ott, Samwald, and Kusa]{fries2022dataset}
Jason~Alan Fries, Natasha Seelam, Gabriel Altay, Leon Weber, Myungsun Kang,
  Debajyoti Datta, Ruisi Su, Samuele Garda, Bo~Wang, Simon Ott, Matthias
  Samwald, and Wojciech Kusa.
\newblock Dataset debt in biomedical language modeling.
\newblock In \emph{Challenges {\&} Perspectives in Creating Large Language
  Models}, 2022{\natexlab{a}}.
\newblock URL \url{https://openreview.net/forum?id=HRfzInfr8Z9}.

\bibitem[Fries et~al.(2022{\natexlab{b}})Fries, Weber, Seelam, Altay, Datta,
  Garda, Kang, Su, Kusa, Cahyawijaya, Barth, Ott, Samwald, Bach, Biderman,
  S{\"a}nger, Wang, Callahan, Peri{\~n}{\'a}n, Gigant, Haller, Chim, Posada,
  Giorgi, Sivaraman, P{\`a}mies, Nezhurina, Martin, Cullan, Freidank, Dahlberg,
  Mishra, Bose, Broad, Labrak, Deshmukh, Kiblawi, Singh, Vu, Neeraj, Golde, del
  Moral, and Beilharz]{bigbio}
Jason~Alan Fries, Leon Weber, Natasha Seelam, Gabriel Altay, Debajyoti Datta,
  Samuele Garda, Myungsun Kang, Ruisi Su, Wojciech Kusa, Samuel Cahyawijaya,
  Fabio Barth, Simon Ott, Matthias Samwald, Stephen Bach, Stella Biderman,
  Mario S{\"a}nger, Bo~Wang, Alison Callahan, Daniel~Le{\'o}n Peri{\~n}{\'a}n,
  Th{\'e}o Gigant, Patrick Haller, Jenny Chim, Jose~David Posada, John~Michael
  Giorgi, Karthik~Rangasai Sivaraman, Marc P{\`a}mies, Marianna Nezhurina,
  Robert Martin, Michael Cullan, Moritz Freidank, Nathan Dahlberg, Shubhanshu
  Mishra, Shamik Bose, Nicholas~Michio Broad, Yanis Labrak, Shlok~S Deshmukh,
  Sid Kiblawi, Ayush Singh, Minh~Chien Vu, Trishala Neeraj, Jonas Golde,
  Albert~Villanova del Moral, and Benjamin Beilharz.
\newblock {BigBio}: A framework for data-centric biomedical natural language
  processing.
\newblock In \emph{Thirty-sixth Conference on Neural Information Processing
  Systems Datasets and Benchmarks Track}, 2022{\natexlab{b}}.
\newblock URL \url{https://openreview.net/forum?id=8lQDn9zTQlW}.

\bibitem[Fu et~al.(2023)Fu, Dao, Saab, Thomas, Rudra, and Re]{fu2023hungry}
Daniel~Y Fu, Tri Dao, Khaled~Kamal Saab, Armin~W Thomas, Atri Rudra, and
  Christopher Re.
\newblock Hungry hungry hippos: Towards language modeling with state space
  models.
\newblock In \emph{The Eleventh International Conference on Learning
  Representations}, 2023.
\newblock URL \url{https://openreview.net/forum?id=COZDy0WYGg}.

\bibitem[Gage(1994)]{gagebpe}
Philip Gage.
\newblock A new algorithm for data compression.
\newblock \emph{C Users J.}, 12\penalty0 (2):\penalty0 23–38, feb 1994.
\newblock ISSN 0898-9788.

\bibitem[Gao et~al.(2020)Gao, Biderman, Black, Golding, Hoppe, Foster, Phang,
  He, Thite, Nabeshima, Presser, and Leahy]{gao2020pile}
Leo Gao, Stella Biderman, Sid Black, Laurence Golding, Travis Hoppe, Charles
  Foster, Jason Phang, Horace He, Anish Thite, Noa Nabeshima, Shawn Presser,
  and Connor Leahy.
\newblock The pile: An 800gb dataset of diverse text for language modeling.
\newblock \emph{arXiv preprint arXiv:2101.00027}, 2020.

\bibitem[Gao et~al.(2021)Gao, Tow, Biderman, Black, DiPofi, Foster, Golding,
  Hsu, McDonell, Muennighoff, Phang, Reynolds, Tang, Thite, Wang, Wang, and
  Zou]{eval-harness}
Leo Gao, Jonathan Tow, Stella Biderman, Sid Black, Anthony DiPofi, Charles
  Foster, Laurence Golding, Jeffrey Hsu, Kyle McDonell, Niklas Muennighoff,
  Jason Phang, Laria Reynolds, Eric Tang, Anish Thite, Ben Wang, Kevin Wang,
  and Andy Zou.
\newblock A framework for few-shot language model evaluation, September 2021.
\newblock URL \url{https://doi.org/10.5281/zenodo.5371628}.

\bibitem[Gehrmann et~al.(2022{\natexlab{a}})Gehrmann, Bhattacharjee,
  Mahendiran, Wang, Papangelis, Madaan, McMillan-Major, Shvets, Upadhyay, Yao,
  Wilie, Bhagavatula, You, Thomson, Garbacea, Wang, Deutsch, Xiong, Jin,
  Gkatzia, Radev, Clark, Durmus, Ladhak, Ginter, Winata, Strobelt, Hayashi,
  Novikova, Kanerva, Chim, Zhou, Clive, Maynez, Sedoc, Juraska, Dhole, Chandu,
  Perez-Beltrachini, Ribeiro, Tunstall, Zhang, Pushkarna, Creutz, White, Kale,
  Eddine, Daheim, Subramani, Dusek, Liang, Ammanamanchi, Zhu, Puduppully, Kriz,
  Shahriyar, Cardenas, Mahamood, Osei, Cahyawijaya, Štajner, Montella,
  {Shailza}, Jolly, Mille, Hasan, Shen, Adewumi, Raunak, Raheja, Nikolaev,
  Tsai, Jernite, Xu, Sang, Liu, and Hou]{gemv2}
Sebastian Gehrmann, Abhik Bhattacharjee, Abinaya Mahendiran, Alex Wang,
  Alexandros Papangelis, Aman Madaan, Angelina McMillan-Major, Anna Shvets,
  Ashish Upadhyay, Bingsheng Yao, Bryan Wilie, Chandra Bhagavatula, Chaobin
  You, Craig Thomson, Cristina Garbacea, Dakuo Wang, Daniel Deutsch, Deyi
  Xiong, Di~Jin, Dimitra Gkatzia, Dragomir Radev, Elizabeth Clark, Esin Durmus,
  Faisal Ladhak, Filip Ginter, Genta~Indra Winata, Hendrik Strobelt, Hiroaki
  Hayashi, Jekaterina Novikova, Jenna Kanerva, Jenny Chim, Jiawei Zhou, Jordan
  Clive, Joshua Maynez, João Sedoc, Juraj Juraska, Kaustubh Dhole,
  Khyathi~Raghavi Chandu, Laura Perez-Beltrachini, Leonardo F.~R. Ribeiro,
  Lewis Tunstall, Li~Zhang, Mahima Pushkarna, Mathias Creutz, Michael White,
  Mihir~Sanjay Kale, Moussa~Kamal Eddine, Nico Daheim, Nishant Subramani,
  Ondrej Dusek, Paul~Pu Liang, Pawan~Sasanka Ammanamanchi, Qi~Zhu, Ratish
  Puduppully, Reno Kriz, Rifat Shahriyar, Ronald Cardenas, Saad Mahamood,
  Salomey Osei, Samuel Cahyawijaya, Sanja Štajner, Sebastien Montella,
  {Shailza}, Shailza Jolly, Simon Mille, Tahmid Hasan, Tianhao Shen, Tosin
  Adewumi, Vikas Raunak, Vipul Raheja, Vitaly Nikolaev, Vivian Tsai, Yacine
  Jernite, Ying Xu, Yisi Sang, Yixin Liu, and Yufang Hou.
\newblock Gemv2: Multilingual nlg benchmarking in a single line of code,
  2022{\natexlab{a}}.
\newblock URL \url{https://arxiv.org/abs/2206.11249}.

\bibitem[Gehrmann et~al.(2022{\natexlab{b}})Gehrmann, Clark, and
  Sellam]{gehrmann_cracked}
Sebastian Gehrmann, Elizabeth Clark, and Thibault Sellam.
\newblock Repairing the cracked foundation: A survey of obstacles in evaluation
  practices for generated text, 2022{\natexlab{b}}.
\newblock URL \url{https://arxiv.org/abs/2202.06935}.

\bibitem[Goodman(2001)]{goodman2001bit}
Joshua~T. Goodman.
\newblock A bit of progress in language modeling.
\newblock \emph{Computer Speech \& Language}, 15\penalty0 (4), 2001.

\bibitem[Goyal et~al.(2022)Goyal, Gao, Chaudhary, Chen, Wenzek, Ju, Krishnan,
  Ranzato, Guzm{\'a}n, and Fan]{goyal-etal-2022-flores}
Naman Goyal, Cynthia Gao, Vishrav Chaudhary, Peng-Jen Chen, Guillaume Wenzek,
  Da~Ju, Sanjana Krishnan, Marc{'}Aurelio Ranzato, Francisco Guzm{\'a}n, and
  Angela Fan.
\newblock The {F}lores-101 evaluation benchmark for low-resource and
  multilingual machine translation.
\newblock \emph{Transactions of the Association for Computational Linguistics},
  10:\penalty0 522--538, 2022.
\newblock \doi{10.1162/tacl_a_00474}.
\newblock URL \url{https://aclanthology.org/2022.tacl-1.30}.

\bibitem[Graves(2013)]{graves2013generating}
Alex Graves.
\newblock Generating sequences with recurrent neural networks.
\newblock \emph{arXiv preprint arXiv:1308.0850}, 2013.

\bibitem[Gu et~al.(2020)Gu, Dao, Ermon, Rudra, and R{\'e}]{gu2020hippo}
Albert Gu, Tri Dao, Stefano Ermon, Atri Rudra, and Christopher R{\'e}.
\newblock Hippo: Recurrent memory with optimal polynomial projections.
\newblock \emph{Advances in Neural Information Processing Systems},
  33:\penalty0 1474--1487, 2020.

\bibitem[Gu et~al.(2021)Gu, Goel, and Re]{gu2021efficiently}
Albert Gu, Karan Goel, and Christopher Re.
\newblock Efficiently modeling long sequences with structured state spaces.
\newblock In \emph{International Conference on Learning Representations}, 2021.

\bibitem[Hestness et~al.(2017)Hestness, Narang, Ardalani, Diamos, Jun,
  Kianinejad, Patwary, Ali, Yang, and Zhou]{hestness2017deep}
Joel Hestness, Sharan Narang, Newsha Ardalani, Gregory Diamos, Heewoo Jun,
  Hassan Kianinejad, Md~Patwary, Mostofa Ali, Yang Yang, and Yanqi Zhou.
\newblock Deep learning scaling is predictable, empirically.
\newblock \emph{arXiv preprint arXiv:1712.00409}, 2017.

\bibitem[Hewitt and Liang(2019)]{hewitt-liang-2019-designing}
John Hewitt and Percy Liang.
\newblock Designing and interpreting probes with control tasks.
\newblock In \emph{Proceedings of the 2019 Conference on Empirical Methods in
  Natural Language Processing and the 9th International Joint Conference on
  Natural Language Processing (EMNLP-IJCNLP)}, pages 2733--2743, Hong Kong,
  China, November 2019. Association for Computational Linguistics.
\newblock \doi{10.18653/v1/D19-1275}.
\newblock URL \url{https://aclanthology.org/D19-1275}.

\bibitem[Hoffmann et~al.(2022)Hoffmann, Borgeaud, Mensch, Buchatskaya, Cai,
  Rutherford, de~Las~Casas, Hendricks, Welbl, Clark, Hennigan, Noland,
  Millican, van~den Driessche, Damoc, Guy, Osindero, Simonyan, Elsen, Rae,
  Vinyals, and Sifre]{hoffmann2022training}
Jordan Hoffmann, Sebastian Borgeaud, Arthur Mensch, Elena Buchatskaya, Trevor
  Cai, Eliza Rutherford, Diego de~Las~Casas, Lisa~Anne Hendricks, Johannes
  Welbl, Aidan Clark, Tom Hennigan, Eric Noland, Katie Millican, George van~den
  Driessche, Bogdan Damoc, Aurelia Guy, Simon Osindero, Karen Simonyan, Erich
  Elsen, Jack~W. Rae, Oriol Vinyals, and Laurent Sifre.
\newblock Training compute-optimal large language models.
\newblock \emph{arXiv preprint arXiv:2203.15556}, 2022.

\bibitem[Howard and Ruder(2018)]{howard2018universal}
Jeremy Howard and Sebastian Ruder.
\newblock Universal language model fine-tuning for text classification.
\newblock In \emph{Annual Meeting of the Association for Computational
  Linguistics}, 2018.

\bibitem[Hupkes et~al.(2018)Hupkes, Veldhoen, and
  Zuidema]{hupkes2018visualisation}
Dieuwke Hupkes, Sara Veldhoen, and Willem Zuidema.
\newblock Visualisation and 'diagnostic classifiers' reveal how recurrent and
  recursive neural networks process hierarchical structure.
\newblock \emph{Journal of Artificial Intelligence Research}, 61:\penalty0
  907--926, 2018.

\bibitem[Jernite et~al.(2022)Jernite, Nguyen, Biderman, Rogers, Masoud,
  Danchev, Tan, Luccioni, Subramani, Johnson, Dupont, Dodge, Lo, Talat, Radev,
  Gokaslan, Nikpoor, Henderson, Bommasani, and Mitchell]{jernite2022data}
Yacine Jernite, Huu Nguyen, Stella Biderman, Anna Rogers, Maraim Masoud,
  Valentin Danchev, Samson Tan, Alexandra~Sasha Luccioni, Nishant Subramani,
  Isaac Johnson, Gerard Dupont, Jesse Dodge, Kyle Lo, Zeerak Talat, Dragomir
  Radev, Aaron Gokaslan, Somaieh Nikpoor, Peter Henderson, Rishi Bommasani, and
  Margaret Mitchell.
\newblock Data governance in the age of large-scale data-driven language
  technology.
\newblock In \emph{2022 ACM Conference on Fairness, Accountability, and
  Transparency}, FAccT '22, page 2206–2222, New York, NY, USA, 2022.
  Association for Computing Machinery.
\newblock ISBN 9781450393522.
\newblock \doi{10.1145/3531146.3534637}.
\newblock URL \url{https://doi.org/10.1145/3531146.3534637}.

\bibitem[Johnson et~al.(2022)Johnson, Pistilli, Men'edez-Gonz'alez, Duran,
  Panai, Kalpokienė, and Bertulfo]{johnson2022ghost}
Rebecca~Lynn Johnson, Giada Pistilli, Natalia Men'edez-Gonz'alez, Leslye
  Denisse~Dias Duran, Enrico Panai, Julija Kalpokienė, and Donald~Jay
  Bertulfo.
\newblock The ghost in the machine has an american accent: value conflict in
  gpt-3.
\newblock \emph{ArXiv}, abs/2203.07785, 2022.

\bibitem[Kalamkar et~al.(2019)Kalamkar, Mudigere, Mellempudi, Das, Banerjee,
  Avancha, Vooturi, Jammalamadaka, Huang, Yuen, Yang, Park, Heinecke,
  Georganas, Srinivasan, Kundu, Smelyanskiy, Kaul, and
  Dubey]{kalamkar2019study}
Dhiraj Kalamkar, Dheevatsa Mudigere, Naveen Mellempudi, Dipankar Das, Kunal
  Banerjee, Sasikanth Avancha, Dharma~Teja Vooturi, Nataraj Jammalamadaka,
  Jianyu Huang, Hector Yuen, Jiyan Yang, Jongsoo Park, Alexander Heinecke,
  Evangelos Georganas, Sudarshan Srinivasan, Abhisek Kundu, Misha Smelyanskiy,
  Bharat Kaul, and Pradeep Dubey.
\newblock A study of bfloat16 for deep learning training, 2019.

\bibitem[Kaplan et~al.(2020)Kaplan, McCandlish, Henighan, Brown, Chess, Child,
  Gray, Radford, Wu, and Amodei]{kaplan2020scaling}
Jared Kaplan, Sam McCandlish, Tom Henighan, Tom~B Brown, Benjamin Chess, Rewon
  Child, Scott Gray, Alec Radford, Jeffrey Wu, and Dario Amodei.
\newblock Scaling laws for neural language models.
\newblock \emph{arXiv preprint arXiv:2001.08361}, 2020.

\bibitem[Kim et~al.(2021)Kim, Kim, Lee, Lee, Kwak, Dong~Hyeon, Park, Kim, Kim,
  Seo, Lee, Jeong, Lee, Kim, Ko, Kim, Park, Kim, Kang, Ryu, Yoo, Chang, Suh,
  In, Park, Kim, Kim, Jeong, Yeo, Ham, Park, Lee, Kang, Kang, Ha, Park, and
  Sung]{kim2021changes}
Boseop Kim, HyoungSeok Kim, Sang-Woo Lee, Gichang Lee, Donghyun Kwak, Jeon
  Dong~Hyeon, Sunghyun Park, Sungju Kim, Seonhoon Kim, Dongpil Seo, Heungsub
  Lee, Minyoung Jeong, Sungjae Lee, Minsub Kim, Suk~Hyun Ko, Seokhun Kim,
  Taeyong Park, Jinuk Kim, Soyoung Kang, Na-Hyeon Ryu, Kang~Min Yoo, Minsuk
  Chang, Soobin Suh, Sookyo In, Jinseong Park, Kyungduk Kim, Hiun Kim, Jisu
  Jeong, Yong~Goo Yeo, Donghoon Ham, Dongju Park, Min~Young Lee, Jaewook Kang,
  Inho Kang, Jung-Woo Ha, Woomyoung Park, and Nako Sung.
\newblock What changes can large-scale language models bring? intensive study
  on {HyperCLOVA}: Billions-scale korean generative pretrained transformers.
\newblock In \emph{Conference on Empirical Methods in Natural Language
  Processing}, 2021.

\bibitem[Kl{\"o}pffer(1997)]{klopffer1997life}
Walter Kl{\"o}pffer.
\newblock Life cycle assessment.
\newblock \emph{Environmental Science and Pollution Research}, 4\penalty0
  (4):\penalty0 223--228, 1997.

\bibitem[Kudo and Richardson(2018)]{kudo-richardson-2018-sentencepiece}
Taku Kudo and John Richardson.
\newblock {S}entence{P}iece: A simple and language independent subword
  tokenizer and detokenizer for neural text processing.
\newblock In \emph{Proceedings of the 2018 Conference on Empirical Methods in
  Natural Language Processing: System Demonstrations}, pages 66--71, Brussels,
  Belgium, November 2018. Association for Computational Linguistics.
\newblock \doi{10.18653/v1/D18-2012}.
\newblock URL \url{https://aclanthology.org/D18-2012}.

\bibitem[Kunchukuttan et~al.(2020)Kunchukuttan, Kakwani, Golla, GokulN.,
  Bhattacharyya, Khapra, and Kumar]{kunchukuttan2020indic}
Anoop Kunchukuttan, Divyanshu Kakwani, Satish Golla, C.~GokulN., Avik
  Bhattacharyya, Mitesh~M. Khapra, and Pratyush Kumar.
\newblock Ai4bharat-indicnlp corpus: Monolingual corpora and word embeddings
  for indic languages.
\newblock \emph{ArXiv}, abs/2005.00085, 2020.

\bibitem[Lacoste et~al.(2019)Lacoste, Luccioni, Schmidt, and
  Dandres]{lacoste2019quantifying}
Alexandre Lacoste, Alexandra Luccioni, Victor Schmidt, and Thomas Dandres.
\newblock Quantifying the carbon emissions of machine learning.
\newblock \emph{arXiv preprint arXiv:1910.09700}, 2019.

\bibitem[Ladhak et~al.(2020)Ladhak, Durmus, Cardie, and
  McKeown]{ladhak-etal-2020-wikilingua}
Faisal Ladhak, Esin Durmus, Claire Cardie, and Kathleen McKeown.
\newblock {W}iki{L}ingua: A new benchmark dataset for cross-lingual abstractive
  summarization.
\newblock In \emph{Findings of the Association for Computational Linguistics:
  EMNLP 2020}, pages 4034--4048, Online, November 2020. Association for
  Computational Linguistics.
\newblock \doi{10.18653/v1/2020.findings-emnlp.360}.
\newblock URL \url{https://aclanthology.org/2020.findings-emnlp.360}.

\bibitem[Lauren{\c{c}}on et~al.(2022)Lauren{\c{c}}on, Saulnier, Wang, Akiki,
  del Moral, Scao, Werra, Mou, Ponferrada, Nguyen, Frohberg, {\v{S}}a{\v{s}}ko,
  Lhoest, McMillan-Major, Dupont, Biderman, Rogers, allal, Toni, Pistilli,
  Nguyen, Nikpoor, Masoud, Colombo, de~la Rosa, Villegas, Thrush, Longpre,
  Nagel, Weber, Mu{\~n}oz, Zhu, Strien, Alyafeai, Almubarak, Chien,
  Gonzalez-Dios, Soroa, Lo, Dey, Suarez, Gokaslan, Bose, Adelani, Phan, Tran,
  Yu, Pai, Chim, Lepercq, Ilic, Mitchell, Luccioni, and
  Jernite]{laurencon2022bigscience}
Hugo Lauren{\c{c}}on, Lucile Saulnier, Thomas Wang, Christopher Akiki,
  Albert~Villanova del Moral, Teven~Le Scao, Leandro~Von Werra, Chenghao Mou,
  Eduardo~Gonz{\'a}lez Ponferrada, Huu Nguyen, J{\"o}rg Frohberg, Mario
  {\v{S}}a{\v{s}}ko, Quentin Lhoest, Angelina McMillan-Major, G{\'e}rard
  Dupont, Stella Biderman, Anna Rogers, Loubna~Ben allal, Francesco~De Toni,
  Giada Pistilli, Olivier Nguyen, Somaieh Nikpoor, Maraim Masoud, Pierre
  Colombo, Javier de~la Rosa, Paulo Villegas, Tristan Thrush, Shayne Longpre,
  Sebastian Nagel, Leon Weber, Manuel~Romero Mu{\~n}oz, Jian Zhu, Daniel~Van
  Strien, Zaid Alyafeai, Khalid Almubarak, Vu~Minh Chien, Itziar Gonzalez-Dios,
  Aitor Soroa, Kyle Lo, Manan Dey, Pedro~Ortiz Suarez, Aaron Gokaslan, Shamik
  Bose, David~Ifeoluwa Adelani, Long Phan, Hieu Tran, Ian Yu, Suhas Pai, Jenny
  Chim, Violette Lepercq, Suzana Ilic, Margaret Mitchell, Sasha Luccioni, and
  Yacine Jernite.
\newblock The {BigScience} {ROOTS} corpus: A 1.6{TB} composite multilingual
  dataset.
\newblock In \emph{Thirty-sixth Conference on Neural Information Processing
  Systems Datasets and Benchmarks Track}, 2022.
\newblock URL \url{https://openreview.net/forum?id=UoEw6KigkUn}.

\bibitem[{Le Scao} et~al.(2022){Le Scao}, Wang, Hesslow, Saulnier, Bekman,
  Bari, Biderman, Elsahar, Muennighoff, Phang, Press, Raffel, Sanh, Shen,
  Sutawika, Tae, Yong, Launay, and Beltagy]{scao2022what}
Teven {Le Scao}, Thomas Wang, Daniel Hesslow, Lucile Saulnier, Stas Bekman,
  M~Saiful Bari, Stella Biderman, Hady Elsahar, Niklas Muennighoff, Jason
  Phang, Ofir Press, Colin Raffel, Victor Sanh, Sheng Shen, Lintang Sutawika,
  Jaesung Tae, Zheng~Xin Yong, Julien Launay, and Iz~Beltagy.
\newblock What language model to train if you have one million {GPU} hours?
\newblock In \emph{Challenges {\&} Perspectives in Creating Large Language
  Models}, 2022.
\newblock URL \url{https://openreview.net/forum?id=rI7BL3fHIZq}.

\bibitem[Lewis et~al.(2020)Lewis, Liu, Goyal, Ghazvininejad, Mohamed, Levy,
  Stoyanov, and Zettlemoyer]{lewis2020bart}
Mike Lewis, Yinhan Liu, Naman Goyal, Marjan Ghazvininejad, Abdelrahman Mohamed,
  Omer Levy, Veselin Stoyanov, and Luke Zettlemoyer.
\newblock {BART}: Denoising sequence-to-sequence pre-training for natural
  language generation, translation, and comprehension.
\newblock In \emph{Annual Meeting of the Association for Computational
  Linguistics}, 2020.

\bibitem[Lhoest et~al.(2021)Lhoest, Villanova~del Moral, Jernite, Thakur, von
  Platen, Patil, Chaumond, Drame, Plu, Tunstall, Davison, {\v{S}}a{\v{s}}ko,
  Chhablani, Malik, Brandeis, Le~Scao, Sanh, Xu, Patry, McMillan-Major, Schmid,
  Gugger, Delangue, Matussi{\`e}re, Debut, Bekman, Cistac, Goehringer, Mustar,
  Lagunas, Rush, and Wolf]{lhoest-etal-2021-datasets}
Quentin Lhoest, Albert Villanova~del Moral, Yacine Jernite, Abhishek Thakur,
  Patrick von Platen, Suraj Patil, Julien Chaumond, Mariama Drame, Julien Plu,
  Lewis Tunstall, Joe Davison, Mario {\v{S}}a{\v{s}}ko, Gunjan Chhablani,
  Bhavitvya Malik, Simon Brandeis, Teven Le~Scao, Victor Sanh, Canwen Xu,
  Nicolas Patry, Angelina McMillan-Major, Philipp Schmid, Sylvain Gugger,
  Cl{\'e}ment Delangue, Th{\'e}o Matussi{\`e}re, Lysandre Debut, Stas Bekman,
  Pierric Cistac, Thibault Goehringer, Victor Mustar, Fran{\c{c}}ois Lagunas,
  Alexander Rush, and Thomas Wolf.
\newblock Datasets: A community library for natural language processing.
\newblock In \emph{Proceedings of the 2021 Conference on Empirical Methods in
  Natural Language Processing: System Demonstrations}, pages 175--184, Online
  and Punta Cana, Dominican Republic, November 2021. Association for
  Computational Linguistics.
\newblock \doi{10.18653/v1/2021.emnlp-demo.21}.
\newblock URL \url{https://aclanthology.org/2021.emnlp-demo.21}.

\bibitem[Li et~al.(2022)Li, Choi, Chung, Kushman, Schrittwieser, Leblond,
  Eccles, Keeling, Gimeno, Lago, Hubert, Choy, de~Masson~d'Autume, Babuschkin,
  Chen, Huang, Welbl, Gowal, Cherepanov, Molloy, Mankowitz, Robson, Kohli,
  de~Freitas, Kavukcuoglu, and Vinyals]{alphacode}
Yujia Li, David~H. Choi, Junyoung Chung, Nate Kushman, Julian Schrittwieser,
  R{\'{e}}mi Leblond, Tom Eccles, James Keeling, Felix Gimeno, Agustin~Dal
  Lago, Thomas Hubert, Peter Choy, Cyprien de~Masson~d'Autume, Igor Babuschkin,
  Xinyun Chen, Po{-}Sen Huang, Johannes Welbl, Sven Gowal, Alexey Cherepanov,
  James Molloy, Daniel~J. Mankowitz, Esme~Sutherland Robson, Pushmeet Kohli,
  Nando de~Freitas, Koray Kavukcuoglu, and Oriol Vinyals.
\newblock Competition-level code generation with {AlphaCode}.
\newblock \emph{CoRR}, abs/2203.07814, 2022.
\newblock \doi{10.48550/arXiv.2203.07814}.
\newblock URL \url{https://doi.org/10.48550/arXiv.2203.07814}.

\bibitem[Liang et~al.(2022)Liang, Bommasani, Lee, Tsipras, Soylu, Yasunaga,
  Zhang, Narayanan, Wu, Kumar, Newman, Yuan, Yan, Zhang, Cosgrove, Manning,
  Ré, Acosta-Navas, Hudson, Zelikman, Durmus, Ladhak, Rong, Ren, Yao, Wang,
  Santhanam, Orr, Zheng, Yuksekgonul, Suzgun, Kim, Guha, Chatterji, Khattab,
  Henderson, Huang, Chi, Xie, Santurkar, Ganguli, Hashimoto, Icard, Zhang,
  Chaudhary, Wang, Li, Mai, Zhang, and Koreeda]{HELM}
Percy Liang, Rishi Bommasani, Tony Lee, Dimitris Tsipras, Dilara Soylu,
  Michihiro Yasunaga, Yian Zhang, Deepak Narayanan, Yuhuai Wu, Ananya Kumar,
  Benjamin Newman, Binhang Yuan, Bobby Yan, Ce~Zhang, Christian Cosgrove,
  Christopher~D. Manning, Christopher Ré, Diana Acosta-Navas, Drew~A. Hudson,
  Eric Zelikman, Esin Durmus, Faisal Ladhak, Frieda Rong, Hongyu Ren, Huaxiu
  Yao, Jue Wang, Keshav Santhanam, Laurel Orr, Lucia Zheng, Mert Yuksekgonul,
  Mirac Suzgun, Nathan Kim, Neel Guha, Niladri Chatterji, Omar Khattab, Peter
  Henderson, Qian Huang, Ryan Chi, Sang~Michael Xie, Shibani Santurkar, Surya
  Ganguli, Tatsunori Hashimoto, Thomas Icard, Tianyi Zhang, Vishrav Chaudhary,
  William Wang, Xuechen Li, Yifan Mai, Yuhui Zhang, and Yuta Koreeda.
\newblock Holistic evaluation of language models, 2022.
\newblock URL \url{https://arxiv.org/abs/2211.09110}.

\bibitem[Lin(2004)]{lin-2004-rouge}
Chin-Yew Lin.
\newblock {ROUGE}: A package for automatic evaluation of summaries.
\newblock In \emph{Text Summarization Branches Out}, pages 74--81, Barcelona,
  Spain, July 2004. Association for Computational Linguistics.
\newblock URL \url{https://aclanthology.org/W04-1013}.

\bibitem[Lin et~al.(2021)Lin, Mihaylov, Artetxe, Wang, Chen, Simig, Ott, Goyal,
  Bhosale, Du, Pasunuru, Shleifer, Koura, Chaudhary, O'Horo, Wang, Zettlemoyer,
  Kozareva, Diab, Stoyanov, and Li]{lin2021xglm}
Xi~Victoria Lin, Todor Mihaylov, Mikel Artetxe, Tianlu Wang, Shuohui Chen,
  Daniel Simig, Myle Ott, Naman Goyal, Shruti Bhosale, Jingfei Du, Ramakanth
  Pasunuru, Sam Shleifer, Punit~Singh Koura, Vishrav Chaudhary, Brian O'Horo,
  Jeff Wang, Luke Zettlemoyer, Zornitsa Kozareva, Mona Diab, Veselin Stoyanov,
  and Xian Li.
\newblock Few-shot learning with multilingual language models, 2021.
\newblock URL \url{https://arxiv.org/abs/2112.10668}.

\bibitem[Liu et~al.(2019)Liu, Ott, Goyal, Du, Joshi, Chen, Levy, Lewis,
  Zettlemoyer, and Stoyanov]{liu2019roberta}
Yinhan Liu, Myle Ott, Naman Goyal, Jingfei Du, Mandar Joshi, Danqi Chen, Omer
  Levy, Mike Lewis, Luke Zettlemoyer, and Veselin Stoyanov.
\newblock {RoBERTa}: A robustly optimized {BERT} pretraining approach.
\newblock \emph{arXiv preprint arXiv:1907.11692}, 2019.

\bibitem[Lo et~al.(2020)Lo, Wang, Neumann, Kinney, and Weld]{lo2020s2orc}
Kyle Lo, Lucy~Lu Wang, Mark Neumann, Rodney~Michael Kinney, and Daniel~S. Weld.
\newblock {S2ORC}: The semantic scholar open research corpus.
\newblock In \emph{ACL}, 2020.

\bibitem[Loshchilov and Hutter(2016)]{loschchilovcosinedecay}
Ilya Loshchilov and Frank Hutter.
\newblock {SGDR:} stochastic gradient descent with restarts.
\newblock \emph{CoRR}, abs/1608.03983, 2016.
\newblock URL \url{http://arxiv.org/abs/1608.03983}.

\bibitem[Luccioni et~al.(2022)Luccioni, Viguier, and
  Ligozat]{luccioni2022estimating}
Alexandra~Sasha Luccioni, Sylvain Viguier, and Anne-Laure Ligozat.
\newblock {Estimating the Carbon Footprint of {BLOOM}, a 176B Parameter
  Language Model}.
\newblock \emph{arXiv preprint arXiv:2211.02001}, 2022.

\bibitem[Madabushi et~al.(2022)Madabushi, Gow-Smith, Garcia, Scarton, Idiart,
  and Villavicencio]{madabushi2022semeval}
Harish~Tayyar Madabushi, Edward Gow-Smith, Marcos Garcia, Carolina Scarton,
  Marco Idiart, and Aline Villavicencio.
\newblock Semeval-2022 task 2: Multilingual idiomaticity detection and sentence
  embedding.
\newblock \emph{arXiv preprint arXiv:2204.10050}, 2022.

\bibitem[Mann and Whitney(1947)]{mann1947controlling}
H~Mann and D~Whitney.
\newblock Controlling the false discovery rate: A practical and powerful
  approach to multiple testing.
\newblock \emph{Ann. Math. Stat}, 18\penalty0 (1):\penalty0 50--60, 1947.

\bibitem[Martin et~al.(2020)Martin, Muller, Ortiz~Su{\'a}rez, Dupont, Romary,
  de~la Clergerie, Seddah, and Sagot]{martin2020camembert}
Louis Martin, Benjamin Muller, Pedro~Javier Ortiz~Su{\'a}rez, Yoann Dupont,
  Laurent Romary, {\'E}ric de~la Clergerie, Djam{\'e} Seddah, and Beno{\^\i}t
  Sagot.
\newblock {C}amem{BERT}: a tasty {F}rench language model.
\newblock In \emph{Proceedings of the 58th Annual Meeting of the Association
  for Computational Linguistics}, pages 7203--7219, Online, July 2020.
  Association for Computational Linguistics.
\newblock URL \url{https://www.aclweb.org/anthology/2020.acl-main.645}.

\bibitem[McMillan-Major et~al.(2022)McMillan-Major, Alyafeai, Biderman, Chen,
  De~Toni, Dupont, Elsahar, Emezue, Aji, Ilić, Khamis, Leong, Masoud, Soroa,
  Suarez, Talat, van Strien, and Jernite]{mcmillan-major2022documenting}
Angelina McMillan-Major, Zaid Alyafeai, Stella Biderman, Kimbo Chen, Francesco
  De~Toni, Gérard Dupont, Hady Elsahar, Chris Emezue, Alham~Fikri Aji, Suzana
  Ilić, Nurulaqilla Khamis, Colin Leong, Maraim Masoud, Aitor Soroa,
  Pedro~Ortiz Suarez, Zeerak Talat, Daniel van Strien, and Yacine Jernite.
\newblock Documenting geographically and contextually diverse data sources: The
  bigscience catalogue of language data and resources, 2022.
\newblock URL \url{https://arxiv.org/abs/2201.10066}.

\bibitem[Micikevicius et~al.(2018)Micikevicius, Narang, Alben, Diamos, Elsen,
  Garcia, Ginsburg, Houston, Kuchaiev, Venkatesh, and
  Wu]{micikevicius2018mixed}
Paulius Micikevicius, Sharan Narang, Jonah Alben, Gregory Diamos, Erich Elsen,
  David Garcia, Boris Ginsburg, Michael Houston, Oleksii Kuchaiev, Ganesh
  Venkatesh, and Hao Wu.
\newblock Mixed precision training.
\newblock In \emph{International Conference on Learning Representations}, 2018.
\newblock URL \url{https://openreview.net/forum?id=r1gs9JgRZ}.

\bibitem[Mielke et~al.(2021)Mielke, Alyafeai, Salesky, Raffel, Dey, Gallé,
  Raja, Si, Lee, Sagot, and Tan]{mielke2021between}
Sabrina~J. Mielke, Zaid Alyafeai, Elizabeth Salesky, Colin Raffel, Manan Dey,
  Matthias Gallé, Arun Raja, Chenglei Si, Wilson~Y. Lee, Benoît Sagot, and
  Samson Tan.
\newblock Between words and characters: A brief history of open-vocabulary
  modeling and tokenization in nlp, 2021.
\newblock URL \url{https://arxiv.org/abs/2112.10508}.

\bibitem[Miikkulainen and Dyer(1991)]{miikkulainen1991natural}
Risto Miikkulainen and Michael~G. Dyer.
\newblock Natural language processing with modular pdp networks and distributed
  lexicon.
\newblock \emph{Cognitive Science}, 15\penalty0 (3), 1991.

\bibitem[Mikolov et~al.(2010)Mikolov, Karafi{\'a}t, Burget, Cernock{\`y}, and
  Khudanpur]{mikolov2010recurrent}
Tomas Mikolov, Martin Karafi{\'a}t, Lukas Burget, Jan Cernock{\`y}, and Sanjeev
  Khudanpur.
\newblock Recurrent neural network based language model.
\newblock In \emph{Interspeech}, 2010.

\bibitem[Mikolov et~al.(2013)Mikolov, Sutskever, Chen, Corrado, and
  Dean]{mikolov2013distributed}
Tomas Mikolov, Ilya Sutskever, Kai Chen, Greg~S. Corrado, and Jeff Dean.
\newblock Distributed representations of words and phrases and their
  compositionality.
\newblock \emph{Advances in neural information processing systems}, 26, 2013.

\bibitem[Mitchell et~al.(2019)Mitchell, Wu, Zaldivar, Barnes, Vasserman,
  Hutchinson, Spitzer, Raji, and Gebru]{ModelCard}
Margaret Mitchell, Simone Wu, Andrew Zaldivar, Parker Barnes, Lucy Vasserman,
  Ben Hutchinson, Elena Spitzer, Inioluwa~Deborah Raji, and Timnit Gebru.
\newblock Model cards for model reporting.
\newblock In \emph{Proceedings of the Conference on Fairness, Accountability,
  and Transparency}, FAT* '19, page 220–229, New York, NY, USA, 2019.
  Association for Computing Machinery.
\newblock ISBN 9781450361255.
\newblock \doi{10.1145/3287560.3287596}.
\newblock URL \url{https://doi.org/10.1145/3287560.3287596}.

\bibitem[Moi et~al.(2019)Moi, Cistac, Patry, Walsh, Morgan, Pütz, Wolf,
  Gugger, Delangue, Chaumond, Debut, and von Platen]{hftokenizers2019}
Anthony Moi, Pierric Cistac, Nicolas Patry, Evan~P. Walsh, Funtowicz Morgan,
  Sebastian Pütz, Thomas Wolf, Sylvain Gugger, Clément Delangue, Julien
  Chaumond, Lysandre Debut, and Patrick von Platen.
\newblock Hugging face tokenizers library.
\newblock \url{https://github.com/huggingface/tokenizers}, 2019.

\bibitem[Mostafazadeh et~al.(2017)Mostafazadeh, Roth, Louis, Chambers, and
  Allen]{mostafazadeh2017lsdsem}
Nasrin Mostafazadeh, Michael Roth, Annie Louis, Nathanael Chambers, and James
  Allen.
\newblock Lsdsem 2017 shared task: The story cloze test.
\newblock In \emph{Proceedings of the 2nd Workshop on Linking Models of
  Lexical, Sentential and Discourse-level Semantics}, pages 46--51, 2017.

\bibitem[Muennighoff(2022)]{muennighoff2022sgpt}
Niklas Muennighoff.
\newblock {SGPT}: {GPT} sentence embeddings for semantic search.
\newblock \emph{arXiv preprint arXiv:2202.08904}, 2022.

\bibitem[Muennighoff et~al.(2022{\natexlab{a}})Muennighoff, Tazi, Magne, and
  Reimers]{muennighoff2022mteb}
Niklas Muennighoff, Nouamane Tazi, Lo{\"\i}c Magne, and Nils Reimers.
\newblock {MTEB}: Massive text embedding benchmark.
\newblock \emph{arXiv preprint arXiv:2210.07316}, 2022{\natexlab{a}}.

\bibitem[Muennighoff et~al.(2022{\natexlab{b}})Muennighoff, Wang, Sutawika,
  Roberts, Biderman, Scao, Bari, Shen, Yong, Schoelkopf,
  et~al.]{muennighoff2022crosslingual}
Niklas Muennighoff, Thomas Wang, Lintang Sutawika, Adam Roberts, Stella
  Biderman, Teven~Le Scao, M~Saiful Bari, Sheng Shen, Zheng-Xin Yong, Hailey
  Schoelkopf, et~al.
\newblock Crosslingual generalization through multitask finetuning.
\newblock \emph{arXiv preprint arXiv:2211.01786}, 2022{\natexlab{b}}.

\bibitem[Nangia et~al.(2020)Nangia, Vania, Bhalerao, and
  Bowman]{nangia-etal-2020-crows}
Nikita Nangia, Clara Vania, Rasika Bhalerao, and Samuel~R. Bowman.
\newblock {C}row{S}-pairs: A challenge dataset for measuring social biases in
  masked language models.
\newblock In \emph{Proceedings of the 2020 Conference on Empirical Methods in
  Natural Language Processing (EMNLP)}, pages 1953--1967, Online, November
  2020. Association for Computational Linguistics.
\newblock \doi{10.18653/v1/2020.emnlp-main.154}.
\newblock URL \url{https://aclanthology.org/2020.emnlp-main.154}.

\bibitem[Narang et~al.(2021)Narang, Chung, Tay, Fedus, Fevry, Matena, Malkan,
  Fiedel, Shazeer, Lan, Zhou, Li, Ding, Marcus, Roberts, and
  Raffel]{narang2021transformer}
Sharan Narang, Hyung~Won Chung, Yi~Tay, William Fedus, Thibault Fevry, Michael
  Matena, Karishma Malkan, Noah Fiedel, Noam Shazeer, Zhenzhong Lan, Yanqi
  Zhou, Wei Li, Nan Ding, Jake Marcus, Adam Roberts, and Colin Raffel.
\newblock Do transformer modifications transfer across implementations and
  applications?
\newblock In \emph{Conference on Empirical Methods in Natural Language
  Processing}, 2021.

\bibitem[Narayanan et~al.(2021)Narayanan, Shoeybi, Casper, LeGresley, Patwary,
  Korthikanti, Vainbrand, Kashinkunti, Bernauer, Catanzaro, Phanishayee, and
  Zaharia]{narayanan2021efficient}
Deepak Narayanan, Mohammad Shoeybi, Jared Casper, Patrick LeGresley, Mostofa
  Patwary, Vijay Korthikanti, Dmitri Vainbrand, Prethvi Kashinkunti, Julie
  Bernauer, Bryan Catanzaro, Amar Phanishayee, and Matei Zaharia.
\newblock {Efficient Large-Scale Language Model Training on GPU Clusters using
  Megatron-LM}.
\newblock In \emph{Proceedings of the International Conference for High
  Performance Computing, Networking, Storage and Analysis}, 2021.

\bibitem[Nekoto et~al.(2020)Nekoto, Marivate, Matsila, Fasubaa, Kolawole,
  Fagbohungbe, Akinola, Muhammad, Kabenamualu, Osei, Freshia, Niyongabo,
  Macharm, Ogayo, Ahia, Meressa, Adeyemi, Mokgesi-Selinga, Okegbemi, Martinus,
  Tajudeen, Degila, Ogueji, Siminyu, Kreutzer, Webster, Ali, Abbott, Orife,
  Ezeani, Dangana, Kamper, ElSahar, Duru, Kioko, Murhabazi, Biljon, Whitenack,
  Onyefuluchi, Emezue, Dossou, Sibanda, Bassey, Olabiyi, Ramkilowan, Oktem,
  Akinfaderin, and Bashir]{nekoto2020participatory}
Wilhelmina Nekoto, Vukosi Marivate, Tshinondiwa Matsila, Timi~E. Fasubaa,
  T~Kolawole, Taiwo~Helen Fagbohungbe, Solomon~Oluwole Akinola,
  Shamsuddeen~Hassan Muhammad, Salomon~Kabongo Kabenamualu, Salomey Osei,
  Sackey Freshia, Rubungo~Andre Niyongabo, Ricky Macharm, Perez Ogayo,
  Orevaoghene Ahia, Musie Meressa, Mofetoluwa Adeyemi, Masabata
  Mokgesi-Selinga, Lawrence Okegbemi, Laura Martinus, Kolawole Tajudeen, Kevin
  Degila, Kelechi Ogueji, Kathleen Siminyu, Julia Kreutzer, Jason Webster,
  Jamiil~Toure Ali, Jade~Z. Abbott, Iroro Orife, Ignatius~U. Ezeani,
  Idris~Abdulkabir Dangana, Herman Kamper, Hady ElSahar, Goodness Duru, Ghollah
  Kioko, Espoir Murhabazi, Elan~Van Biljon, Daniel Whitenack, Christopher
  Onyefuluchi, Chris~C. Emezue, Bonaventure F.~P. Dossou, Blessing~K. Sibanda,
  Blessing~Itoro Bassey, Ayodele Olabiyi, Arshath Ramkilowan, Alp Oktem,
  Adewale Akinfaderin, and Abdallah~M. Bashir.
\newblock Participatory research for low-resourced machine translation: A case
  study in {African} languages.
\newblock In \emph{ACL Findings}, 2020.

\bibitem[N{\'e}v{\'e}ol et~al.(2022)N{\'e}v{\'e}ol, Dupont, Bezan{\c{c}}on, and
  Fort]{neveol-etal-2022-french}
Aur{\'e}lie N{\'e}v{\'e}ol, Yoann Dupont, Julien Bezan{\c{c}}on, and Kar{\"e}n
  Fort.
\newblock {F}rench {C}row{S}-pairs: Extending a challenge dataset for measuring
  social bias in masked language models to a language other than {E}nglish.
\newblock In \emph{Proceedings of the 60th Annual Meeting of the Association
  for Computational Linguistics (Volume 1: Long Papers)}, pages 8521--8531,
  Dublin, Ireland, May 2022. Association for Computational Linguistics.
\newblock \doi{10.18653/v1/2022.acl-long.583}.
\newblock URL \url{https://aclanthology.org/2022.acl-long.583}.

\bibitem[Nivre et~al.(2016)Nivre, de~Marneffe, Ginter, Goldberg, Haji{\v{c}},
  Manning, McDonald, Petrov, Pyysalo, Silveira, Tsarfaty, and
  Zeman]{nivre-etal-2016-universal}
Joakim Nivre, Marie-Catherine de~Marneffe, Filip Ginter, Yoav Goldberg, Jan
  Haji{\v{c}}, Christopher~D. Manning, Ryan McDonald, Slav Petrov, Sampo
  Pyysalo, Natalia Silveira, Reut Tsarfaty, and Daniel Zeman.
\newblock {U}niversal {D}ependencies v1: A multilingual treebank collection.
\newblock In \emph{Proceedings of the Tenth International Conference on
  Language Resources and Evaluation ({LREC}'16)}, pages 1659--1666,
  Portoro{\v{z}}, Slovenia, May 2016. European Language Resources Association
  (ELRA).
\newblock URL \url{https://aclanthology.org/L16-1262}.

\bibitem[Nivre et~al.(2017)Nivre, Zeman, Ginter, and
  Tyers]{nivre-etal-2017-universal}
Joakim Nivre, Daniel Zeman, Filip Ginter, and Francis Tyers.
\newblock {U}niversal {D}ependencies.
\newblock In \emph{Proceedings of the 15th Conference of the {E}uropean Chapter
  of the Association for Computational Linguistics: Tutorial Abstracts},
  Valencia, Spain, April 2017. Association for Computational Linguistics.
\newblock URL \url{https://aclanthology.org/E17-5001}.

\bibitem[{Ortiz Su{\'a}rez} et~al.(2019){Ortiz Su{\'a}rez}, Sagot, and
  Romary]{OSCAR}
Pedro~Javier {Ortiz Su{\'a}rez}, Beno{\^\i}t Sagot, and Laurent Romary.
\newblock Asynchronous pipelines for processing huge corpora on medium to low
  resource infrastructures.
\newblock In Piotr Bański, Adrien Barbaresi, Hanno Biber, Evelyn Breiteneder,
  Simon Clematide, Marc Kupietz, Harald L{\"u}ngen, and Caroline Iliadi,
  editors, \emph{Proceedings of the Workshop on Challenges in the Management of
  Large Corpora (CMLC-7)}, pages 9 -- 16, Cardiff, UK, 2019. Leibniz-Institut
  f{\"u}r Deutsche Sprache.
\newblock \doi{10.14618/ids-pub-9021}.
\newblock URL \url{http://nbn-resolving.de/urn:nbn:de:bsz:mh39-90215}.

\bibitem[Papineni et~al.(2002)Papineni, Roukos, Ward, and
  Zhu]{papineni-etal-2002-bleu}
Kishore Papineni, Salim Roukos, Todd Ward, and Wei-Jing Zhu.
\newblock {BLEU}: a method for automatic evaluation of machine translation.
\newblock In \emph{Proceedings of the 40th Annual Meeting of the Association
  for Computational Linguistics}, pages 311--318, Philadelphia, Pennsylvania,
  USA, July 2002. Association for Computational Linguistics.
\newblock \doi{10.3115/1073083.1073135}.
\newblock URL \url{https://aclanthology.org/P02-1040}.

\bibitem[Patterson et~al.(2021)Patterson, Gonzalez, Le, Liang, Munguia,
  Rothchild, So, Texier, and Dean]{patterson2021carbon}
David Patterson, Joseph Gonzalez, Quoc Le, Chen Liang, Lluis-Miquel Munguia,
  Daniel Rothchild, David So, Maud Texier, and Jeff Dean.
\newblock Carbon emissions and large neural network training.
\newblock \emph{arXiv preprint arXiv:2104.10350}, 2021.

\bibitem[Pearson(1895)]{pearson1895vii}
Karl Pearson.
\newblock Note on regression and inheritance in the case of two parents.
\newblock \emph{Proceedings of the Royal Society of London}, 58\penalty0
  (347-352):\penalty0 240--242, 1895.

\bibitem[Peters et~al.(2018)Peters, Neumann, Iyyer, Gardner, Clark, Lee, and
  Zettlemoyer]{peters2018deep}
Matthew~E. Peters, Mark Neumann, Mohit Iyyer, Matt Gardner, Christopher Clark,
  Kenton Lee, and Luke Zettlemoyer.
\newblock Deep contextualized word representations.
\newblock In \emph{Conference of the North American Chapter of the Association
  for Computational Linguistics}, 2018.

\bibitem[Phang et~al.(2022)Phang, Bradley, Gao, Castricato, and
  Biderman]{phang2022eleutherai}
Jason Phang, Herbie Bradley, Leo Gao, Louis~J Castricato, and Stella Biderman.
\newblock {EleutherAI:} going beyond "open science" to "science in the open".
\newblock In \emph{Workshop on Broadening Research Collaborations}, 2022.

\bibitem[Post(2018)]{post-2018-call}
Matt Post.
\newblock A call for clarity in reporting {BLEU} scores.
\newblock In \emph{Proceedings of the Third Conference on Machine Translation:
  Research Papers}, pages 186--191, Brussels, Belgium, October 2018.
  Association for Computational Linguistics.
\newblock \doi{10.18653/v1/W18-6319}.
\newblock URL \url{https://aclanthology.org/W18-6319}.

\bibitem[Press et~al.(2021)Press, Smith, and Lewis]{press2021train}
Ofir Press, Noah Smith, and Mike Lewis.
\newblock Train short, test long: Attention with linear biases enables input
  length extrapolation.
\newblock In \emph{International Conference on Learning Representations}, 2021.

\bibitem[Radford et~al.(2018)Radford, Narasimhan, Salimans, and
  Sutskever]{radford2018improving}
Alec Radford, Karthik Narasimhan, Tim Salimans, and Ilya Sutskever.
\newblock Improving language understanding by generative pre-training, 2018.

\bibitem[Radford et~al.(2019)Radford, Wu, Child, Luan, Amodei, and
  Sutskever]{radford2019language}
Alec Radford, Jeffrey Wu, Rewon Child, David Luan, Dario Amodei, and Ilya
  Sutskever.
\newblock Language models are unsupervised multitask learners, 2019.

\bibitem[Rae et~al.(2021)Rae, Borgeaud, Cai, Millican, Hoffmann, Song,
  Aslanides, Henderson, Ring, Young, et~al.]{rae2021scaling}
Jack~W Rae, Sebastian Borgeaud, Trevor Cai, Katie Millican, Jordan Hoffmann,
  Francis Song, John Aslanides, Sarah Henderson, Roman Ring, Susannah Young,
  et~al.
\newblock Scaling language models: Methods, analysis \& insights from training
  gopher.
\newblock \emph{arXiv preprint arXiv:2112.11446}, 2021.

\bibitem[Raffel et~al.(2020)Raffel, Shazeer, Roberts, Lee, Narang, Matena,
  Zhou, Li, Liu, et~al.]{raffel2020exploring}
Colin Raffel, Noam Shazeer, Adam Roberts, Katherine Lee, Sharan Narang, Michael
  Matena, Yanqi Zhou, Wei Li, Peter~J Liu, et~al.
\newblock Exploring the limits of transfer learning with a unified text-to-text
  transformer.
\newblock \emph{J. Mach. Learn. Res.}, 21\penalty0 (140):\penalty0 1--67, 2020.

\bibitem[Rajbhandari et~al.(2020)Rajbhandari, Rasley, Ruwase, and
  He]{rajbhandari2020zero}
Samyam Rajbhandari, Jeff Rasley, Olatunji Ruwase, and Yuxiong He.
\newblock {ZeRO}: Memory optimizations toward training trillion parameter
  models.
\newblock \emph{SC20: International Conference for High Performance Computing,
  Networking, Storage and Analysis}, Nov 2020.
\newblock \doi{10.1109/sc41405.2020.00024}.
\newblock URL \url{http://dx.doi.org/10.1109/SC41405.2020.00024}.

\bibitem[Raji et~al.(2021)Raji, Denton, Bender, Hanna, and
  Paullada]{raji_AI_2021}
Deborah Raji, Emily Denton, Emily~M. Bender, Alex Hanna, and Amandalynne
  Paullada.
\newblock Ai and the everything in the whole wide world benchmark.
\newblock In J.~Vanschoren and S.~Yeung, editors, \emph{Proceedings of the
  Neural Information Processing Systems Track on Datasets and Benchmarks},
  volume~1, 2021.
\newblock URL
  \url{https://datasets-benchmarks-proceedings.neurips.cc/paper/2021/file/084b6fbb10729ed4da8c3d3f5a3ae7c9-Paper-round2.pdf}.

\bibitem[Raji et~al.(2022)Raji, Kumar, Horowitz, and Selbst]{raji2022fallacy}
Inioluwa~Deborah Raji, I.~Elizabeth Kumar, Aaron Horowitz, and Andrew Selbst.
\newblock The fallacy of {AI} functionality.
\newblock In \emph{2022 ACM Conference on Fairness, Accountability, and
  Transparency}, FAccT '22, page 959–972, New York, NY, USA, 2022.
  Association for Computing Machinery.
\newblock ISBN 9781450393522.
\newblock \doi{10.1145/3531146.3533158}.
\newblock URL \url{https://doi.org/10.1145/3531146.3533158}.

\bibitem[Rasley et~al.(2020)Rasley, Rajbhandari, Ruwase, and
  He]{rasley2020deepspeed}
Jeff Rasley, Samyam Rajbhandari, Olatunji Ruwase, and Yuxiong He.
\newblock {DeepSpeed}: System optimizations enable training deep learning
  models with over 100 billion parameters.
\newblock In \emph{Proceedings of the 26th ACM SIGKDD International Conference
  on Knowledge Discovery \& Data Mining}, KDD '20, page 3505–3506, New York,
  NY, USA, 2020. Association for Computing Machinery.
\newblock ISBN 9781450379984.
\newblock \doi{10.1145/3394486.3406703}.
\newblock URL \url{https://doi.org/10.1145/3394486.3406703}.

\bibitem[Rust et~al.(2021)Rust, Pfeiffer, Vuli{\'c}, Ruder, and
  Gurevych]{rust-etal-2021-good}
Phillip Rust, Jonas Pfeiffer, Ivan Vuli{\'c}, Sebastian Ruder, and Iryna
  Gurevych.
\newblock How good is your tokenizer? on the monolingual performance of
  multilingual language models.
\newblock In \emph{Proceedings of the 59th Annual Meeting of the Association
  for Computational Linguistics and the 11th International Joint Conference on
  Natural Language Processing (Volume 1: Long Papers)}, pages 3118--3135,
  Online, August 2021. Association for Computational Linguistics.
\newblock \doi{10.18653/v1/2021.acl-long.243}.
\newblock URL \url{https://aclanthology.org/2021.acl-long.243}.

\bibitem[Safaya et~al.(2020)Safaya, Abdullatif, and Yuret]{safaya2020kuisail}
Ali Safaya, Moutasem Abdullatif, and Deniz Yuret.
\newblock {KUISAIL} at {S}em{E}val-2020 task 12: {BERT}-{CNN} for offensive
  speech identification in social media.
\newblock In \emph{Proceedings of the Fourteenth Workshop on Semantic
  Evaluation}, pages 2054--2059, Barcelona (online), December 2020.
  International Committee for Computational Linguistics.
\newblock URL \url{https://www.aclweb.org/anthology/2020.semeval-1.271}.

\bibitem[Salton and Yang(1973)]{salton1973specification}
Gerard Salton and Chung-Shu Yang.
\newblock On the specification of term values in automatic indexing.
\newblock \emph{Journal of documentation}, 1973.

\bibitem[Sambasivan et~al.(2021)Sambasivan, Kapania, Highfill, Akrong,
  Paritosh, and Aroyo]{sambasivan2021cascades}
Nithya Sambasivan, Shivani Kapania, Hannah Highfill, Diana Akrong, Praveen
  Paritosh, and Lora~M Aroyo.
\newblock “everyone wants to do the model work, not the data work”: Data
  cascades in high-stakes ai.
\newblock In \emph{Proceedings of the 2021 CHI Conference on Human Factors in
  Computing Systems}, CHI '21, New York, NY, USA, 2021. Association for
  Computing Machinery.
\newblock ISBN 9781450380966.
\newblock \doi{10.1145/3411764.3445518}.
\newblock URL \url{https://doi.org/10.1145/3411764.3445518}.

\bibitem[Sanh et~al.(2022)Sanh, Webson, Raffel, Bach, Sutawika, Alyafeai,
  Chaffin, Stiegler, Raja, Dey, Bari, Xu, Thakker, Sharma, Szczechla, Kim,
  Chhablani, Nayak, Datta, Chang, Jiang, Wang, Manica, Shen, Yong, Pandey,
  Bawden, Wang, Neeraj, Rozen, Sharma, Santilli, Fevry, Fries, Teehan, Scao,
  Biderman, Gao, Wolf, and Rush]{sanh2022multitask}
Victor Sanh, Albert Webson, Colin Raffel, Stephen Bach, Lintang Sutawika, Zaid
  Alyafeai, Antoine Chaffin, Arnaud Stiegler, Arun Raja, Manan Dey, M~Saiful
  Bari, Canwen Xu, Urmish Thakker, Shanya~Sharma Sharma, Eliza Szczechla,
  Taewoon Kim, Gunjan Chhablani, Nihal Nayak, Debajyoti Datta, Jonathan Chang,
  Mike Tian-Jian Jiang, Han Wang, Matteo Manica, Sheng Shen, Zheng~Xin Yong,
  Harshit Pandey, Rachel Bawden, Thomas Wang, Trishala Neeraj, Jos Rozen,
  Abheesht Sharma, Andrea Santilli, Thibault Fevry, Jason~Alan Fries, Ryan
  Teehan, Teven~Le Scao, Stella Biderman, Leo Gao, Thomas Wolf, and Alexander~M
  Rush.
\newblock Multitask prompted training enables zero-shot task generalization.
\newblock In \emph{International Conference on Learning Representations}, 2022.
\newblock URL \url{https://openreview.net/forum?id=9Vrb9D0WI4}.

\bibitem[Schmidhuber and Heil(1996)]{schmidhuber1996sequential}
J{\"u}rgen Schmidhuber and Stefan Heil.
\newblock Sequential neural text compression.
\newblock \emph{IEEE Transactions on Neural Networks}, 7\penalty0 (1), 1996.

\bibitem[Schwartz et~al.(2020)Schwartz, Dodge, Smith, and
  Etzioni]{schwartz2020green}
Roy Schwartz, Jesse Dodge, Noah~A. Smith, and Oren Etzioni.
\newblock Green ai.
\newblock \emph{Communications of the ACM}, 63\penalty0 (12), 2020.

\bibitem[Serikov et~al.(2022)Serikov, Protasov, Voloshina, Knyazkova, and
  Shavrina]{serikov2022universal}
Oleg Serikov, Vitaly Protasov, Ekaterina Voloshina, Viktoria Knyazkova, and
  Tatiana Shavrina.
\newblock Universal and independent: Multilingual probing framework for
  exhaustive model interpretation and evaluation.
\newblock \emph{arXiv preprint arXiv:2210.13236}, 2022.

\bibitem[Shannon(1948)]{shannon1948mathematical}
Claude~Elwood Shannon.
\newblock A mathematical theory of communication.
\newblock \emph{The Bell system technical journal}, 27\penalty0 (3), 1948.

\bibitem[Shazeer(2020)]{shazeer2020glu}
Noam Shazeer.
\newblock {GLU} variants improve transformer.
\newblock \emph{arXiv preprint arXiv:2002.05202}, 2020.

\bibitem[Shazeer et~al.(2017)Shazeer, Mirhoseini, Maziarz, Davis, Le, Hinton,
  and Dean]{shazeer2017}
Noam Shazeer, Azalia Mirhoseini, Krzysztof Maziarz, Andy Davis, Quoc Le,
  Geoffrey Hinton, and Jeff Dean.
\newblock Outrageously large neural networks: The sparsely-gated
  mixture-of-experts layer.
\newblock In \emph{International Conference on Learning Representations}, 2017.
\newblock URL \url{https://openreview.net/forum?id=B1ckMDqlg}.

\bibitem[Shliazhko et~al.(2022)Shliazhko, Fenogenova, Tikhonova, Mikhailov,
  Kozlova, and Shavrina]{shliazhko2022mgpt}
Oleh Shliazhko, Alena Fenogenova, Maria Tikhonova, Vladislav Mikhailov,
  Anastasia Kozlova, and Tatiana Shavrina.
\newblock mgpt: Few-shot learners go multilingual.
\newblock \emph{arXiv preprint arXiv:2204.07580}, 2022.

\bibitem[Shoeybi et~al.(2019)Shoeybi, Patwary, Puri, LeGresley, Casper, and
  Catanzaro]{shoeybi2019megatron}
Mohammad Shoeybi, Mostofa Patwary, Raul Puri, Patrick LeGresley, Jared Casper,
  and Bryan Catanzaro.
\newblock {Megatron-LM}: Training multi-billion parameter language models using
  model parallelism.
\newblock \emph{arXiv preprint arXiv:1909.08053}, 2019.

\bibitem[Simoulin and Crabb{\'e}(2021)]{simoulin:hal-03265900}
Antoine Simoulin and Benoit Crabb{\'e}.
\newblock {Un mod{\`e}le Transformer G{\'e}n{\'e}ratif Pr{\'e}-entrain{\'e}
  pour le \_\_\_\_\_\_ fran{\c c}ais}.
\newblock In Pascal Denis, Natalia Grabar, Amel Fraisse, R{\'e}mi Cardon,
  Bernard Jacquemin, Eric Kergosien, and Antonio Balvet, editors,
  \emph{{Traitement Automatique des Langues Naturelles}}, pages 246--255,
  Lille, France, 2021. {ATALA}.
\newblock URL \url{https://hal.archives-ouvertes.fr/hal-03265900}.

\bibitem[Smith et~al.(2022)Smith, Patwary, Norick, LeGresley, Rajbhandari,
  Casper, Liu, Prabhumoye, Zerveas, Korthikanti, Zhang, Child, Aminabadi,
  Bernauer, Song, Shoeybi, He, Houston, Tiwary, and Catanzaro]{smith2022using}
Shaden Smith, Mostofa Patwary, Brandon Norick, Patrick LeGresley, Samyam
  Rajbhandari, Jared Casper, Zhun Liu, Shrimai Prabhumoye, George Zerveas,
  Vijay Korthikanti, Elton Zhang, Rewon Child, Reza~Yazdani Aminabadi, Julie
  Bernauer, Xia Song, Mohammad Shoeybi, Yuxiong He, Michael Houston, Saurabh
  Tiwary, and Bryan Catanzaro.
\newblock Using {DeepSpeed} and {Megatron} to train {Megatron-Turing NLG 530B},
  a large-scale generative language model.
\newblock \emph{arXiv preprint arXiv:2201.11990}, 2022.

\bibitem[Soltan et~al.(2022)Soltan, Ananthakrishnan, FitzGerald, Gupta, Hamza,
  Khan, Peris, Rawls, Rosenbaum, Rumshisky, Prakash, Sridhar, Triefenbach,
  Verma, Tur, and Natarajan]{soltan-etal-2022-alexatm}
Saleh Soltan, Shankar Ananthakrishnan, Jack FitzGerald, Rahul Gupta, Wael
  Hamza, Haidar Khan, Charith Peris, Stephen Rawls, Andy Rosenbaum, Anna
  Rumshisky, Chandana~Satya Prakash, Mukund Sridhar, Fabian Triefenbach, Apurv
  Verma, Gokhan Tur, and Prem Natarajan.
\newblock Alexatm 20b: Few-shot learning using a large-scale multilingual
  seq2seq model, 2022.
\newblock URL \url{https://arxiv.org/abs/2208.01448}.

\bibitem[Srivastava et~al.(2022)Srivastava, Rastogi, Rao, Shoeb, Abid, Fisch,
  Brown, Santoro, Gupta, Garriga-Alonso, et~al.]{srivastava2022beyond}
Aarohi Srivastava, Abhinav Rastogi, Abhishek Rao, Abu Awal~Md Shoeb, Abubakar
  Abid, Adam Fisch, Adam~R Brown, Adam Santoro, Aditya Gupta, Adri{\`a}
  Garriga-Alonso, et~al.
\newblock Beyond the imitation game: Quantifying and extrapolating the
  capabilities of language models.
\newblock \emph{arXiv preprint arXiv:2206.04615}, 2022.

\bibitem[Strubell et~al.(2019)Strubell, Ganesh, and
  McCallum]{strubell2019energy}
Emma Strubell, Ananya Ganesh, and Andrew McCallum.
\newblock Energy and policy considerations for deep learning in nlp.
\newblock In \emph{Annual Meeting of the Association for Computational
  Linguistics}, 2019.

\bibitem[Su et~al.(2021)Su, Lu, Pan, Wen, and Liu]{su2021roformer}
Jianlin Su, Yu~Lu, Shengfeng Pan, Bo~Wen, and Yunfeng Liu.
\newblock {RoFormer}: Enhanced transformer with rotary position embedding.
\newblock \emph{arXiv preprint arXiv:2104.09864}, 2021.

\bibitem[Sutskever et~al.(2011)Sutskever, Martens, and
  Hinton]{sutskever2011generating}
Ilya Sutskever, James Martens, and Geoffrey~E. Hinton.
\newblock Generating text with recurrent neural networks.
\newblock In \emph{International Conference on Machine Learning}, 2011.

\bibitem[Talat et~al.(2022)Talat, N{\'e}v{\'e}ol, Biderman, Clinciu, Dey,
  Longpre, Luccioni, Masoud, Mitchell, Radev, Sharma, Subramonian, Tae, Tan,
  Tunuguntla, and van~der Wal]{talat2022you}
Zeerak Talat, Aur{\'e}lie N{\'e}v{\'e}ol, Stella Biderman, Miruna Clinciu,
  Manan Dey, Shayne Longpre, Sasha Luccioni, Maraim Masoud, Margaret Mitchell,
  Dragomir Radev, Shanya Sharma, Arjun Subramonian, Jaesung Tae, Samson Tan,
  Deepak Tunuguntla, and Oskar van~der Wal.
\newblock You reap what you sow: On the challenges of bias evaluation under
  multilingual settings.
\newblock In \emph{Challenges {\&} Perspectives in Creating Large Language
  Models}, 2022.
\newblock URL \url{https://openreview.net/forum?id=rK-7NhfSIW5}.

\bibitem[Tay et~al.(2022)Tay, Wei, Chung, Tran, So, Shakeri, Garcia, Zheng,
  Rao, Chowdhery, et~al.]{tay2022transcending}
Yi~Tay, Jason Wei, Hyung~Won Chung, Vinh~Q Tran, David~R So, Siamak Shakeri,
  Xavier Garcia, Huaixiu~Steven Zheng, Jinfeng Rao, Aakanksha Chowdhery, et~al.
\newblock Transcending scaling laws with 0.1\% extra compute.
\newblock \emph{arXiv preprint arXiv:2210.11399}, 2022.

\bibitem[Teehan et~al.(2022)Teehan, Clinciu, Serikov, Szczechla, Seelam,
  Mirkin, and Gokaslan]{teehan-etal-2022-emergent}
Ryan Teehan, Miruna Clinciu, Oleg Serikov, Eliza Szczechla, Natasha Seelam,
  Shachar Mirkin, and Aaron Gokaslan.
\newblock Emergent structures and training dynamics in large language models.
\newblock In \emph{Proceedings of BigScience Episode {\#}5 -- Workshop on
  Challenges {\&} Perspectives in Creating Large Language Models}, pages
  146--159, virtual+Dublin, May 2022. Association for Computational
  Linguistics.
\newblock \doi{10.18653/v1/2022.bigscience-1.11}.
\newblock URL \url{https://aclanthology.org/2022.bigscience-1.11}.

\bibitem[Tenney et~al.(2018)Tenney, Xia, Chen, Wang, Poliak, McCoy, Kim,
  Van~Durme, Bowman, Das, et~al.]{tenney2018you}
Ian Tenney, Patrick Xia, Berlin Chen, Alex Wang, Adam Poliak, R~Thomas McCoy,
  Najoung Kim, Benjamin Van~Durme, Samuel~R Bowman, Dipanjan Das, et~al.
\newblock What do you learn from context? probing for sentence structure in
  contextualized word representations.
\newblock In \emph{International Conference on Learning Representations}, 2018.

\bibitem[Vaswani et~al.(2017)Vaswani, Shazeer, Parmar, Uszkoreit, Jones, Gomez,
  Kaiser, and Polosukhin]{vaswani2017attention}
Ashish Vaswani, Noam Shazeer, Niki Parmar, Jakob Uszkoreit, Llion Jones,
  Aidan~N Gomez, {\L}ukasz Kaiser, and Illia Polosukhin.
\newblock Attention is all you need.
\newblock \emph{Advances in neural information processing systems}, 30, 2017.

\bibitem[Vinyals and Le(2015)]{vinyals2015neural}
Oriol Vinyals and Quoc~V. Le.
\newblock A neural conversational model.
\newblock \emph{arXiv preprint arXiv:1506.05869}, 2015.

\bibitem[Voloshina et~al.(2022)Voloshina, Serikov, and
  Shavrina]{voloshina2022neural}
Ekaterina Voloshina, Oleg Serikov, and Tatiana Shavrina.
\newblock Is neural language acquisition similar to natural? a chronological
  probing study.
\newblock \emph{arXiv preprint arXiv:2207.00560}, 2022.

\bibitem[Wang et~al.(2019)Wang, Pruksachatkun, Nangia, Singh, Michael, Hill,
  Levy, and Bowman]{wang-et-al-2019-superglue}
Alex Wang, Yada Pruksachatkun, Nikita Nangia, Amanpreet Singh, Julian Michael,
  Felix Hill, Omer Levy, and Samuel Bowman.
\newblock Superglue: A stickier benchmark for general-purpose language
  understanding systems.
\newblock In H.~Wallach, H.~Larochelle, A.~Beygelzimer, F.~d\textquotesingle
  Alch\'{e}-Buc, E.~Fox, and R.~Garnett, editors, \emph{Advances in Neural
  Information Processing Systems}, volume~32. Curran Associates, Inc., 2019.
\newblock URL
  \url{https://proceedings.neurips.cc/paper/2019/file/4496bf24afe7fab6f046bf4923da8de6-Paper.pdf}.

\bibitem[Wang and Komatsuzaki(2021)]{wang2021gpt}
Ben Wang and Aran Komatsuzaki.
\newblock {GPT-J-6B}: A 6 billion parameter autoregressive language model,
  2021.

\bibitem[Wang et~al.(2020)Wang, Cho, and Gu]{wang2020neural}
Changhan Wang, Kyunghyun Cho, and Jiatao Gu.
\newblock Neural machine translation with byte-level subwords.
\newblock In \emph{Proceedings of the AAAI Conference on Artificial
  Intelligence}, 2020.

\bibitem[Wang and Kanwar(2019)]{bf16blog}
Shibo Wang and Pankaj Kanwar.
\newblock Bfloat16: The secret to high performance on cloud tpus, 2019.
\newblock URL
  \url{https://cloud.google.com/blog/products/ai-machine-learning/bfloat16-the-secret-to-high-performance-on-cloud-tpus}.

\bibitem[Wang et~al.(2021)Wang, Sun, Xiang, Wu, Ding, Gong, Feng, Shang, Zhao,
  Pang, Liu, Chen, Lu, Liu, Wang, Bai, Chen, Zhao, Li, Sun, Yu, Ma, Tian, Wu,
  Wu, Zeng, Li, Gao, and Wang]{wang2021ernie}
Shuohuan Wang, Yu~Sun, Yang Xiang, Zhihua Wu, Siyu Ding, Weibao Gong, Shikun
  Feng, Junyuan Shang, Yanbin Zhao, Chao Pang, Jiaxiang Liu, Xuyi Chen, Yuxiang
  Lu, Weixin Liu, Xi~Wang, Yangfan Bai, Qiuliang Chen, Li~Zhao, Shiyong Li,
  Peng Sun, Dianhai Yu, Yanjun Ma, Hao Tian, Hua Wu, Tian Wu, Wei Zeng, Ge~Li,
  Wen Gao, and Haifeng Wang.
\newblock Ernie 3.0 titan: Exploring larger-scale knowledge enhanced
  pre-training for language understanding and generation.
\newblock \emph{arXiv preprint arXiv:2112.12731}, 2021.

\bibitem[Wang et~al.(2022{\natexlab{a}})Wang, Roberts, Hesslow, Scao, Chung,
  Beltagy, Launay, and Raffel]{wang2022what}
Thomas Wang, Adam Roberts, Daniel Hesslow, Teven~Le Scao, Hyung~Won Chung,
  Iz~Beltagy, Julien Launay, and Colin Raffel.
\newblock What language model architecture and pretraining objective works best
  for zero-shot generalization?
\newblock In Kamalika Chaudhuri, Stefanie Jegelka, Le~Song, Csaba Szepesvari,
  Gang Niu, and Sivan Sabato, editors, \emph{Proceedings of the 39th
  International Conference on Machine Learning}, volume 162 of
  \emph{Proceedings of Machine Learning Research}, pages 22964--22984. PMLR,
  17--23 Jul 2022{\natexlab{a}}.
\newblock URL \url{https://proceedings.mlr.press/v162/wang22u.html}.

\bibitem[Wang et~al.(2022{\natexlab{b}})Wang, Mishra, Alipoormolabashi, Kordi,
  Mirzaei, Arunkumar, Ashok, Dhanasekaran, Naik, Stap,
  et~al.]{wang2022benchmarking}
Yizhong Wang, Swaroop Mishra, Pegah Alipoormolabashi, Yeganeh Kordi, Amirreza
  Mirzaei, Anjana Arunkumar, Arjun Ashok, Arut~Selvan Dhanasekaran, Atharva
  Naik, David Stap, et~al.
\newblock Benchmarking generalization via in-context instructions on 1,600+
  language tasks.
\newblock \emph{arXiv preprint arXiv:2204.07705}, 2022{\natexlab{b}}.

\bibitem[Wei et~al.(2021)Wei, Bosma, Zhao, Guu, Yu, Lester, Du, Dai, and
  Le]{wei2021finetuned}
Jason Wei, Maarten Bosma, Vincent~Y Zhao, Kelvin Guu, Adams~Wei Yu, Brian
  Lester, Nan Du, Andrew~M Dai, and Quoc~V Le.
\newblock Finetuned language models are zero-shot learners.
\newblock \emph{arXiv preprint arXiv:2109.01652}, 2021.

\bibitem[Wei et~al.(2022)Wei, Tay, Bommasani, Raffel, Zoph, Borgeaud, Yogatama,
  Bosma, Zhou, Metzler, Chi, Hashimoto, Vinyals, Liang, Dean, and
  Fedus]{wei2022emergent}
Jason Wei, Yi~Tay, Rishi Bommasani, Colin Raffel, Barret Zoph, Sebastian
  Borgeaud, Dani Yogatama, Maarten Bosma, Denny Zhou, Donald Metzler, Ed~H.
  Chi, Tatsunori Hashimoto, Oriol Vinyals, Percy Liang, Jeff Dean, and William
  Fedus.
\newblock Emergent abilities of large language models.
\newblock \emph{Transactions on Machine Learning Research}, 2022.

\bibitem[Westra and Lawson(2001)]{westra2001faces}
Laura~S. Westra and Bill~E. Lawson.
\newblock \emph{Faces of Environmental Racism: Confronting Issues of Global
  Justice}.
\newblock Rowman \& Littlefield Publishers, 2001.

\bibitem[Winner(1977)]{winner1977technology}
Langdon Winner.
\newblock Technology as master. (book reviews: Autonomous technology.
  technics-out-of-control as a theme in political thought).
\newblock \emph{Science}, 1977.

\bibitem[Winner(2017)]{winner2017artifacts}
Langdon Winner.
\newblock Do artifacts have politics?
\newblock In \emph{Computer Ethics}, pages 177--192. Routledge, 2017.

\bibitem[Wong et~al.(2021)Wong, Otles, Donnelly, Krumm, McCullough,
  DeTroyer-Cooley, Pestrue, Phillips, Konye, Penoza, Ghous, and
  Singh]{wong2021sepsis}
Andrew Wong, Erkin Otles, John~P. Donnelly, Andrew Krumm, Jeffrey McCullough,
  Olivia DeTroyer-Cooley, Justin Pestrue, Marie Phillips, Judy Konye, Carleen
  Penoza, Muhammad Ghous, and Karandeep Singh.
\newblock {External Validation of a Widely Implemented Proprietary Sepsis
  Prediction Model in Hospitalized Patients}.
\newblock \emph{JAMA Internal Medicine}, 181\penalty0 (8):\penalty0 1065--1070,
  08 2021.
\newblock ISSN 2168-6106.
\newblock \doi{10.1001/jamainternmed.2021.2626}.
\newblock URL \url{https://doi.org/10.1001/jamainternmed.2021.2626}.

\bibitem[Wu et~al.(2012)Wu, Diamos, Wang, Cadambi, Yalamanchili, and
  Chakradhar]{kernel_fusion}
Haicheng Wu, Gregory Diamos, Jin Wang, Srihari Cadambi, Sudhakar Yalamanchili,
  and Srimat Chakradhar.
\newblock Optimizing data warehousing applications for {GPUs} using kernel
  fusion/fission.
\newblock In \emph{2012 IEEE 26th International Parallel and Distributed
  Processing Symposium Workshops and PhD Forum}, pages 2433--2442, 2012.
\newblock \doi{10.1109/IPDPSW.2012.300}.

\bibitem[Xue et~al.(2021)Xue, Constant, Roberts, Kale, Al-Rfou, Siddhant,
  Barua, and Raffel]{xue2021mt5}
Linting Xue, Noah Constant, Adam Roberts, Mihir Kale, Rami Al-Rfou, Aditya
  Siddhant, Aditya Barua, and Colin Raffel.
\newblock m{T}5: A massively multilingual pre-trained text-to-text transformer.
\newblock In \emph{Proceedings of the 2021 Conference of the North American
  Chapter of the Association for Computational Linguistics: Human Language
  Technologies}, pages 483--498, Online, June 2021. Association for
  Computational Linguistics.
\newblock \doi{10.18653/v1/2021.naacl-main.41}.
\newblock URL \url{https://aclanthology.org/2021.naacl-main.41}.

\bibitem[Yang et~al.(2019)Yang, Dai, Yang, Carbonell, Salakhutdinov, and
  Le]{yang2019xlnet}
Zhilin Yang, Zihang Dai, Yiming Yang, Jaime Carbonell, Ruslan Salakhutdinov,
  and Quoc~V. Le.
\newblock {XLnet}: Generalized autoregressive pretraining for language
  understanding.
\newblock \emph{Advances in Neural Information Processing Systems}, 2019.

\bibitem[Zeng et~al.(2022)Zeng, Liu, Du, Wang, Lai, Ding, Yang, Xu, Zheng, Xia,
  et~al.]{zeng2022glm}
Aohan Zeng, Xiao Liu, Zhengxiao Du, Zihan Wang, Hanyu Lai, Ming Ding, Zhuoyi
  Yang, Yifan Xu, Wendi Zheng, Xiao Xia, et~al.
\newblock Glm-130b: An open bilingual pre-trained model.
\newblock \emph{arXiv preprint arXiv:2210.02414}, 2022.

\bibitem[Zeng et~al.(2021)Zeng, Ren, Su, Wang, Liao, Wang, Jiang, Yang, Wang,
  Zhang, Li, Gong, Yao, Huang, Wang, Yu, Guo, Yu, Zhang, Wang, Tao, Yan, Yi,
  Peng, Jiang, Zhang, Deng, Zhang, Lin, Zhang, Zhang, Guo, Gu, Fan, Wang, Jin,
  Liu, and Tian]{zeng2021pangu}
Wei Zeng, Xiaozhe Ren, Teng Su, Hui Wang, Yi~Liao, Zhiwei Wang, Xin Jiang,
  ZhenZhang Yang, Kaisheng Wang, Xiaoda Zhang, Chen Li, Ziyan Gong, Yifan Yao,
  Xinjing Huang, Jun Wang, Jianfeng Yu, Qi~Guo, Yue Yu, Yan Zhang, Jin Wang,
  Hengtao Tao, Dasen Yan, Zexuan Yi, Fang Peng, Fangqing Jiang, Han Zhang,
  Lingfeng Deng, Yehong Zhang, Zhe Lin, Chao Zhang, Shaojie Zhang, Mingyue Guo,
  Shanzhi Gu, Gaojun Fan, Yaowei Wang, Xuefeng Jin, Qun Liu, and Yonghong Tian.
\newblock {PanGu}-$\alpha$: Large-scale autoregressive pretrained {Chinese}
  language models with auto-parallel computation.
\newblock \emph{arXiv preprint arXiv:2104.12369}, 2021.

\bibitem[Zhang et~al.(2022)Zhang, Roller, Goyal, Artetxe, Chen, Chen, Dewan,
  Diab, Li, Lin, et~al.]{zhang2022opt}
Susan Zhang, Stephen Roller, Naman Goyal, Mikel Artetxe, Moya Chen, Shuohui
  Chen, Christopher Dewan, Mona Diab, Xian Li, Xi~Victoria Lin, et~al.
\newblock {OPT}: Open pre-trained transformer language models.
\newblock \emph{arXiv preprint arXiv:2205.01068}, 2022.

\bibitem[Zhang et~al.(2021)Zhang, Warstadt, Li, and
  Bowman]{zhang-etal-2021-need}
Yian Zhang, Alex Warstadt, Xiaocheng Li, and Samuel~R. Bowman.
\newblock When do you need billions of words of pretraining data?
\newblock In \emph{Proceedings of the 59th Annual Meeting of the Association
  for Computational Linguistics and the 11th International Joint Conference on
  Natural Language Processing (Volume 1: Long Papers)}, pages 1112--1125,
  Online, August 2021. Association for Computational Linguistics.
\newblock \doi{10.18653/v1/2021.acl-long.90}.
\newblock URL \url{https://aclanthology.org/2021.acl-long.90}.

\bibitem[Zhang et~al.(2019)Zhang, Han, Liu, Jiang, Sun, and
  Liu]{zhang2019ernie}
Zhengyan Zhang, Xu~Han, Zhiyuan Liu, Xin Jiang, Maosong Sun, and Qun Liu.
\newblock {ERNIE}: Enhanced language representation with informative entities.
\newblock In \emph{Annual Meeting of the Association for Computational
  Linguistics}, 2019.

\end{thebibliography}

\appendix

\section{Prompts}\label{app:mt-app}

The following contains prompts used for evaluation. The prompts are also available in PromptSource~\citep{bach2022promptsource}. A sample with a prompt applied as well as the raw prompts are provided. For raw prompts, double curly brackets are filled with content from the sample when used.

\localtableofcontents

\subsection{SuperGLUE/wsc.fixed}

\subsubsection{Data example}

Prompt name: \textbf{GPT-3 Style}
\begin{verbatim}
Passage: I tried to paint a picture of an orchard, with lemons in the lemon 
trees , but they came out looking more like light bulbs.\n\nQuestion: In the 
passage above, does the pronoun "they" refer to lemon trees?
\end{verbatim}
Answer: \textbf{No}

\subsubsection{Prompts}

\textbf{GPT-3 Style}
\begin{verbatim}
Passage: {{ text }} \n\nQuestion: In the passage above, does the pronoun 
"{{ span2_text }}" refer to {{ span1_text }}?\n\nAnswer:
\end{verbatim}

\noindent\textbf{replaced with}
\begin{verbatim}
{{ text }} In the previous sentence, can the pronoun "{{ span2_text }}" 
be replaced with "{{ span1_text }}"? Yes or no?
\end{verbatim}

\noindent\textbf{the pronoun refers to}
\begin{verbatim}
{{ text }} \nIn the passage above, the pronoun "{{ span2_text }}" refers to 
{{ span1_text }}. True or false?
\end{verbatim}

\noindent\textbf{does p stand for}
\begin{verbatim}
{{ text }} Here, does "{{ span2_text.lower() }}" stand for {{ span1_text }}? 
Yes or no?
\end{verbatim}

\noindent\textbf{the pronoun refers to}
\begin{verbatim}
{{ text }} \nIn the passage above, the pronoun "{{ span2_text }}" refers to 
{{ span1_text }}. True or false?
\end{verbatim}

\subsection{SuperGLUE/wic}

\subsubsection{Data example}

Prompt name: \textbf{GPT-3 Style}
\begin{verbatim}
As he called the role he put a check mark by each student's name.
\n\nA check on its dependability under stress.\n\nQuestion: Is the 
word 'check' used in the same sense in the two sentences above?
\end{verbatim}

\subsubsection{Prompts}

\textbf{GPT-3 Style}
\begin{verbatim}
{{sentence1}}\n\n{{sentence2}}\n\nQuestion: Is the word ''{{word}}'' 
used in the same sense in the two sentences above?
\end{verbatim}

\noindent\textbf{question-context-meaning-with-label}
\begin{verbatim}
Does the word "{{word}}" have the same meaning in these two sentences? 
Yes, No?\n\n{{sentence1}}\n\n{{sentence2}}
\end{verbatim}

\noindent\textbf{GPT-3-prompt-with-label}
\begin{verbatim}
{sentence1}}\n\n{{sentence2}}\n\nQuestion: Is the word ''{{word}}'' used 
in the same sense in the two sentences above? Yes, No?
\end{verbatim}

\noindent\textbf{polysemous}
\begin{verbatim}
The word "{{word}}" has multiple meanings. Does it have the same meaning in 
sentences 1 and 2? Yes or no? Sentence 1: {{sentence1}} Sentence 2: 
{{sentence2}}
\end{verbatim}

\noindent\textbf{similar-sense}
\begin{verbatim}
{{sentence1}}\n\n{{sentence2}}\n\nSimilar sense of {{word}}?
\end{verbatim}

\subsection{SuperGLUE/boolq}

\subsubsection{Data example}

Prompt name: \textbf{GPT-3 Style}
\begin{verbatim}
Phantom pain -- Phantom pain sensations are described as perceptions that an 
individual experiences relating to a limb or an organ that is not physically 
part of the body. Limb loss is a result of either removal by amputation or 
congenital limb deficiency. However, phantom limb sensations can also occur 
following nerve avulsion or spinal cord injury.\nQuestion: is pain experienced 
in a missing body part or paralyzed area\nAnswer:
\end{verbatim}
Answer: \textbf{Yes}

\subsubsection{Prompts}

\textbf{GPT-3 Style}
\begin{verbatim}
{{ passage }} \nQuestion: {{ question }}\nAnswer:
\end{verbatim}

\noindent\textbf{yes\_no\_question}
\begin{verbatim}
Text: {{passage}}\n\nAnswer the following yes/no question:
{{question}}? Yes or no?
\end{verbatim}      

\noindent\textbf{exam}
\begin{verbatim}
EXAM\n1. Answer by yes or no.\n\nDocument: {{passage}}\n
Question: {{question}}?
\end{verbatim}

\noindent\textbf{based on the following passage}
\begin{verbatim}
Based on the following passage, {{ question }}? {{ passage }}
\end{verbatim}

\noindent\textbf{could you tell me…}
\begin{verbatim}
{ passage }} \n\nHaving read that, could you tell me {{ question }}?
\end{verbatim}

\subsection{SuperGLUE/axb \& SuperGLUE/axg}

\subsubsection{Data example}

Prompt name: \textbf{GPT-3 style}
\begin{verbatim}
The taxpayer met with the accountant to get help filing his taxes.\n\n
Question: The accountant sought help filing taxes. True or False?
\end{verbatim}
Answer: \textbf{False}

\subsubsection{Prompts}

\noindent\textbf{GPT-3 style}
\begin{verbatim}
{{sentence1}}\n\nQuestion: {{sentence2}} True or False?
\end{verbatim}

\noindent\textbf{MNLI Crowdsource}
\begin{verbatim}
{{sentence1}} Using only the above description and what you know about 
the world, is "{{sentence2}}" definitely correct? Yes or no?
\end{verbatim}

\noindent\textbf{can we infer}
\begin{verbatim}
Suppose {{sentence1}} Can we infer that "{{sentence2}}"? Yes or no?
\end{verbatim}

\noindent\textbf{guaranteed true}
\begin{verbatim}
Given {{sentence1}} Is it guaranteed true that "{{sentence2}}"? Yes or no?
\end{verbatim}

\noindent\textbf{justified in saying}
\begin{verbatim}
{{sentence1}} Are we justified in saying that "{{sentence2}}"? Yes or no?
\end{verbatim}

\subsection{XNLI \& SuperGLUE/CB}

\subsubsection{Data example}

Prompt name: \noindent\textbf{GPT-3 style}
\begin{verbatim}
Well, I wasn't even thinking about that, but I was so frustrated, and, I ended up 
talking to him again.\n\nQuestion: I havent spoken to him again. True, False, or
Neither?
\end{verbatim}
Answer: \textbf{False}

\subsubsection{Prompts}
\noindent\textbf{GPT-3 style}
\begin{verbatim}
{{premise}}\n\nQuestion: {{hypothesis}} True, False, or Neither?
\end{verbatim}

\noindent\textbf{MNLI crowdsource}
\begin{verbatim}
{{premise}} Using only the above description and what you know about the world, 
"{{hypothesis}}" is definitely correct, incorrect, or inconclusive?
\end{verbatim}

\noindent\textbf{can we infer}
\begin{verbatim}
Suppose {{premise}} Can we infer that "{{hypothesis}}"? Yes, no, or maybe?
\end{verbatim}

\noindent\textbf{guaranteed/possible/impossible}
\begin{verbatim}
Assume it is true that {{premise}} \n\nTherefore, \"{{hypothesis}}\" is 
{{\"guaranteed\"}}, {{\"possible\"}}, or {{\"impossible\"}}?
\end{verbatim}

\noindent\textbf{justified in saying}
\begin{verbatim}
{{premise}} Are we justified in saying that "{{hypothesis}}"? Yes, no, 
or maybe?
\end{verbatim}

\subsection{XWinograd}

\subsubsection{Data example}

Prompt name: \textbf{Replace}
\begin{verbatim}
The city councilmen refused the demonstrators a permit because _ feared 
violence.\nReplace the _ in the above sentence with the correct option: 
\n- the demonstrators\n- The city councilmen
\end{verbatim}
Answer: \textbf{The city councilmen}

\subsubsection{Prompts}

\textbf{Replace}
\begin{verbatim}
{{sentence}}\nReplace the _ in the above sentence with the correct option:
\n- {{option1}}\n- {{option2}}
\end{verbatim}

\noindent\textbf{True or False}
\begin{verbatim}
The _ in the sentence below refers to {{option1}}. True or False? 
{{sentence}}
\end{verbatim}

\noindent\textbf{does underscore refer to}
\begin{verbatim}
{{sentence}} In the previous sentence, does _ refer to 
{{ option1 }} or {{ option2 }}?
\end{verbatim}

\noindent\textbf{underscore refer to}
\begin{verbatim}
{{sentence}}\n What does the _ in the above sentence refer to? 
{{ option1 }} or {{ option2 }}?
\end{verbatim}

\noindent\textbf{stand for}
\begin{verbatim}
In the sentence below, does the _ stand for {{answer_choices[0]}} or 
{{answer_choices[1]}}? {{sentence}}
\end{verbatim}

\subsection{XCOPA \& SuperGLUE/COPA}

\subsubsection{Data example}

Prompt name: \textbf{C1 or C2? premise, so/because...}
\begin{verbatim}
"It was fragile." or "It was small."? The item was packaged in bubble wrap.  
because
\end{verbatim}
Answer: \textbf{It was fragile.}

\subsubsection{Prompts}

\noindent\textbf{C1 or C2? premise, so/because...}
\begin{verbatim}
{{ answer_choices[0] }}" or "{{ answer_choices[1] }}"? {{ premise }} 
{% if question == "cause" %} because {% else %} so {% endif %}
\end{verbatim}

\noindent\textbf{best\_option}
\begin{verbatim}
{{ premise }} \n\nWhat's the best option?\n- {{choice1}}\n- {{choice2}}\n\
\nWe are looking for {% if question == \"cause\" %} a cause {% else %} 
an effect {% endif %}    
\end{verbatim}

\noindent\textbf{cause\_effect}
\begin{verbatim}
{{ premise }}\nSelect the most plausible {% if question == "cause" %} cause: 
{% else %} effect: {% endif %}\n- {{choice1}}\n- {{choice2}}
\end{verbatim}

\noindent\textbf{i\_am\_hesitating}
\begin{verbatim}
{{ premise }} \n\nI am hesitating between two options. Help me choose the 
more  likely {% if question == \"cause\" %} cause: {% else %} 
effect: {% endif %}\n- {{choice1}}\n- {{choice2}}      
\end{verbatim}

\noindent\textbf{plausible\_alternatives}
\begin{verbatim}
{{ premise }} {% if question == "cause" %} This happened because... 
{% else %} As a consequence... {% endif %} Help me pick the more 
plausible option:\n- {{choice1}}\n- {{choice2}}
\end{verbatim}

\subsection{XStoryCloze \& Story Cloze}

\subsubsection{Data example}

XStoryCloze and Story Cloze are not publicly available datasets. Please contact the authors of \citet{lin2021xglm} for XStoryCloze and \citet{mostafazadeh2017lsdsem} for Story Cloze samples.

\subsubsection{Prompts}

\noindent\textbf{Answer Given options}
\begin{verbatim}
{{input_sentence_1}} {{input_sentence_2}} {{input_sentence_3}} 
{{input_sentence_4}} What is a possible continuation for the story 
given the following options ? - {{answer_choices | join("\n- ")}}
\end{verbatim}

\noindent\textbf{Choose Story Ending}
\begin{verbatim}
Read the following story :\n\n{{input_sentence_1}}\n{{input_sentence_2}}\n
{{input_sentence_3}}\n{{input_sentence_4}}\n\nChoose a possible ending for the
previous story from the following options: \n- {{answer_choices | join(\"\\\n- \")}}
\end{verbatim}

\noindent\textbf{Story Continuation and Options}
\begin{verbatim}
What is a possible continuation for the following story ? \n\n{{input_sentence_1}}
\n\{{input_sentence_2}}\n{{input_sentence_3}}\n{{input_sentence_4}}\n\nChoose from 
the following options:\n- {{answer_choices | join(\"\\n- \")}}
\end{verbatim}
     
\noindent\textbf{Generate Ending}
\begin{verbatim}
Generate a possible ending for the following story: {{input_sentence_1}} 
{{input_sentence_2}} {{input_sentence_3}} {{input_sentence_4}}
\end{verbatim}

\noindent\textbf{Novel Correct Ending}
\begin{verbatim}
I read the following novel: {{input_sentence_1}} {{input_sentence_2}}
{{input_sentence_3}} {{input_sentence_4}} What do you think is the most probable
ending? You can choose from the following options: - {{answer_choices | join("\n-")}}
\end{verbatim}

\subsection{WMT}

Prompts for Section~\ref{sec:eval:mt:wmt}, where we compare prompts in both zero-shot and 1-shot settings for four language directions (en$\leftrightarrow$\{hi,fr\}).

\subsubsection{Data example}

The prompt names and content are specific to the language direction. The prompts below each exist in four versions, where ``l1'' and ``l2'' are replaced by the language codes of the source and target languages respectively (en, fr or hi) and ``L1'' and ``L2'' are replaced by the language names of the source and target languages respectively (English, French or Hindi).

Prompt name: \textbf{a\_good\_translation-l1-l2-source+target}
\begin{verbatim}
Given the following source text in English: Spectacular Wingsuit Jump Over 
Bogota , a good French translation is:
\end{verbatim}
Answer: \textbf{Spectaculaire saut en "wingsuit" au-dessus de Bogota}

\subsubsection{Prompts}

\textbf{a\_good\_translation-l1-l2-source+target}
\begin{verbatim}
Given the following source text in L1:  {{translation[l1]}} , a 
good L2 translation is: ||| {{translation[l2]}}
\end{verbatim}

\noindent\textbf{gpt-3-l1-l2-target}
\begin{verbatim}
Q: What is the {{L2}} translation of {{translation[l2]}} A:
\end{verbatim}

\noindent\textbf{version-l1-l2-target}
\begin{verbatim}
If the original version says: {{translation[l1]}}; then the L2 
version should say:
\end{verbatim}

\noindent\textbf{xglm-l1-l2-source+target}
\begin{verbatim}
{{L1}}: {{translation[l1]}} = {{L2}}:
\end{verbatim}

\subsection{DiaBLa}

Prompts for contextual MT results shown in Table~\ref{tab:diabla-context-results}.

\subsubsection{Data example}

Prompt name: \textbf{xglm-source+target}
\begin{verbatim}
English: We appear to have stopped moving. = French:
\end{verbatim}
Answer: \textbf{J'ai l'impression qu'on s'est arrêtés.}

\subsubsection{Prompt}

\textbf{xglm-source+target}
\begin{verbatim}
{% set trg_lang ="French" %}{% set src_lang ="English" %}
{% if utterance_meta.lang == "french" %}
  {% set trg_lang = "English" %}{% set src_lang = "French" %}
{% endif %}
{{ src_lang }}: {{ orig }} = {{ trg_lang }}: ||| {{ ref }}
\end{verbatim}

The dialogue set is bilingual (between native English and native French speakers). In few-shot setups, few-shot examples are by default in either language direction (English$\rightarrow$French or French$\rightarrow$English) regardless of the direction of the current example. We implemented some additional DiaBLa tasks that control the language direction of the few-shot example to be the same or the opposite direction to the current example (\texttt{diabla\_1\_shot\_context\_same} and \texttt{diabla\_1\_shot\_context\_opposite}).

\subsection{Flores-101 (MT)}
The prompts are specific to each language pair as indicated by the source and target languages in the prompt name. Below is an example for French$\rightarrow$Catalan translation.

\subsubsection{Data example}

Prompt name: \textbf{xglm-French-Catalan-source+target}
\begin{verbatim}
French: L'aménagement vieillot, l'absence d'équipements récents et un certain 
charme à l'ancienne font également partie de leur caractère. = Catalan: ||| 
Els accessoris vintage, la manca de les comoditats modernes i un cert envelliment 
elegant també formen part del seu caràcter
\end{verbatim}
\subsubsection{Prompt}

\textbf{xglm-French-Catalan-source+target}
\begin{verbatim}
French: {{ sentence_fra }} = Catalan: ||| {{ sentence_cat }}
\end{verbatim}

\subsection{CrowS-Pairs}

\subsubsection{Data example}

\begin{verbatim}
Sentence A: He couldn't figure out the issue with the rope, but he tried not to get 
too down on himself, most black kids don't have the best education growing up.\n
Sentence B: He couldn't figure out the issue with the rope, but he tried not to get 
too down on himself, most white kids don't have the best education growing up.\n
Select the option that is the most plausible:\n- Sentence A demonstrates 
a stereotype\n- Sentence B violates a stereotype
\end{verbatim}
Answer: \textbf{Sentence A demonstrates a stereotype}

\subsubsection{Prompts}

\textbf{demontraste\_or\_violate}
\begin{verbatim}
Sentence A: {{sent_more}}\nSentence B: {{sent_less}}\nSelect the option 
that is the most plausible:\n- {{answer_choices[0]}}\n- {{answer_choices[1]}}
\end{verbatim}

\end{document}